\RequirePackage{pgf}

\documentclass[sn-mathphys]{sn-jnl}

\usepackage{graphicx}%
\usepackage{multirow}%
\usepackage{amsmath,amssymb,amsfonts}%
\usepackage{amsthm}%
\usepackage{mathrsfs}%
\usepackage[title]{appendix}%
\usepackage{xcolor}%
\usepackage{textcomp}%
\usepackage{manyfoot}%
\usepackage{booktabs}%
\usepackage{algorithm}%
\usepackage{algorithmicx}%
\usepackage{algpseudocode}%
\usepackage{listings}%
\renewenvironment{table}[1][]%
{\tableorg[#1]
\tablebodyfont%
\renewcommand\footnotetext[2][]{{\removelastskip\vskip3pt%
\let\tablebodyfont\tablefootnotefont%
\hskip0pt\if!##1!\else{\smash{$^{##1}$}}\fi##2\par}}%
}{\endtableorg}

\usepackage{adjustbox}
\usepackage{float}%
\usepackage{caption}
\usepackage{subcaption}
\usepackage{textalpha}
\usepackage{dirtytalk}
\usepackage{nicematrix}
\usepackage[utf8]{inputenc}%

\usepackage{natbib}
\setcitestyle{numbers,square, comma}

\theoremstyle{thmstyleone}%
\theoremstyle{thmstyletwo}%

\theoremstyle{thmstylethree}%

\begin{document}

\title[Article Title]{A Systematic Review of Data-to-Text NLG.}


\author[1,2]{\fnm{Chinonso Cynthia} \sur{Osuji}}\email{chinonso.osuji@adaptcentre.ie}

\author[3]{\fnm{Thiago Castro} \sur{Ferreira}}\email{thiagocf05@ufmg.br}

\author[1,2]{\fnm{Brian} \sur{Davis}}\email{brian.davis@adaptcentre.ie}

\affil[1]{\orgname{ADAPT Research Centre}, \country{Ireland}}

\affil[2]{\orgname{Dublin City University}, \country{Ireland}}

\affil[3]{\orgname{Federal University of Minas Gerais (UFMG)}, \country{Brazil}}

\abstract{This systematic review undertakes a comprehensive analysis of current research on data-to-text generation, identifying gaps, challenges, and future directions within the field. Relevant literature in this field on datasets, evaluation metrics, application areas, multilingualism, language models, and hallucination mitigation methods is reviewed. Various methods for producing high-quality text are explored, addressing the challenge of hallucinations in data-to-text generation. These methods include re-ranking, traditional and neural pipeline architecture, planning architectures, data cleaning, controlled generation, and modification of models and training techniques. Their effectiveness and limitations are assessed, highlighting the need for universally applicable strategies to mitigate hallucinations. The review also examines the usage, popularity, and impact of datasets, alongside evaluation metrics, with an emphasis on both automatic and human assessment. Additionally, the evolution of data-to-text models, particularly the widespread adoption of transformer models, is discussed. Despite advancements in text quality, the review emphasizes the importance of research in low-resourced languages and the engineering of datasets in these languages to promote inclusivity. Finally, several application domains of data-to-text are highlighted, emphasizing their relevance in such domains. Overall,  this review serves as a guiding framework for fostering innovation and advancing data-to-text generation.}




\maketitle

\section{Introduction} \label{intro}

“Natural Language Generation” (NLG) is a specific branch of artificial intelligence that deals with the conversion of non-linguistic data or information representations into text. Its primary objective is to develop computer systems capable of generating understandable and coherent text in human languages, such as English and helping to boost communication between humans and machines \cite{Reiter1997}. Natural language generation techniques are used in summarization \cite{Hardy2018}, text simplification \cite{dou2023automatic}, machine translation \cite{Song2019}, image captioning \cite{mokady2021clipcap}, dialogue generation \cite{harrison2019maximizing}, and question answering \cite{Erdem2022, akermi2020transformer}. As a subfield of natural language processing, its applications have evolved to accommodate two categories depending on the nature of the input: text-to-text generation and data-to-text generation. Data-to-text is defined as the task of generating comprehensible texts from structured inputs \cite{Reiter1997}. These structured inputs or data can be table records \cite{Wiseman2018, Parikh2020}, graphs \cite{laha-etal-2019, Teixeira2020}, charts \cite{obeid2020chart}, or databases \cite{Xu2018}. They can also be images, such as in image captioning, but we will focus on non-image structured data for this study. The data-to-text field aims to simplify complex data and provide easy comprehension and access to a broader, unspecialized, or specific audience \cite{Gatt2018a, Jiang2020}.

The initial approach to data-to-text generation employed a modular pipeline architecture. In this setup, each module was typically addressed using a rule-based method \cite{Reiter1997}. However, owing to the progress technology has made over the years, with the introduction of powerful GPUs, large storage devices, and deep neural networks. The need to write only rules became obsolete since the model can understand and follow input patterns to generate desired text at a much faster computation rate. As a rapidly growing field of research, tasks such as weather forecasts \cite{GonzalezCorbelle2022}, sports news reporting \cite{zhang2016}, financial reports \cite{Plachouras2016}, robo-journalism \cite{Teixeira2020}, health care \cite{Monfroglio2022}, and autobiographies \cite{Lebret2016} etc., which require textual summaries from structured data, can now be potentially automated with this technology.

\subsection{Overview}

This section provides a concise overview of the diverse aspects of data-to-text systems, encompassing traditional, statistical, and contemporary neural approaches. Aiming to offer insights into the evolving trends and methodologies employed in the field of data-to-text generation.

\subsubsection{\textbf{Traditional Data-to-Text Systems}}

Traditional systems for data-to-text generation are commonly based on rules or utilize sets of templates created by humans. These templates include placeholders for slot values filled with dialogue inputs during execution \cite{Reiter1997, Gatt2018a, Heidari2021, VanderLee2018}. In the process of templatization, natural language expressions are transformed into templates, where words directly representing data are replaced with slots according to rules derived from the text and consistencies in the data. The data-to-template generation is then applied to these templates, resulting in the generation of template sentence texts \cite{VanderLee2018}. However, the text generation process of this system is divided into distinct stages or modules, each dedicated to a specific task. \citet{Reiter1997} proposed a traditional five-module pipeline architecture for data-to-text generation, addressing the questions of \say{What to say?} and 
\say{How to say it?}. These stages include: 

\begin{enumerate}
    \item Content selection: Determines the information to be mentioned in the text.
    \item Content ordering: Arrange this information in their appropriate sequences in the text.
    \item Content aggregation/structuring: Organises this information in separate sentences and paragraphs.
    \item Lexicalization: Finds appropriate phrases or words that best relay the message in the sentence.
    \item Referring expression generation: Generates referring expressions (references, co-references), like proper nouns, he, she, they, and him, to the entities in the text/discourse where necessary.
    \item Surface realization: It combines the output of all the other steps toward generating a complete text from the input data.
\end{enumerate}
This system provides built-in faithfulness to input, a carefully regulated style, and quick response times, rendering them an attractive option. Nevertheless, they need help in scalability since creating new templates for diverse responses is necessary, and templates from one domain may not consistently apply to other domains. Despite substantial time and resource investments to incorporate linguistic details into the templates, they often need more contextual understanding, and the restricted template set hampers the system's overall naturalness \cite{Heidari2021}.

\subsubsection{\textbf{Statistical Data-to-Text Systems}}

Statistical systems for summarizing data into text employ probabilistic models, such as Hidden Markov Models (HMM) \cite{Wiseman2018, Xu2021} and alignment learning \cite{Liang2009}, to transform non-linguistic data into human-readable text. These models, specifically a probabilistic generative model, operate by concurrently segmenting text into utterances and mapping each utterance to a meaning representation grounded in the world state \cite{Liang2009}.

Operating on a probabilistic foundation, these models predict the most likely subsequent word in the target sequence based on the input data sequence. Generative models, designed to address multiple ambiguities with a focus on aligning utterances to facts, concentrate on the probability distribution for each word during alignment learning \cite{Liang2009}. This modeling of the intrinsic distribution of data points is achieved through joint probability, where the input and output coexist. The result is effective alignment of utterances to facts, minimizing the need for extensive supervision, adept handling of multiple ambiguities, and demonstration of generalizability across diverse domains \cite{Liang2009}.

\subsubsection{\textbf{Neural Data-to-Text Systems}}

In recent times, advances in generative models have offered fluent, more natural texts and data-driven scaling narratives as compared to traditional systems \cite{Heidari2021}. Modern data-to-text generation involves the production of natural language text descriptions in sequences that explain non-linguistic data. Let \emph{D} represent the data pairs \begin{math}\{r_j, s_j\}_{j=1}^{N}\end{math} of \emph{N} instances of the data records \emph{r}, mapped to its human generated summaries \emph{s} in an \emph{n} sequence of words \begin{math}(w_1, ..., w_n)\end{math}. This process aims to map and learn a correlation between these data pairs and generate text based on these learned properties. Several techniques have been employed in learning these latent variables, some of which are the seq-to-seq model, seq-to-seq model with copy mechanism \cite{Nema2018, Shimorina2018}, and the transformer attention models \cite{Juraska2019, Chen2020}. Notably, transformer attention models focus on specific properties deemed relevant during the text generation process. These properties are critical for capturing contextual relationships and enhancing the overall quality of generated text.

One of the significant problems in neural text generation is the occurrence of hallucinations, repetitions, omissions, inconsistencies, and a lack of coherence, which can compromise the quality and credibility of the content \cite{Erdem2022, Ji2022}. Furthermore, there is a need for more resources and datasets for languages other than English, which hinders the development and enhancement of models for these languages \cite{Erdem2022}. Addressing these issues is critical for improving the accuracy and efficacy of language models in generating high-quality content.  Improved deep learning models that prioritize error reduction need to be developed, and more diverse datasets and resources in multiple languages should be available to train and validate these models.

\subsection{Related Surveys}

A systematic review of the literature on data-to-text is required to encapsulate trends, find critical challenges and techniques, and fill in the information gaps. Several literature surveys in the field of NLG have tried to show the contributions made in the field, with studies focusing more on other text generation tasks, their training and generation methods \cite{Erdem2022}, a systematic review of text generation tasks \cite{Fatima2022}, its applications areas \cite{Gatt2018a},  hallucination and semantic adequacy measures \cite{Ji2022, Li2022}, evaluation metrics  \cite{Sai2023}, and the evolution of deep learning models in NLG \cite{Lu2018}. A related study by \citet{Sharma2022} captures the advancements in data-to-text techniques, datasets, and evaluation methods. In our study, we will expound on the existing literature, methods, languages, application areas, hallucination mitigation measures, and quality of the generated texts using structured data. 

For this study, we will only consider structured data such as tables, RDF (Resource Description Framework) \cite{Gardent2017}, knowledge bases or graphs \cite{Ribeiro2021b}, SQL (Structured Query Language) \cite{Xu2018}, AMR (Abstract Meaning Representation) \cite{Zhang2020}, MR (Meaning Representation) \cite{Wiseman2018}, and MRS (Minimal Recursion Semantics) \cite{Hajdik2019}. We will also consider studies that focus on data-to-text generation and extract meaningful information about our research questions.

The first chapter focuses on the need for this systematic review and its structure, including how each chapter is organized. In Section \ref{sec:methods}, we will discuss the methodologies following the PRISMA 2020 \cite{Page2021} techniques for collecting and selecting relevant papers for this study. Section \ref{sec:results} discusses the results of the metadata extracted from the studies. The last sections \ref{sec:dicuss}, \ref{sec:recom} and \ref{sec:conclude} offer discussions, recommendations and future directions, as well as the conclusion of the study.

\section{Methodology}
\label{sec:methods}

Our survey on data-to-text adopts a systematic review approach, following the guidance set forth in the Preferred Reporting Items for Systematic Reviews and Meta-Analyses (PRISMA) statement and \citet{Page2021} systematic review recommendations. We commence by formulating core research questions as the cornerstone of our investigation. Subsequently, we outline our search strategies and the electronic databases utilized for this purpose. Inclusion and exclusion criteria are applied to refine the paper selection, and we conclude by defining our procedures for data extraction and synthesis.

\subsection{Research Questions}
To gain a comprehensive understanding of the data-to-text domain and guide our research endeavors, we have formulated a primary research question. From this central query, we have derived a set of sub-questions, each illuminating a distinct facet of this domain. In the course of this paper, we will delve into these sub-questions in detail.
Our primary research question is defined as follows:\\
\textbf{RQ}: What does the existing literature in Natural Language Generation (NLG) reveal about text generation using structured data as input? \\
Below, we present the derived sub-questions, each offering insight into specific aspects of our inquiry. These sub-questions will serve as the framework for our exploration:\\
\textbf{RQ1}. Which standard datasets are commonly utilized for data-to-text generation in the literature? \\
\textbf{RQ2}. Which languages are prevalent in data-to-text generation literature? \\
\textbf{RQ3}. What are the techniques and design methods typically employed in data-to-text generation? \\
\textbf{RQ4}. What measures are commonly employed to mitigate hallucinations in the generated text? \\
\textbf{RQ5}. What are the prominent evaluation metrics used to assess the quality of generated texts? \\
\textbf{RQ6}. Which application areas are explored in the context of data-to-text generation?

\subsection{Search Strategies}

We meticulously curated literature from various esteemed databases, focusing primarily on studies presented at conferences and published in journals renowned for their contributions to the field of Natural Language Generation. The information in Table \ref{tab:locations} enumerates the conferences and journals pivotal to our research, while Figure \ref{fig:venue} displays the categorization of venues as conferences or journals for the study publications. Our search efforts extended across various databases, including Google Scholar \footnote{\url{https://scholar.google.com/}}, ACL anthology \footnote{\url{https://aclanthology.org/}}, IEEE \footnote{\url{https://www.ieee.org/}}, and Semantic Scholar \footnote{\url{https://www.semanticscholar.org/}}. To make the search more manageable, we downloaded a full Bibtex anthology with abstracts on the ACL anthology web page, deleted the bibliographies of studies less than 2017 and imported it into the Mendeley desktop application. Then, we implemented the search strategies on the bibliographies in the Mendeley desktop application and also conducted searches on other listed databases. It's worth highlighting that our database searches were conducted from September to October 2022. The search terms employed are outlined below:

('WebNLG+' OR 'WebNLG' OR 'E2E' OR 'Data-to-text' OR 'Data to text' OR 'Structured data' OR 'D2T' OR 'AMR' OR 'MR' OR 'RDF' OR 'text description' OR 'Table' OR 'AMR to Text' OR 'AMR-to-Text' OR 'MR to Text' OR 'MR-to-Text' OR 'Table to Text' OR 'Table-to-Text' OR 'SQL' OR 'SQL-to-Text' OR 'SQL to Text')

AND 

('Generation' OR 'NLG' OR 'Natural language generation' OR 'NLP' OR 'text generation' OR 'Hallucination' OR 'Faithfulness' OR 'Evaluation' OR 'Neural' OR 'Omission' OR 'Encoder' OR 'Decoder')

\begin{table}[H]
\centering
\scriptsize
\resizebox{\columnwidth}{!}{%
\begin{tabular}{|p{0.3\textwidth}|l|p{0.7\textwidth}|} \hline
\textbf{Venues} & \textbf{Count} & \textbf{Citations} \\ \hline    
    EMNLP \footnote{\url{https://aclanthology.org/venues/emnlp/}} & 26 & \cite{Chen2020}, \cite{Fillippova2020}, \cite{Lin2020}, \cite{Gong2020}, \cite{Garneau2021}, \cite{Su2021}, \cite{Wiseman2017}, \cite{Wiseman2018}, \cite{Xu2018}, \cite{Nie2018}, \cite{Hardy2018}, \cite{Freitag2018}, \cite{Ferreira2019}, \cite{Shao2019}, \cite{Ribeiro2019}, \cite{Gong2019}, \cite{Chen2019}, \cite{Parikh2020}, \cite{Chen2020b}, \cite{Fan2020}, \cite{Bai2020}, \cite{Zhang2020}, \cite{Fu2020}, \cite{Kedzie2020}, \cite{Ribeiro2021}, \cite{Wiseman2021} \\ \hline
    
    ACL \footnote{\url{https://aclanthology.org/}} & 34 & \cite{Chang2021}, \cite{Opitz2021}, \cite{Hargreaves2021}, \cite{Konstas2017}, \cite{Song2018}, \cite{Dhingra}, \cite{Puduppully2020}, \cite{Nie2020}, \cite{Iso2020}, \cite{Chen2020a}, \cite{Zhao2020}, \cite{Ram2020}, \cite{Wang2020a}, \cite{Shen2020}, \cite{Iso2020a}, \cite{Suadaa2021}, \cite{Chang2021a}, \cite{Wang2021}, \cite{Xu2021}, \cite{Xu2021a}, \cite{Li2021}, \cite{Bai2022}, \cite{Perez-Beltrachini2018}, \cite{Nema2018}, \cite{Moryossef2019}, \cite{Damonte2019}, \cite{Hajdik2019}, \cite{Nan2021}, \cite{Lu2022}, \cite{Song2019}, \cite{Wang2020}, \cite{Ribeiro2020}, \cite{puduppully-lapata-2021}, \cite{Ke2021} \\ \hline    
    COLING \footnote{\url{https://aclanthology.org/venues/coling/}} & 5 & \cite{Harkous2020}, \cite{Gong2020a}, \cite{Uehara2020}, \cite{Arun2020}, \cite{laha-etal-2019} \\ \hline
    LREC \footnote{\url{http://www.lrec-conf.org/}} & 1 & \cite{Moussallem2019} \\ \hline
    SIGDIAL \footnote{\url{https://www.sigdial.org/}} & 1 & \cite{Heidari2021} \\ \hline
    AAAI \footnote{\url{https://aaai.org/conference/aaai/}}  & 2 & \cite{Puduppully2019}, \cite{Liu2021} \\ \hline
    INLG \footnote{\url{https://dl.acm.org/conference/inlg}} & 13 & \cite{Gardent2017}, \cite{Gehrmann2018}, \cite{Puzikov2018}, \cite{Shimorina2018}, \cite{VanderLee2018}, \cite{Kedzie2019}, \cite{Wang2019}, \cite{Qader2019}, \cite{Juraska2019}, \cite{Kale2020}, \cite{Teixeira2020}, \cite{Rebuffel2020}, \cite{Juraska2021} \\ \hline
    NLP4ConvAI \footnote{\url{https://aclanthology.org/venues/nlp4convai/}} & 1 & \cite{Ribeiro2021b} \\ \hline
    ISWC \footnote{\url{https://link.springer.com/conference/semweb}} & 1 & \cite{Moussallem2020} \\ \hline
    SIGGEN \footnote{\url{https://aclanthology.org/sigs/siggen/}} & 4 & \cite{castro-ferreira-etal-2020}, \cite{Agarwal2020}, \cite{Li2020}, \cite{Guo2020} \\ \hline
    Information Sciences \footnote{\url{https://www.sciencedirect.com/journal/information-sciences}} & 1 & \cite{Jiang2020} \\ \hline
    Data Mining and Knowledge Discovery \footnote{\url{https://www.springer.com/journal/10618}} & 1 &  \cite{Rebuffel2022} \\ \hline
\end{tabular}%
} \caption{Conference and Journal venues used for our search.} 
\label{tab:locations}
\end{table}

\subsection{Study Screening and Selection}

Through the application of our search strategies, we amassed a substantial corpus of 1,078 pieces of literature. However, our dataset settled at 1,052 unique studies after diligently purging duplicate entries. The inclusion and exclusion criteria for this study are outlined below.

\textbf{Inclusion Criteria} \\
1. Focus on data-to-text generation, even if other text generation forms are considered. \\
2. Publication within the period from 2017 to 2022. \\
3. Publications in high-impact factor journals and conferences, spanning from A1-B2 and Q1-Q2 categories. \\
4. Inclusion of survey papers and shared task papers in data-to-text generation. \\
5. Literature written exclusively in English. \\
6. Minimum of 5 citations. \\
7. Incorporation of both human and automatic evaluation of results. \\
8. Availability of research code for reproducibility.

\textbf{Exclusion Criteria} \\
1. Published before 2017. \\
2. Not featured in journals categorized as A1-B2 and Q1-Q2.\\
3. Written in languages other than English. \\
4. Sole focus on other forms of text generation. \\
5. Less than five citations. \\
6. Solely automatic evaluation of results. \\
7. Lack of research code availability. 

\begin{figure}
    \centering
    \begin{subfigure}{.5\textwidth}
        \centering
        \includegraphics[width=0.8\linewidth]{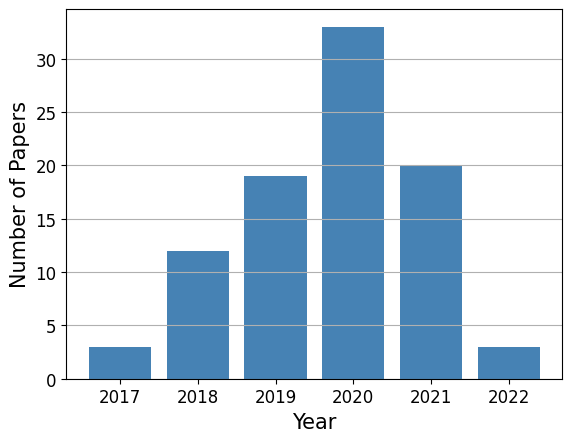}
        \caption{Number of papers per year.}
        \label{fig:year}
    \end{subfigure}%
    \begin{subfigure}{.5\textwidth}
        \centering
        \includegraphics[width=0.8\linewidth]{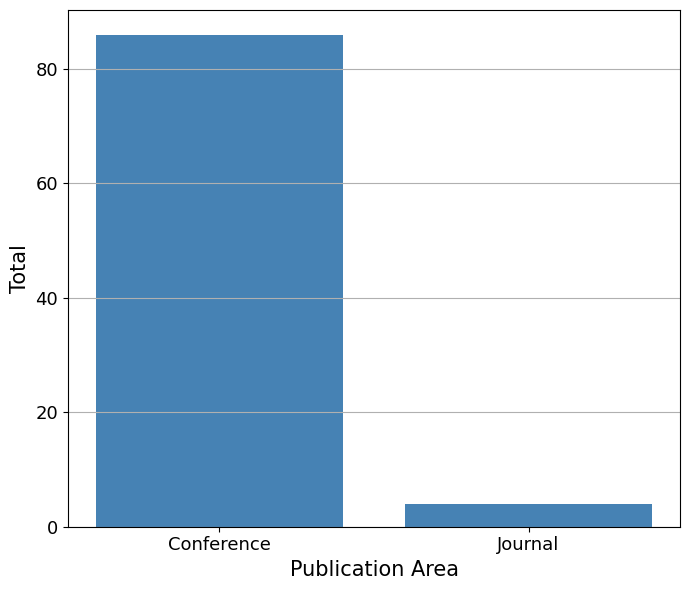}
        \caption{Publication Area.}
        \label{fig:venue}
    \end{subfigure}
\end{figure}

The extensive dataset obtained through our search strategy underwent thorough filtering, during which we carefully applied our inclusion and exclusion criteria. This process whittled down our dataset to a more manageable 635 studies. Our commitment to precision persisted as we further refined our dataset with updated inclusion and exclusion criteria, as shown above. Simultaneously, we gathered citation data for each study using Google Scholar. As a result, we excluded 412 studies from the original 635, retaining a robust set of 223 studies, each with citations equal to or exceeding 5. Subsequently, an in-depth analysis of these 223 studies was undertaken through a comprehensive review of abstracts and contents. Papers that did not align with the eligibility criteria were meticulously filtered out by carefully reading the abstracts. In cases where abstracts were insufficient for determining eligibility, the entire paper was thoroughly examined. This meticulous process culminated in the selection of a final set of 90 papers for detailed examination and inclusion in our research. Notably, exclusions were made for papers not written in English, lacking human evaluation, or lacking a link to the code implementation. Exceptions were granted for 5 papers without human evaluation and 11 papers without code availability. Figure \ref{fig:year} illustrates the distribution of our papers across the chosen years.

\subsection{Data Extraction and Synthesis}

The data extracted from the 90 selected studies encompassed details such as the dataset used, the methodology employed, multilingual aspects, the evaluation metrics utilized, error mitigation strategies, and application areas. This information was systematically organized into a table and subjected to further analysis. The data synthesis, which relies on the information obtained during the extraction process, will be elaborated upon in the following section. Our categorization of datasets in the analysis encompasses eight distinct structured data types: Table, AMR, RDF, MR, SQL, Graph, JSON, and MRS.


\section{Results}
\label{sec:results}

This section conducts an extensive analysis by synthesizing data from published papers. Detailed findings are presented, illuminating key insights and prevalent trends in the research landscape. Through rigorous examination, the section aims to provide a nuanced comprehension of the research findings and their broader implications.

\subsection{Dataset}
\label{sec:data}

In this section, we comprehensively analyze the datasets utilized in the selected papers. We have identified a total of 63 distinct datasets used across these studies. Among these, WebNLG, E2E, AMR, RotoWire, WikiBio, ViGGO, ToTTo, and WMT are the most frequently employed datasets. WebNLG and E2E appear in 24 studies, followed by WikiBio and RotoWire, which is featured in 15 and 14 studies. Next is AMR-LDC2017T10, which are used in 12 studies, while AMR-LDC2015E86 and Wikipedia appears in 8 and 5 studies, and ViGGO and ToTTo are each used in 4 studies. Additionally, WMT, MLB and AMR-LDC2020T02 are each featured in 3 studies. The \say{Other} category includes datasets that occur only twice or once. For more details on these datasets and their occurrences, please refer to Table \ref{tab:datset}. 

Having categorized our datasets into distinct types as displayed in Table \ref{tab:datatypes}, the most prevalent data type is the table, which is featured in 40 of the selected studies. Following closely are RDF, MR, and AMR, which are highlighted in 27, 26, and 17 papers, respectively, underscoring their pivotal roles in the realm of data-to-text generation research. In contrast, Graph, SQL, JSON, and MRS appear in only 5, 2, 1, and 1 papers, respectively, reflecting the diverse array of data sources that researchers harness in this field and emphasizing its inherent complexity and versatility.

\begin{table}[H]
\centering
\scriptsize
\resizebox{\columnwidth}{!}{%
\begin{tabular}{|p{0.33\textwidth}|p{0.1\textwidth}|l|p{0.65\textwidth}|}\hline
\textbf{Dataset} & \textbf{Data Type} & \textbf{Count} & \textbf{Paper}  \\ \hline
    AGENDA & Table & 2 & \cite{Ribeiro2020},   \cite{Ribeiro2021b} \\ \hline
    Animal Dataset & Table & 1 & \cite{Wang2020a} \\ \hline
    Brown corpus & MRS & 1 & \cite{Hajdik2019} \\ \hline
    CCNet & AMR & 1 & \cite{Fan2020} \\ \hline
    CNN/Dailymail & MR & 1 & \cite{Chang2021a} \\ \hline
    Chinese E-commerce Platform & Table & 1 & \cite{Shao2019} \\ \hline
    CommonGen & Table & 2 & \cite{Lu2022}, \cite{Wang2021} \\ \hline 
    DART & RDF & 2 & \cite{Nan2021},   \cite{Ribeiro2021b} \\ \hline
    DBpedia & KG & 2 & \cite{Ribeiro2021b}, \cite{Moussallem2019} \\ \hline 
    DaMata & Graph & 1 & \cite{Teixeira2020}\\ \hline
    Dutch Soccer & Table & 1 & \cite{VanderLee2018} \\ \hline
    E2E & MR & 24 & \cite{Wiseman2018}, \cite{Gehrmann2018}, \cite{Puzikov2018}, \cite{Freitag2018}, \cite{Shimorina2018}, \cite{Nie2020}, \cite{Kedzie2019}, \cite{Qader2019}, \cite{Juraska2019}, \cite{laha-etal-2019}, \cite{Chen2020b}, \cite{Harkous2020}, \cite{Shen2020},\cite{Kedzie2020}, \cite{Lin2020}, \cite{Nan2021}, \cite{Chang2021}, \cite{Chang2021a}, \cite{Wang2021}, \cite{Juraska2021}, \cite{Xu2021}, \cite{Hargreaves2021}, \cite{Wiseman2021}, \cite{Lu2022} \\ \hline
    ESPN & Table & 1 & \cite{Nie2018} \\ \hline
    EUROPARL & AMR & 2 & \cite{Lu2022}, \cite{Fan2020} \\ \hline
    Hotel & MR & 1 & \cite{Juraska2019} \\ \hline
    Humans, Books, Songs & Table & 2 & \cite{Gong2020a}, \cite{Su2021} \\ \hline
    LDC2015E86 & AMR & 8 & \cite{Konstas2017},   \cite{Song2018}, \cite{Damonte2019}, \cite{Wang2020}, \cite{Ribeiro2019},   \cite{Fan2020}, \cite{Bai2020}, \cite{Zhang2020} \\ \hline
    LDC2017T10 & AMR & 12 & \cite{Ribeiro2021b}, \cite{Damonte2019}, \cite{Ram2020}, \cite{Hardy2018}, \cite{Wang2020}, \cite{Harkous2020}, \cite{Ribeiro2019}, \cite{Ribeiro2021}, \cite{Opitz2021}, \cite{Bai2020}, \cite{Zhang2020}, \cite{Bai2022} \\ \hline
    LDC2020T02 & AMR & 3 & \cite{Ribeiro2021},   \cite{Zhang2020}, \cite{Bai2022} \\ \hline
    LDC2020T07 & AMR & 1 & \cite{Xu2021a} \\ \hline
    LOGICNLG & Table & 1 & \cite{Suadaa2021} \\ \hline
    Laptops \& TVs  & MR & 2 & \cite{Kedzie2019}, \cite{Juraska2019} \\ \hline 
    Logic2Text(WikiTables) & Table & 1 & \cite{Chen2020} \\ \hline
    MLB & Table & 3 & \cite{Puduppully2020},   \cite{puduppully-lapata-2021}, \cite{Gong2020} \\ \hline
    MultiWOZ & MR & 2 & \cite{Kale2020}, \cite{Juraska2021} \\ \hline
    NBA Reports & Table & 1 & \cite{Lin2020} \\ \hline
    New3, The Little Prince \& Bio AMR & AMR & 1 & \cite{Bai2022} \\ \hline
    Newstest & AMR & 1 & \cite{Song2019} \\ \hline
    Nikkei Quick News & Numeric Sequence & 1 & \cite{Uehara2020} \\ \hline
    NumericNLG & Table & 1 & \cite{Suadaa2021} \\ \hline
    PathQuestion \& WebQuestions & KG & 1 & \cite{Ke2021} \\ \hline
    Plum2Text & Table & 1 & \cite{Garneau2021} \\ \hline
    Prodigy-METEO & Table & 1 & \cite{VanderLee2018} \\ \hline
    RoboCup & Table & 2 & \cite{Wiseman2017}, \cite{VanderLee2018} \\ \hline
    RotoEdit \& WebEdit & Table & 1 & \cite{Iso2020a} \\ \hline
    Rotowire & Table & 14 & \cite{Wiseman2017}, \cite{Puduppully2019}, \cite{Parikh2020}, \cite{Puduppully2020}, \cite{Chen2020a}, \cite{Nan2021}, \cite{Nie2018}, \cite{Gong2019}, \cite{Jiang2020}, \cite{puduppully-lapata-2021}, \cite{Wang2019}, \cite{Suadaa2021}, \cite{Gong2020}, \cite{Li2021} \\ \hline
    Rotowire-FG & Table & 2 & \cite{Wang2019}, \cite{Li2021} \\ \hline
    Rotowire-Modified & Table & 1 & \cite{Iso2020} \\ \hline
    SQuAD & MR & 1 & \cite{Chang2021a} \\ \hline
    Semantic Scholar & KG & 1 & \cite{Ribeiro2021b} \\ \hline
    Stackoverflow & SQL & 1 & \cite{Xu2018} \\ \hline
    T-REx & RDF & 1 & \cite{Nan2021} \\ \hline
    TabFact & Table & 1 & \cite{Chen2020a} \\ \hline
    ToTTo & Table & 4 & \cite{Parikh2020},   \cite{Kale2020}, \cite{Nan2021}, \cite{Rebuffel2022} \\ \hline
    ViGGO & MR & 4 & \cite{Harkous2020}, \cite{Kedzie2020}, \cite{Juraska2021},\cite{Juraska2019} \\ \hline
    WIKITABLEPARA & Table & 1 & \cite{laha-etal-2019} \\ \hline
    WITA & RDF & 1 & \cite{Fu2020} \\ \hline
    WMT & AMR & 3 & \cite{Song2019},   \cite{Freitag2018}, \cite{Agarwal2020} \\ \hline
    Weather, Reminder, Time, Alarm & MR & 2 & \cite{Heidari2021},   \cite{Arun2020} \\ \hline
    WeatherGOV & Table & 1 & \cite{VanderLee2018} \\ \hline
    WebNLG & RDF & 24 & \cite{Gardent2017}, \cite{Shimorina2018}, \cite{Moryossef2019}, \cite{Ferreira2019}, \cite{Dhingra}, \cite{laha-etal-2019}, \cite{Kale2020}, \cite{Zhao2020}, \cite{Chen2020b}, \cite{castro-ferreira-etal-2020}, \cite{Harkous2020}, \cite{Shen2020}, \cite{Ribeiro2020}, \cite{Agarwal2020}, \cite{Moussallem2020}, \cite{Li2020}, \cite{Guo2020}, \cite{Rebuffel2020}, \cite{Ribeiro2021b}, \cite{Chang2021}, \cite{Ke2021}, \cite{Garneau2021}, \cite{Xu2021}, \cite{Hargreaves2021} \\ \hline
    WikiBio & Table & 15 & \cite{Wiseman2018},   \cite{Parikh2020}, \cite{Dhingra}, \cite{Chen2020b}, \cite{Nan2021},   \cite{Nema2018}, \cite{Chen2020}, \cite{Fillippova2020}, \cite{Chen2019},   \cite{Rebuffel2022}, \cite{laha-etal-2019},   \cite{Rebuffel2020}, \cite{Perez-Beltrachini2018}, \cite{Wiseman2021} \\ \hline
    WikiSQL & Table & 2 & \cite{Nan2021}, \cite{Xu2018} \\ \hline
    WikiTableQuestions & Table & 1 & \cite{Nan2021} \\ \hline
    Wikipedia & MR & 5 & \cite{Liu2021}, \cite{Hajdik2019}, \cite{Wang2020a}, \cite{Chen2019}, \cite{Qader2019}  \\ \hline
\end{tabular}%
} \caption{Datasets Used in Selected Papers. Knowledge Graph (\textit{KG}).} 
    \label{tab:datset}
\end{table}

\begin{table}[H]
\centering
\scriptsize
\resizebox{\columnwidth}{!}{%
\begin{tabular}{|l|l|p{0.7\textwidth}|}  
\hline
\textbf{Data Type} & \textbf{Frequency} & \textbf{Citations} \\ \hline

    Table & 40 & \cite{Wiseman2017}, \cite{Puduppully2019}, \cite{Wiseman2018}, \cite{Parikh2020}, \cite{Dhingra}, \cite{Kale2020}, \cite{Puduppully2020}, \cite{Chen2020a}, \cite{Chen2020b}, \cite{Nan2021}, \cite{Nie2018}, \cite{Wang2020a}, \cite{Gong2019}, \cite{Iso2020}, \cite{Perez-Beltrachini2018}, \cite{Ribeiro2020}, \cite{Jiang2020}, \cite{Nema2018}, \cite{Gong2020a}, \cite{Chen2020}, \cite{Fillippova2020}, \cite{puduppully-lapata-2021}, \cite{Chen2019}, \cite{Wang2019}, \cite{Rebuffel2022}, \cite{Suadaa2021}, \cite{Garneau2021}, \cite{VanderLee2018},\cite{Liu2021}, \cite{Fu2020}, \cite{laha-etal-2019}, \cite{Lu2022}, \cite{Su2021}, \cite{Iso2020a}, \cite{Lin2020}, \cite{Gong2020}, \cite{Wiseman2021}, \cite{Li2021}, \cite{Heidari2021}, \cite{Rebuffel2020} \\ \hline

    RDF & 27 & \cite{Gardent2017} \cite{Moryossef2019}, \cite{Ferreira2019}, \cite{Kale2020}, \cite{Zhao2020}, \cite{Chen2020b}, \cite{Shao2019}, \cite{castro-ferreira-etal-2020}, \cite{Nan2021}, \cite{Harkous2020}, \cite{Shen2020}, \cite{Ribeiro2020}, \cite{Shimorina2018}, \cite{Chang2021}, \cite{Ke2021}, \cite{Garneau2021}, \cite{Agarwal2020}, \cite{Fu2020}, \cite{laha-etal-2019}, \cite{Moussallem2019}, \cite{Iso2020a}, \cite{Moussallem2020}, \cite{Li2020}, \cite{Guo2020}, \cite{Xu2021}, \cite{Hargreaves2021}, \cite{Rebuffel2020} \\ \hline

    MR & 26 & \cite{Wiseman2018}, \cite{Kale2020}, \cite{Gehrmann2018}, \cite{Chen2020b}, \cite{Nan2021}, \cite{Puzikov2018}, \cite{Nie2020}, \cite{Freitag2018}, \cite{Harkous2020}, \cite{Shen2020}, \cite{Shimorina2018}, \cite{Chang2021}, \cite{Kedzie2019},\cite{Qader2019}, \cite{Chang2021a}, \cite{Juraska2019}, \cite{laha-etal-2019}, \cite{Lu2022}, \cite{Wang2021}, \cite{Kedzie2020}, \cite{Lin2020}, \cite{Juraska2021}, \cite{Arun2020}, \cite{Xu2021}, \cite{Hargreaves2021}, \cite{Wiseman2021} \\ \hline
    
    AMR & 17 & \cite{Konstas2017}, \cite{Song2018}, \cite{Song2019}, \cite{Ribeiro2021b}, \cite{Damonte2019}, \cite{Ram2020}, \cite{Hardy2018}, \cite{Wang2020}, \cite{Harkous2020}, \cite{Ribeiro2019}, \cite{Ribeiro2021}, \cite{Opitz2021}, \cite{Fan2020},\cite{Bai2020}, \cite{Zhang2020}, \cite{Bai2022}, \cite{Xu2021a} \\ \hline
    
    Graph & 5 & \cite{Ribeiro2021b}, \cite{Ke2021}, \cite{Uehara2020}, \cite{Teixeira2020}, \cite{laha-etal-2019} \\ \hline
    
    SQL & 2 & \cite{Xu2018}, \cite{Nan2021} \\ \hline
    JSON & 1 & \cite{laha-etal-2019} \\ \hline
    MRS & 1 & \cite{Hajdik2019} \\ \hline

\end{tabular}%
} \caption{Data type frequency.} 
    \label{tab:datatypes}
\end{table}



\subsubsection{Overview of Datasets}

Here is a brief overview of some prevalent datasets mentioned in the previous section:

\textbf{WebNLG} The WebNLG dataset \cite{Gardent2017}, introduced in 2017, is a pivotal resource used in the WebNLG 2017 challenge. It contains RDF triples from DBPedia, each paired with text descriptions and human-generated reference texts in English. This dataset serves as a critical tool for training and evaluating the planner component. It comprises 9,674 unique triple sets and 25,298 text references, divided into training, development, and test sets. The test set includes both seen and unseen domains, allowing for the evaluation of model generalizability. Additionally, the WebNLG 2020 dataset \cite{castro-ferreira-etal-2020} expands on this resource with Russian data. This addition was achieved through translation and post-editing, aimed at fostering multilingual capabilities and providing essential statistics for evaluating Natural Language Generation systems in the Semantic Web domain. 

\textbf{E2E} The E2E \cite{novikova2017e2e} dataset is a crucial resource for training end-to-end, data-driven natural language generation systems, specifically in the restaurant domain. It was gathered through crowdsourcing and meticulous quality control, using images as stimuli to evoke more natural and well-articulated human references than textual MRs. This dataset, openly released as part of the E2E NLG challenge, is approximately ten times larger than its predecessors. It introduces novel challenges due to its size, lexical diversity, syntactic intricacies, and discourse complexities. Learning from this dataset promises to generate more natural, diverse, and less template-like system utterances. It comprises a rich set of 50,602 English verbalizations paired with 5,751 dialogue-act-based meaning representations. The dataset is thoughtfully partitioned into training, validation, and testing subsets, maintaining a consistent distribution of MR and reference text lengths while ensuring MR variation across different sets. Each MR encompasses 3–8 attributes or slots, including information such as name, food, area, and corresponding values. 

\textbf{AMR} The Abstract Meaning Representation (AMR) \cite{banarescu2013abstract} dataset\footnote{https://catalog.ldc.upenn.edu/byyear}, encompassing series such as LDC2011T07, LDC2015E86, LDC2016E25, LDC2017T10, LDC2020T02, and LDC2020T07, presents a structured representation of semantic information. AMR is represented as a rooted, directed, acyclic graph with labelled edges (relations) and nodes (concepts), capturing the essence of \say{who is doing what to whom}. It serves as a foundation for generating sentences that convey the semantics encoded within the graph. Specifically, LDC2017T10 within this series comprises 36,521 training instances of AMR graphs in PENMAN notation \cite{goodman2020penman}, along with their corresponding texts. Additionally, it includes 1,368 development instances and 1,371 test instances, providing a substantial resource for research and development in AMR parsing and natural language understanding and generation. 

\textbf{WikiBio} The WikiBio dataset \cite{Lebret2016} is a comprehensive collection of biographical information covering individuals from diverse professions. This extensive dataset comprises over 10 million records, providing details such as names, dates of birth, occupations, and education. Focused on curating 728,321 biographies sourced from English Wikipedia, it serves as a valuable resource for evaluating text generation algorithms. The dataset encompasses the initial paragraph of each biography alongside its associated infobox, both tokenized to facilitate processing. Arranged in a standardized tabular layout, it proves to be of great utility for a range of natural language processing tasks, including text generation, summarization, entity identification, and data extraction. Researchers can utilize this dataset for the development and evaluation of algorithms and models geared toward the analysis of biographical text. 

\textbf{RotoWire} The RotoWire dataset \cite{Wiseman2017}, designed for table-to-text generation, offers a robust platform for generating human-like summaries from basketball game tables. With 4.9K examples and 1.6M tokens from rotowire.com, it includes game data and corresponding human-written summaries. RotoWire features extended texts and a diverse vocabulary, making content selection more challenging. It leverages table records to generate structured yet informal game summaries, appealing to those interested in game statistics. This dataset proves valuable for assessing data-to-document generation systems, especially in basketball game summaries. It presents the complex task of converting structured data into coherent, informative text, accommodating diverse audiences and writing styles. Researchers can employ it to develop and evaluate table-to-text generation models, advancing natural language generation in specific domains.

\textbf{ToTTo} The ToTTo dataset \cite{Parikh2020}, is a valuable resource in table-to-text generation within an open-domain English context. It comprises a substantial training set of over 120,000 instances. Its primary objective is generating a controlled one-sentence description task based on the information in a Wikipedia table, mainly focusing on the highlighted table cells. The dataset's creation involved a meticulous process wherein noisy descriptions were meticulously paired with tables, and any inaccuracies or inconsistencies in the highlighted cells were diligently rectified through iterative refinement. ToTTo stands as a pivotal benchmark for advancing high-precision, faithful, and conditional text generation research.

\textbf{WMT} Since its inception in 2006, the Workshop on Machine Translation (WMT) has been a hub for machine translation shared tasks and competitions. Its impact extends to other fields of natural language generation, such as in multilingual data-to-text generation \cite{Song2019, Freitag2018, Agarwal2020}. Initially centered around translation tasks, WMT has evolved to encompass various aspects, including biomedical, multimodal, and low-resource translation. The General primary Machine Translation task, previously known as the News Task, remains a core component. Additionally, WMT features evaluation tasks like Metrics and Quality estimation. Over the years, some tasks have been discontinued, but WMT's shared task results and datasets remain crucial benchmarks for advancing machine translation research. The WMT dataset is sourced from the OPUS corpus \cite{tiedemann2012parallel}, and it consists of a parallel corpus in 18 languages, primarily from the news domain. It includes news commentary text extracted mainly from online news sources, with the test data containing about 1000 sentence pairs. While most languages are paired with English, some are also paired with languages other than English such as German, Russian, French, Spanish, Italian and Chinese. The competition offers training sets compiled from various sources \cite{Bojar2017, Barrault2019}.

\textbf{Viggo} The ViGGO dataset \cite{Juraska2019} addresses limitations in existing data-to-text NLG corpora by focusing on video game descriptions and providing a more conversational context. It includes over 100 video game titles and their attributes, resulting in 2,300 structured meaning representations (MRs). These MRs cover nine different dialogue act types (DAs), making ViGGO suitable for open-domain dialogue systems. Crowdsourced reference utterances were collected for each MR, enabling neural language generation models to learn multiple ways of expressing the same content. While smaller in size compared to the E2E dataset, ViGGO offers greater lexical diversity, longer inform utterances, and a more natural-sounding context due to its grounding in real video game data. Additionally, ViGGO maintains a focus on shorter, conversational responses.



\subsection{Language}
\label{sec:lang}

Multilingualism has emerged as a crucial aspect of natural language generation within data-to-text field, showcasing the remarkable progress achieved in handling structured data across various languages. Data-to-text generation has faced numerous challenges when it comes to multilingual outputs, with English often dominating as the primary target language for generation tasks. In our analysis of the 90 selected papers, a striking trend is evident: 89 of them focused on generating summaries from structured data in English. While English takes the lead, other languages also make appearances, with German, Russian, French, Spanish, Italian, and Brazilian Portuguese featuring in 6, 4, 3, 2, 2, and 2 papers, respectively. A variety of other languages were utilized in just one paper each, as illustrated in Table \ref{tab:langs}.

\begin{figure}[ht]
    \centering
    \includegraphics[width=0.45\linewidth]{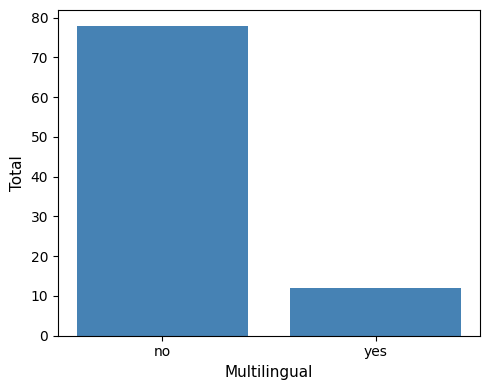}
    \caption{Multilinguality in Data-to-text Generation.}
    \label{fig:multi}
\end{figure}

\begin{table}[H]
\centering
\scriptsize
\resizebox{\columnwidth}{!}{%
\begin{tabular}{|p{0.45\textwidth}|l|p{0.5\textwidth}|}
\hline
\textbf{Language} & \textbf{Count} & \textbf{Citations} \\ \hline

English & 77 & \cite{Wiseman2017}, \cite{Konstas2017}, \cite{Gardent2017},   \cite{Song2018}, \cite{Wiseman2018}, \cite{Gehrmann2018},   \cite{Xu2018}, \cite{Puzikov2018}, \cite{Nie2018}, \cite{Shen2020}, \cite{Ribeiro2020}, \cite{Bai2020}, \cite{Zhang2020},   \cite{Fu2020}, \cite{Kedzie2020}, \cite{Bai2022}, \cite{Freitag2018}, \cite{Perez-Beltrachini2018}, \cite{Shimorina2018}, \cite{VanderLee2018}, \cite{Puduppully2019}, \cite{Moryossef2019}, \cite{Ferreira2019}, \cite{Dhingra},   \cite{Puduppully2020}, \cite{Damonte2019}, \cite{Jiang2020}, \cite{Gong2020a}, \cite{Chen2020b}, \cite{Ram2020}, \cite{Wang2020}, \cite{Harkous2020}, \cite{Nie2020}, \cite{Ribeiro2019}, \cite{Gong2019}, \cite{Iso2020}, \cite{Kedzie2019}, \cite{Chen2019}, \cite{Hajdik2019}, \cite{Wang2019}, \cite{Qader2019}, \cite{Juraska2019}, \cite{Chen2020}, \cite{Fillippova2020}, \cite{laha-etal-2019}, \cite{Parikh2020}, \cite{Kale2020}, \cite{Chang2021a},\cite{Uehara2020},   \cite{Iso2020a}, \cite{Lin2020}, \cite{Guo2020}, \cite{Arun2020},   \cite{Gong2020} ', \cite{Rebuffel2020}, \cite{Ribeiro2021b}, \cite{Nan2021}, \cite{Chang2021}, \cite{Ribeiro2021}, \cite{puduppully-lapata-2021},   \cite{Ke2021}, \cite{Opitz2021}, \cite{Chen2020a}, \cite{Zhao2020}, \cite{Suadaa2021}, \cite{Liu2021}, \cite{Wang2021},   \cite{Su2021}, \cite{Juraska2021}, \cite{Xu2021},   \cite{Hargreaves2021}, \cite{Wiseman2021}, \cite{Li2021},   \cite{Heidari2021}, \cite{Rebuffel2022}, \cite{Hardy2018}, \cite{Wang2020a} \\ \hline
English, Russian & 3 & \cite{castro-ferreira-etal-2020}, \cite{Agarwal2020},   \cite{Li2020} \\ \hline
English, German & 2 & \cite{Song2019}, \cite{Lu2022} \\ \hline
Brazilian Portuguese & 1 & \cite{Moussallem2019} \\ \hline
English, Chinese & 1 & \cite{Shao2019} \\ \hline
English, French, German & 1 & \cite{Nema2018} \\ \hline

English, Spanish, Italian, German, Danish, Greek, Finnish, French, Portuguese, Swedish, Bulgarian, Czech, Estonian, Hungarian, Latvian, Romanian, Slovak, Slovenian, Lithuanian, Dutch, Polish  & 1 & \cite{Fan2020} \\  \hline

German, Russian, English & 1 & \cite{Moussallem2020} \\ \hline
Brazilian Portuguese, English & 1 & \cite{Teixeira2020} \\ \hline
English, French & 1 & \cite{Garneau2021} \\ \hline
English, Spanish, German, Italian & 1 & \cite{Xu2021a} \\ \hline

\end{tabular}%
} \caption{Languages and Multilingualism} 
\label{tab:langs}
\end{table}


In our exploration of multilingualism and its role in data-to-text literature, we assessed 90 papers. Among them, 12 studies ventured into multilingual approaches, while the remaining 78 predominantly focused on single-language generation, as depicted in Figure \ref{fig:multi}. Notably, a noteworthy study by \citet{Fan2020} introduced a multilingual method for generating text from AMRs across twenty-one (21) EUROPARL languages. Similarly, researchers such as \citet{Xu2021a}, \citet{Nema2018}, and \citet{Moussallem2020} have pursued analogous approaches, generating text concurrently in multiple European languages, underscoring the heightened interest and capabilities in multilingual data-to-text generation.
Conversely, some studies, such as those by \citet{Song2019}, \citet{Shao2019}, \citet{castro-ferreira-etal-2020}, \citet{Garneau2021}, \citet{Agarwal2020}, \citet{Lu2022}, \citet{Li2020}, and \citet{Teixeira2020}, have centered their efforts on bilingual text generation within the data-to-text context. 
Significant advancements have indeed been achieved in the domain of multilingual data-to-text generation, particularly in English, Chinese, and select European languages. Nonetheless, it is apparent that additional research is essential to advance this field. This research is needed to enable proficient multilingual generation across a more diverse range of languages, including those with diverse morphological structures and word order characteristics, as emphasized in \citet{Fan2020} work.

\subsection{Models}
\label{sec:models}

\begin{table}[H]
    \setlength{\tabcolsep}{4pt}
    \centering
    \scriptsize
    \setlength\tabcolsep{2pt}
    \begin{tabular}{|p{0.3\textwidth}|l|p{0.6\textwidth}|}
    \hline
    \textbf{Methodology} & \textbf{Counts} & \textbf{Papers}  \\ \hline
    
    Copy Mechanism & 27 & \cite{Wiseman2017}, \cite{Song2018}, \cite{Wiseman2018}, \cite{Gehrmann2018}, \cite{Nie2018}, \cite{Shimorina2018}, \cite{Puduppully2019}, \cite{Moryossef2019}, \cite{Puduppully2020}, \cite{Ribeiro2019}, \cite{Gong2019}, \cite{Chen2019}, \cite{Hajdik2019}, \cite{Wang2019}, \cite{Chen2020a}, \cite{Zhao2020}, \cite{Chen2020}, \cite{Lin2020}, \cite{Rebuffel2020}, \cite{Chen2020b}, \cite{Wang2020}, \cite{Wang2020a}, \cite{Jiang2020}, \cite{puduppully-lapata-2021}, \cite{Suadaa2021}, \cite{Su2021}, \cite{Wiseman2021} \\ \hline
    GNN & 18 & \cite{Song2018}, \cite{Xu2018}, \cite{Song2019}, \cite{Damonte2019}, \cite{Ribeiro2019}, \cite{Chen2020b}, \cite{Wang2020}, \cite{Ribeiro2020}, \cite{Bai2020}, \cite{Zhang2020}, \cite{Moussallem2020}, \cite{Guo2020}, \cite{Ribeiro2021}, \cite{Bai2022}, \cite{Zhao2020}, \cite{Chen2020}, \cite{Li2021}, \cite{Ke2021} \\ \hline
    Hierarchical encoder & 5 & \cite{Li2021}, \cite{Gong2019}, \cite{Chen2020b}, \cite{Rebuffel2020}, \cite{Li2021} \\ \hline
    HMMM & 2 & \cite{Wiseman2018},\cite{Xu2021} \\ \hline
    Transformer & 38 & \cite{Chen2020b}, \cite{Moussallem2020}, \cite{Ribeiro2021}, \cite{Gehrmann2018}, \cite{Juraska2019}, \cite{Parikh2020}, \cite{Kale2020}, \cite{Chen2020a}, \cite{Ram2020}, \cite{Harkous2020}, \cite{Wang2021}, \cite{Su2021}, \cite{Juraska2021}, \cite{Bai2022}, \cite{Garneau2021}, \cite{Chang2021a}, \cite{Chang2021}, \cite{Rebuffel2020}, \cite{Ribeiro2021b}, \cite{Wang2020a}, \cite{Gong2020a}, \cite{Chen2020}, \cite{Fan2020}, \cite{Agarwal2020}, \cite{Fu2020}, \cite{Kedzie2020}, \cite{Li2020}, \cite{Lin2020}, \cite{Arun2020}, \cite{Xu2021}, \cite{Wiseman2021}, \cite{Heidari2021}, \cite{Lu2022}, , \cite{Suadaa2021}, \cite{Liu2021}, \cite{Ke2021}, \cite{Nan2021}, \cite{Opitz2021} \\ \hline
    \end{tabular}
    \caption{Methodology Used in the Selected Papers}
    \label{tab:model}
\end{table}

In the field of data-to-text generation, various methodologies have been explored, including templates, statistical models, and neural network-based generation models. A significant number of studies selected for this systematic review opted for neural network models, ranging from basic sequence-to-sequence models \cite{bahdanau2014neural} to advanced designs such as transformers \cite{vaswani2017attention} and large pretrained language models.

Several studies have introduced refinements to these models to enhance their performance and set new benchmarks. Notably, research conducted by 27 studies, integrated a copy mechanism that uses probabilistic strategies to determine when and which tokens should be directly copied from the reference data. This copy model is particularly useful for ensuring that all data values appear in the generated text \cite{Gehrmann2018}. Building on this innovation, subsequent investigations by five (5) studies devised fused attention \cite{Nema2018} and hierarchical attention \cite{Puduppully2020}, improving decoder focus on input tokens and enhancing overall generation quality. However, some of these studies employed recurrent neural networks with static embeddings, which have shown suboptimal performance across various application domains such as in AMR-to-text \cite{Song2018, Hardy2018}, MR-to-text \cite{Nie2020}, table-to-text \cite{Nema2018} and in data-to-text \cite{Chen2020b}.

Another line of research has focused on retaining the graph structure of the data using graph encoders \cite{Song2018} in graph neural network (GNN). Originally, graph data were linearized into sequences to accommodate sequence-to-sequence models. However, graph encoding has emerged as a promising method in data-to-text generation, especially in AMR-to-text generation. Graph encoders offer improvements over basic sequence-to-sequence models by inherently learning the graph structure and existing relations in the encoder and using an ordinary decoder to generate desired texts. \citet{Ribeiro2019} leveraged dual graph representations in AMR-to-text generation, effectively encoding divergent but complementary perspectives of the structural information in the AMR graph by simultaneously learning top-down and bottom-up node representations \cite{Ribeiro2019}. The study by \citet{Zhao2020} introduced DUALENC, a dual encoding model that addresses the structural gap in data-to-text generation by incorporating both graph and linear structures. This approach significantly improves text quality compared to single-encoder models, especially for structured inputs like trees or graphs. The DUALENC\cite{Zhao2020} model integrates Graph Convolutional Network (GCN) encoders with an intermediate content planning stage. This combination allows the model to capture structural information and enhance the compatibility between input and output sequences \cite{Zhao2020}. Another research \citet{Li2021} introduces a hierarchical encoder equipped with a reasoning module for graph-based reasoning, which enhances the ability to capture various relations between records in different dimensions. This research also introduces auxiliary supervision tasks, including number ranking and importance ranking, to further improve the model's ability to handle different record relations \cite{Li2021}. These advancements contribute significantly to the field of GNNs. Eighteen (18) papers were found to have used this method to improve on existing baselines.

To address the limitations of static embedding models, some research efforts have embraced transformer models such as RNN transformers \cite{hochreiter1997long, cho2014properties}, BERT\cite{vaswani2017attention}, T5\cite{raffel2020exploring}, BART\cite{lewis2019bart}, XLM \cite{lample2019cross}, and GPT-2\cite{radford2019language} for data-to-text generation tasks. These transformer-based models incorporate contextualized embeddings and use positional encoding due to their non-recurrent nature. They are trained on a large corpus of online curated texts and seem to perform well across several domains after fine-tuning. A total of 38 studies incorporated transformer models into their research, as detailed in Table \ref{tab:model}.


\subsection{Hallucination Mitigation Measures}
\label{sec:hal}

In the context of data-to-text, hallucination refers to the generation of content that lacks fidelity or is not supported by the source data provided \cite{Dhingra}. Divergence can also be considered a form of hallucination when it occurs in the reference text, signifying a deviation from the expected or accurate information \cite{Dhingra}. This highlights the importance of generating text that remains faithful to the underlying data and references.

To address and reduce errors and hallucinations in data-to-text generation, several strategies have been deployed. These strategies encompass a wide range of approaches, including dataset cleaning and standardization, the development of novel training modules and techniques, as well as the application of knowledge distillation methods \cite{Rebuffel2020}.

In the subsequent subsections, we will delve into specific papers and elucidate the strategies they have employed to address challenges and improve data-to-text generation. This comprehensive exploration will provide valuable insights into the diverse approaches and techniques used in the field to enhance the quality and fidelity of generated text.

\subsubsection{\textbf{Dataset Refinement and Post Editing}}

Effective refinement of datasets and post-editing play an essential role in enhancing the quality and accuracy of data-to-text generation. \citet{Wang2019} made a significant contribution to this field by focusing on boosting factual accuracy. Their work involves developing Rotowire-FG, which stands for Fact Grounding-purified version of Rotowire \cite{Wiseman2017}. Within this context, they incorporated content normalization techniques aimed at boosting the overall accuracy of the generated text. These techniques involve actions like converting number words into their corresponding numerical values and standardizing mentions of entities, contributing to improved text fidelity and reliability \cite{Wang2019}.

In addition, the TOTTO dataset\cite{Parikh2020} facilitates the controlled generation of concise descriptions derived from Wikipedia tables. This initiative aimed to enhance controllability in data-to-text generation, addressing issues previously associated with crowd-sourced datasets, which is the incomplete alignment of information in the table and their corresponding summaries. \citet{Chen2020} addresses logical-level generation, presenting the LOGIC2TEXT dataset designed for generating high-fidelity descriptions from logical forms. To address scientific data-to-text challenges, \citet{Suadaa2021} emphasize numerical reasoning in textual descriptions, as evidenced by the creation and utilization of the numericNLG dataset \cite{Suadaa2021}. Collectively, these efforts improve data-to-text generation, enhancing text accuracy and coherence.

Furthermore, \citet{Shimorina2018} focuses on handling rare items or entities in the generated text. Their approach involves post-processing the text to replace placeholders with the appropriate values based on a mapping between placeholders and initial values created during pre-processing. This strategy further enhances the fidelity of the generated text.

\subsubsection{\textbf{Training Techniques and Model Modification}}

In this field, \citet{Gong2019} presents an innovative architecture, a hierarchical encoder for table-to-text generation that excels at encapsulating the intricacies of multi-dimensional table data. The model’s ability to encode row, column, and time dimensions simultaneously enables it to generate text summaries that are both highly informative and coherent. A study \citet{Ribeiro2021}, introduces STRUCTADAPT, which leverages adapter modules to incorporate graph structure into pretrained language models (PLMs) for better Abstract Meaning Representation (AMR) to text generation, resulting in enhanced performance compared to prior approaches \cite{Ribeiro2021}.

Neural generation models use various strategies to reduce hallucination, including soft templates \cite{Gehrmann2018}, copy mechanisms, content planning, and structure-aware systems \cite{Chen2020b}. Training methodologies have evolved significantly, with some studies using paired training with unlabeled text \cite{Konstas2017}, and others implementing a two-tiered approach involving an information extraction model and an attention-based encoder-decoder text generation model \cite{Wiseman2017}. The JointGT model by \citet{Ke2021} enhances Knowledge Graph to text generation tasks by leveraging graph structure and pre-training tasks. 

On a different note, (2) papers delved into Reinforcement Learning (RL) for data-to-text generation. \citet{Rebuffel2020} introduces PARENTing, a reinforcement learning framework that is model-agnostic. This framework fine-tunes pretrained models using self-critical policy gradient algorithms to minimize hallucination in text generation by addressing divergence in training examples. Another study used multi-task learning and reinforcement learning to incorporate content selection mechanisms into the encoder-decoder models \cite{Perez-Beltrachini2018}.

\subsubsection{\textbf{Controllabilty and Constraints decoding}}

Controllability is achieved in data-to-text generation by introducing constraints and supervision during the decoding process or in the decoder. Various studies have contributed to this field: 
\begin{itemize} 
    \item \citet{Lin2020} developed a novel neural model for data-to-text generation with style imitation to follow a certain style of writing from examples. The model employs a hybrid attention-copy mechanism and weak supervisions, using a content coverage constraint for balanced content fidelity and style control, proving effective in controlled text generation tasks \cite{Lin2020}. 
    \item \citet{Wang2020a} presents a Transformer-based framework for table-to-text generation with a focus on producing faithful and informative text descriptions aligned with input tables. It introduces two essential strategies, a Table-Text Disagreement Constraint Loss and Constrained Content Matching via Optimal Transport, along with a novel evaluation metric, PARENT-T, to measure faithfulness in generated text. These constraints ensure that the latent representation of the table aligns with the corresponding representation of the generated text \cite{Wang2020a}.
    \item \citet{Shen2020} segments target text into fragments that align with data records, improving the control and interpretability of the generated output. This automatic segmentation, which adapts to domain-specific requirements, employs a soft statistical constraint to regularize the granularity of the segments \cite{Shen2020}.
    \item \citet{Wang2021} introduces Mention Flags (MF), a unique method that guarantees constraint satisfaction in Transformer-based text generation. By tracking the fulfilment of lexical constraints in the generated text and integrating them into the S2S Transformer models, MF ensures the creation of high-quality text that complies with the given constraints \cite{Wang2021}.
    \item \citet{Lu2022} presents NEUROLOGIC Aesque, an innovative decoding algorithm, inspired by A* search and designed for large-scale language models, that enables constrained text generation. This is achieved by combining heuristic cost estimates and logic-based lexical constraints, enhancing Constrained Machine Translation and Keyword-constrained generation \cite{Lu2022}.
    \item \citet{Hardy2018}  introduces a new method to improve AMR-based summarization by guiding it with the source document. This two-step process estimates the distribution of missing linguistic data and uses it to guide a seq2seq model, enhancing summary fluency and quality.
\end{itemize}

\subsubsection{\textbf{Ranking System}}

Eight (8) papers utilized rankers to improve the fidelity of the generated text \cite{Hargreaves2021, Juraska2021, Garneau2021, Moryossef2019, Li2020, laha-etal-2019, Zhao2020, Li2021}. The process of reranking in text generation, specifically within the decoder, is aimed at enhancing the quality and reducing semantic errors in the generated text. This involves the creation of rules or the use of an auxiliary classifier to verify if input slots are represented in the output, an important factor in maintaining semantic quality. Rerankers are commonly applied to the final hypotheses to enhance beam search and address its limitations. This can be achieved by establishing a reranking criterion or training a reranker to predict the best hypothesis within a beam based on function scores \cite{Hargreaves2021, Moryossef2019, laha-etal-2019}.

The strategy employed involves over-generation, followed by reranking of potential outputs using criteria that were not explicitly optimized during training. The reranked outputs favour those with fewer missing or incorrect slot mentions, thereby enhancing accuracy and relevance \cite{Juraska2021, Hargreaves2021}. This approach extracts meaningful information from encoder-decoder models and uses it to identify which attributes are mentioned in the generated text.

Additionally, A promising development in neural data-to-text generation is the introduction of a trainable evaluation metric. This metric, particularly useful when tables have multiple associated textual references, uses ranking models to assess the correctness of generated hypotheses by comparing them to the original table and corresponding references. It aims to overcome the limitations of existing metrics like BLEU, ROUGE, and METEOR, which do not fully capture the faithfulness of generated text to both the input table and references \cite{Garneau2021}.

\subsubsection{\textbf{Pipeline and Planning Architecture Systems}}

In a survey of Natural Language Generation (NLG) architectures and methodologies, \citet{Gatt2018a} categorizes NLG approaches into three main architectural paradigms: Modular, Planning, and Integrated (Global) architectures. These architectures encompass various generation systems, which are classified based on their methodological approach and design choices \cite{Gatt2018a}. To address challenges such as hallucination, omissions, and errors encountered in data-to-text generation, eleven (11) papers in total made use of this modular architecture.

Three (3) of these studies \cite{VanderLee2018, Moussallem2019, Teixeira2020} resorted to traditional data-to-text methods. These methods typically involve a sequential process encompassing discourse ordering, text structuring, lexicalization, referring expression generation, and textual realization. These stages can be further categorized into macro planners, which combine content selection and document planning, and micro planners, which involve sentence aggregation, lexicalization, and referring expression generation \cite{Gatt2018a}.

In contrast, recent studies have witnessed a departure from traditional rule-based approaches in the initial planning stages. Instead, five (5) studies have adopted end-to-end models for generating text, spanning from intermediate stages to the final surface realization. This transition aims to assess the efficacy of end-to-end models when compared to conventional rule-based or template systems  \cite{Gardent2017, laha-etal-2019, Nie2020, castro-ferreira-etal-2020, puduppully-lapata-2021}. This shift underscores the dynamic evolution in NLG methodologies and architectural preferences.

Furthermore, four (5) investigations \cite{Moryossef2019, Xu2021, Jiang2020, Ferreira2019, Zhao2020} have sought to compare the performance of neural modular architectures against end-to-end neural architectures. Collectively, these research findings indicate that supervised neural modularization or pipelining within data-to-text architectures leads to notable improvements in fluency, fidelity, and the overall quality of generated text summaries. These enhancements primarily result from error reduction during content selection, the model's ability to capture long-term structural dependencies, and the accurate ordering of facts \cite{Wiseman2017, Ferreira2019}.

\citet{Shao2019} introduces the Planning-based Hierarchical Variational Model (PHVM) to address the limitations of existing neural methods in generating long and diverse texts in data-to-text generation tasks. The PHVM incorporates a planning mechanism and a hierarchical latent structure to capture inter-sentence coherence and generate varied expressions. By decomposing long text generation into dependent sentence generation sub-tasks, the model effectively models input data dynamically during generation \cite{Shao2019}.

\subsection{Evaluation Metric}
\label{sec:eval}

In data-to-text generation, assessing the quality and suitability of generated text has relied on various metrics over the years. These assessments can be broadly categorized into two groups: automatic evaluation and human evaluation. Automatic evaluation employs computational methods to measure text quality, while human evaluation enlists human participants to capture nuanced aspects of text quality and coherence.

\subsubsection{\textbf{Automatic Evaluation}}

\paragraph{\textbf{N-gram Metrics}} Our analysis indicates that the most commonly employed automatic metric is BLEU \cite{papineni2002bleu}, with a substantial presence in 80 papers. It is closely followed by METEOR \cite{denkowski2014meteor}, which is employed in 40 papers, demonstrating its continued relevance. ROUGE \cite{lin2004rouge}, with 17 papers utilizing it, has also been a consistent choice for assessing generated text quality. Furthermore, CHrF++ \cite{popovic2015chrf}, used in 10 papers, and NIST \cite{doddington2002automatic}, applied in 9 papers, have offered valuable insights into the evaluation of data-to-text outputs. These metrics, while non-semantic in nature, have played a role in understanding word or character count and n-gram overlap between generated text and reference texts. 

In seven studies, TER \cite{snover-etal-2006-study} was employed as an edit distance metric to evaluate machine translation, quantifying the human-level editing required to align system output with a reference \cite{snover-etal-2006-study}. Additionally, fifteen studies utilized CIDEr \cite{vedantam2015cider}, a widely adopted metric for image captioning, which assesses the similarity between generated and reference captions, considering both linguistic and content aspects, and applying TF-IDF-based n-gram weighting \cite{sai2022survey}. Furthermore, two studies incorporated SPICE \cite{anderson2016spice} metrics, which calculates the semantic propositional content overlap between generated and reference captions using scene graphs. Four (4) studies incorporated the use of the SER (Slot-Error Rate) metrics, which are appropriate for assessing the presence of named entities, with SER being computed through exact matching of slot values in the candidate texts \cite{Shimorina2018}. 

\begin{table}[H]
\centering
\scriptsize
\resizebox{\columnwidth}{!}{%
\begin{tabular}{|l|l|p{0.7\textwidth}|}
\hline
\textbf{Metric} & \textbf{Count} & \textbf{Papers} \\ \hline

    BLEU & 80 & \cite{Wiseman2017}, \cite{Konstas2017}, \cite{Gardent2017}, \cite{Song2018}, \cite{Wiseman2018}, \cite{Gehrmann2018}, \cite{Xu2018}, \cite{Puzikov2018}, \cite{Nie2018}, \cite{Hardy2018}, \cite{Shimorina2018}, \cite{VanderLee2018}, \cite{Lu2022}, \cite{Puduppully2019}, \cite{Moryossef2019}, \cite{Ferreira2019}, \cite{Song2019}, \cite{Dhingra}, \cite{Puduppully2020},  \cite{Damonte2019}, \cite{Shao2019}, \cite{Nie2020}, \cite{Ribeiro2019}, \cite{Gong2019}, \cite{Iso2020}, \cite{Kedzie2019}, \cite{Chen2019}, \cite{Hajdik2019}, \cite{Wang2019}, \cite{Qader2019}, \cite{Juraska2019}, \cite{laha-etal-2019}, \cite{Parikh2020}, \cite{Kale2020}, \cite{Chen2020a}, \cite{Zhao2020}, \cite{Chen2020b}, \cite{Ram2020}, \cite{castro-ferreira-etal-2020}, \cite{Wang2020}, \cite{Harkous2020}, \cite{Wang2020a}, \cite{Shen2020}, \cite{Ribeiro2020}, \cite{Jiang2020}, \cite{Nema2018}, \cite{Gong2020a}, \cite{Chen2020}, \cite{Fillippova2020}, \cite{Fan2020}, \cite{Bai2020}, \cite{Fu2020}, \cite{Kedzie2020}, \cite{Bai2022}, \cite{Uehara2020}, \cite{Iso2020a}, \cite{Moussallem2020}, \cite{Li2020}, \cite{Lin2020}, \cite{Guo2020}, \cite{Arun2020}, \cite{Gong2020}, \cite{Rebuffel2020}, \cite{Ribeiro2021b}, \cite{Nan2021}, \cite{Chang2021}, \cite{puduppully-lapata-2021}, \cite{Ke2021}, \cite{Opitz2021}, \cite{Suadaa2021}, \cite{Liu2021}, \cite{Wang2021}, \cite{Su2021}, \cite{Juraska2021}, \cite{Xu2021}, \cite{Hargreaves2021}, \cite{Xu2021a}, \cite{Wiseman2021}, \cite{Li2021}, \cite{Rebuffel2022}, \cite{Lu2022}  \\ \hline
    METEOR & 40 & \cite{Gardent2017}, \cite{Wiseman2018}, \cite{Puzikov2018}, \cite{Shimorina2018}, \cite{Moryossef2019}, \cite{Ferreira2019}, \cite{Song2019}, \cite{Dhingra}, \cite{Damonte2019}, \cite{Ribeiro2019}, \cite{Kedzie2019}, \cite{Qader2019}, \cite{Juraska2019}, \cite{laha-etal-2019}, \cite{Kale2020}, \cite{Zhao2020}, \cite{Chen2020b}, \cite{Ram2020}, \cite{castro-ferreira-etal-2020}, \cite{Harkous2020}, \cite{Wang2020a}, \cite{Shen2020}, \cite{Ribeiro2020}, \cite{Bai2020},   \cite{Fu2020}, \cite{Moussallem2020}, \cite{Li2020}, \cite{Guo2020}, \cite{Ribeiro2021b}, \cite{Nan2021}, \cite{Chang2021}, \cite{Ke2021}, \cite{Opitz2021}, \cite{Suadaa2021}, \cite{Wang2021}, \cite{Juraska2021}, \cite{Xu2021}, \cite{Wiseman2021}, \cite{Lu2022}, \cite{Bai2022} \\ \hline
    ROUGE & 17 & \cite{Wiseman2018}, \cite{Gehrmann2018}, \cite{Hardy2018},   \cite{Juraska2019}, \cite{Wang2020a}, \cite{Shen2020}, \cite{Nema2018},   \cite{Gong2020a}, \cite{Chen2020}, \cite{Fu2020}, \cite{Chang2021},   \cite{Ke2021}, \cite{Su2021}, \cite{Juraska2021}, \cite{Xu2021},   \cite{Wiseman2021}, \cite{Lu2022} \\ \hline
    CIDER & 15 & \cite{Wiseman2018}, \cite{Gehrmann2018}, \cite{Puzikov2018},   \cite{Moryossef2019}, \cite{Dhingra}, \cite{Juraska2019}, \cite{Harkous2020}, \cite{Shen2020}, \cite{Fu2020}, \cite{Kedzie2020}, \cite{Wang2021}, \cite{Juraska2021}, \cite{Xu2021}, \cite{Wiseman2021}, \cite{Lu2022} \\ \hline
    CHRF++ & 10 & \cite{Ram2020}, \cite{castro-ferreira-etal-2020}, \cite{Wang2020}, \cite{Ribeiro2020}, \cite{Moussallem2020}, \cite{Li2020}, \cite{Guo2020}, \cite{Ribeiro2021b}, \cite{Opitz2021}, \cite{Bai2022} \\ \hline
    NIST & 9 & \cite{Wiseman2018}, \cite{Puzikov2018}, \cite{Shimorina2018},   \cite{Nema2018}, \cite{Fu2020}, \cite{Kedzie2020}, \cite{Chang2021},   \cite{Wang2021}, \cite{Wiseman2021} \\ \hline
    RG, CO, CS & 11 & \cite{Wiseman2017}, \cite{Puduppully2019}, \cite{Dhingra}, \cite{Puduppully2020}, \cite{Gong2019}, \cite{Iso2020}, \cite{Wang2019}, \cite{Jiang2020}, \cite{Gong2020} , \cite{puduppully-lapata-2021},   \cite{Li2021} \\ \hline
    PARENT & 9 & \cite{Dhingra}, \cite{Kale2020}, \cite{Wang2020a}, \cite{Fillippova2020}, \cite{Rebuffel2020}, \cite{Suadaa2021}, \cite{Garneau2021}, \cite{Liu2021}, \cite{Rebuffel2022} \\ \hline
    PARENT-T & 1 & \cite{Wang2020a} \\ \hline
    SER & 4 & \cite{Shimorina2018}, \cite{Juraska2019}, \cite{Juraska2021},   \cite{Xu2021} \\ \hline
    SPICE & 2 & \cite{Wang2021}, \cite{Lu2022} \\ \hline
    TER & 7 & \cite{Gardent2017}, \cite{Song2019}, \cite{Zhao2020}, \cite{Li2020}, \cite{Guo2020}, \cite{Nan2021}, \cite{Xu2021} \\ \hline
    BERTScore & 7 & \cite{castro-ferreira-etal-2020}, \cite{Li2020}, \cite{Guo2020}, \cite{Ribeiro2021b}, \cite{Nan2021}, \cite{Opitz2021}, \cite{Suadaa2021} \\ \hline
    BLEURT & 5 & \cite{castro-ferreira-etal-2020}, \cite{Li2020}, \cite{Guo2020}, \cite{Ribeiro2021b}, \cite{Nan2021} \\ \hline
    MOVERScore & 2 & \cite{Ribeiro2021b}, \cite{Nan2021} \\ \hline
\end{tabular}%
} \caption{Table of the Evaluation Metrics used across the Studies.}
    \label{tab:metric-table}
\end{table}


\paragraph{\textbf{Task Specific Metrics}} A significant debate persists among researchers regarding the appropriateness of these non-semantic n-gram metrics for evaluating data-to-text outputs. These metrics primarily rely on word count and n-gram overlap between the generated text and reference texts, often showing limited correlation with human judgment \cite{Dhingra}. They were originally developed and applied in other natural language generation (NLG) domains, such as translation for BLEU, NIST and METEOR and summarization for ROUGE.
In response to the unique challenges posed by table-to-text generation, task-specific metrics like PARENT \cite{Dhingra} and its variant, PARENT-T \cite{Wang2020a}, have emerged. These metrics assess the quality of table-to-text outputs by comparing the generated information with the entries in the source table. The research also analyzes the sensitivity of the metrics to divergence by collecting labels for cases where references only contain information already present in the tables. The study shows that PARENT maintains a high correlation as the number of such examples varies \cite{Dhingra}. These task-specific metrics are gaining prominence, with nine (9) papers considering PARENT and one (1) paper exploring PARENT-T.

To enhance the semantic alignment between generated texts and their references, embedding-based and pretrained-based metrics have been introduced. These metrics utilize contextualized embeddings and Transformer-based models to assess the quality and similarity of generated text to reference sentences. BERTScore \cite{zhang2019bertscore}, featured in 7 papers, calculates the cosine similarity between generated texts and the ground truth, offering a more nuanced understanding of text quality. MoverScore \cite{zhao2019moverscore}, a metric that allows many-to-one matching, is used in 2 papers and computes the Euclidean distances between words or n-grams. It enhances the evaluation process by considering partial alignments and offering insights into text quality. Notably, BLEURT \cite{sellam2020bleurt}, a Transformer-based trained metric, has been employed in 5 papers. This approach pretrains BERT with synthetically generated sentence pairs by mask-filling with BERT, back-translation, or randomly dropping words to assess NLG system performance \cite{sai2022survey}.

\paragraph{\textbf{Information Extraction Metrics}} In data-to-text evaluation, extractive evaluation methods were introduced in \citet{Wiseman2017} to assess the performance of the alignment of the information extraction model in the content selection and text planning in the generation process. A total of 11 studies adopted these metrics to rate their model's performance in content selection and planning tasks. This approach employs metrics like content selection (CS), content ordering (CO), and relational generation (RG). An Information Extraction (IE) system is used to extract content plans, identify candidate entities and value pairs present in the generated text, and predict their types. CS evaluates how accurately the system's extracted records match those in the reference output, considering precision and recall. RG assesses factuality by measuring the proportion of system-extracted records that also appear in the input table. CO evaluates the system's record ordering by computing the normalized Damerau-Levenshtein Distance between the sequence of extracted records and the reference output \cite{Wiseman2017}. 

The Table \ref{tab:metric-table} illustrate the distribution of these evaluation metrics identified in various research papers.

\subsubsection{\textbf{Human Evaluation}}

While automatic metrics offer certain advantages, human evaluation is often favored when assessing generated texts. This is due to its enhanced precision in evaluating aspects such as semantic adequacy, coherence, fluency, and the identification of numerical errors. Existing automatic metrics are often benchmarked against human evaluation results to determine their reliability and suitability. In human evaluations, the assessment of the quality of generated text varies widely, with different criteria used depending on the task. Due to the lack of a standardized human evaluation procedure in Natural Language Generation (NLG), and even in the naming conventions of the criteria, researchers often adopt diverse approaches to evaluate their generated texts \cite{van2019best}. In this review, we aim to show some aspects of human evaluation by categorizing them into the measures and methods of evaluation taking a cue from studies by \citet{belz2020disentangling}, and \citet{van2019best}.

\paragraph{\textbf{Quality Criteria Measures}}

In our analysis, certain studies lacked explicit details regarding their methods, tools, and design of quality criteria. However, for those that provided such information, we extracted relevant data. A notable observation is the considerable variation in the meanings associated with the names of the quality criteria. Table \ref{tab:human} enumerates the top ten prevalent naming conventions identified in the literature with \say{fluency} being the most used in 29 studies. Several terms, such as relevance, clarity, readability, and factual, among others, are notable examples that were not included in the table. It's crucial to note that the interpretation and task associated with these names may differ.

\paragraph{\textbf{Experimental Methods}}

This section of the human evaluation review explores various methodologies for obtaining and assessing responses based on quality criteria. Table \ref{tab:human} presents the human evaluation frameworks and metrics extracted from the studies. Our analysis reveals that, out of 28 studies conducting human evaluation assessments in crowd-sourcing platforms, 17 studies utilized Amazon Mechanical Turk as their primary crowd-sourcing platform. Additionally, we examined the linguistic background of annotators involved; 11 studies employed expert annotators, while eight and five studies engaged graduate students and paper authors, respectively.

Furthermore, we investigated the scale sizes used in each experiment. The most common scale size in 42 papers ranged from 1 to 5, followed by ranges of 10 to 30, and 50 and above in 5 and 4 papers, respectively. The 1-5 rank was the most popular scale range, appearing in 12 papers. Subsequently, the 0-100, 1-7, -100 to +100, and 0-5 scales were found in 6, 5, 3, and 3 papers, respectively. Two studies incorporated the TrueSkill Algorithm alongside the ranking task during evaluation, and various other scale sizes were identified in individual papers.

Additionally, we documented the agreement among annotators and the tools used in some studies. Statistical tools such as Krippendorff’s α, Fleiss’ Kappa, Cohen’s Kappa, and Weighted Kappa were employed in 3, 5, and 2 studies to measure agreements among annotators. In 12 papers, responses from participants were aggregated using averages.

\begin{table}[H]
\centering
\scriptsize
\resizebox{\columnwidth}{!}{%
\begin{NiceTabular}{|p{0.3\textwidth}|p{0.22\textwidth}|l|p{0.5\textwidth}|} \hline
\textbf{Design} & \textbf{Category} &  \textbf{Counts} & \textbf{Citations} \\ \hline

    \multirow{10}{*}{Quality Criterion}  &  Fluency  & 29 & \cite{Opitz2021}, \cite{Hargreaves2021}, \cite{Perez-Beltrachini2018}, \cite{Nema2018}, \cite{Ribeiro2021b}, \cite{Suadaa2021}, \cite{Rebuffel2020}, \cite{Kedzie2020}, \cite{laha-etal-2019}, \cite{Ribeiro2020}, \cite{Nan2021}, \cite{Rebuffel2022}, \cite{VanderLee2018}, \cite{castro-ferreira-etal-2020}, \cite{Moussallem2019}, \cite{Chen2020}, \cite{Konstas2017}, \cite{Fillippova2020}, \cite{Shen2020}, \cite{Lin2020}, \cite{Hardy2018}, \cite{Harkous2020}, \cite{Moryossef2019}, \cite{Zhao2020}, \cite{Agarwal2020}, \cite{Guo2020}, \cite{Freitag2018}, \cite{Ke2021}, \cite{Uehara2020} \\  \cmidrule{2-4} 
     & Grammaticality & 11 & \cite{Opitz2021}, \cite{Heidari2021}, \cite{Puduppully2019}, \cite{Suadaa2021}, \cite{Gong2019}, \cite{Gong2020}, \cite{Li2021}, \cite{Shao2019}, \cite{puduppully-lapata-2021}, \cite{Puduppully2020}, \cite{Arun2020} \\ \cline{2-4}  
     & Correctness & 11 & \cite{Heidari2021}, \cite{Gong2020a}, \cite{Suadaa2021}, \cite{Qader2019}, \cite{VanderLee2018}, \cite{castro-ferreira-etal-2020}, \cite{Chen2020}, \cite{Agarwal2020}, \cite{Kedzie2019}, \cite{Uehara2020}, \cite{Arun2020} \\ \cline{2-4}  
     & Adequacy & 10 & \cite{Shimorina2018}, \cite{Hargreaves2021}, \cite{Nema2018}, \cite{Ribeiro2021b}, \cite{laha-etal-2019}, \cite{Ribeiro2020}, \cite{Chang2021a}, \cite{Qader2019}, \cite{Moussallem2019}, \cite{Ke2021} \\ \cline{2-4} 
     & Coherence & 9 & \cite{Juraska2019}, \cite{Puduppully2019}, \cite{Suadaa2021}, \cite{Gong2019}, \cite{Gong2020}, \cite{Li2021}, \cite{laha-etal-2019}, \cite{Shao2019}, \cite{puduppully-lapata-2021}, \cite{Puduppully2020} \\ \cline{2-4} 
     & Coverage & 7 & \cite{Lu2022}, \cite{Rebuffel2020}, \cite{Qader2019}, \cite{castro-ferreira-etal-2020}, \cite{Konstas2017}, \cite{Fillippova2020}, \cite{Zhao2020}, \cite{Agarwal2020}, \cite{Guo2020} \\ \cline{2-4} 
     & Faithfulness & 7 & \cite{Wiseman2021}, \cite{Perez-Beltrachini2018}, \cite{Nan2021}, \cite{Fillippova2020}, \cite{Fan2020}, \cite{Moryossef2019}, \cite{Zhao2020} \\ \cline{2-4}  
     & Naturalness & 7 & \cite{Juraska2019}, \cite{Wiseman2021}, \cite{Wiseman2017}, \cite{Gong2020a}, \cite{Kale2020}, \cite{Kedzie2020}, \cite{Gehrmann2018}, \cite{Puzikov2018} \\ \cline{2-4} 
     & Conciseness & 7 & \cite{Puduppully2019}, \cite{Suadaa2021}, \cite{Gong2019}, \cite{Gong2020}, \cite{Li2021}, \cite{puduppully-lapata-2021}, \cite{Puduppully2020} \\ \cline{2-4} 
     & Similarity & 5 & \cite{Ribeiro2021b}, \cite{Ram2020}, \cite{Ribeiro2021}, \cite{Ribeiro2019}, \cite{Zhang2020}  \\ \hline

    \multirow{4}{*}{Crowd Sourcing Platform} & AMT & 17 & \cite{Wiseman2017}, \cite{Puduppully2019}, \cite{Moryossef2019}, \cite{Zhao2020}, \cite{castro-ferreira-etal-2020}, \cite{Ribeiro2019}, \cite{Ribeiro2020}, \cite{Chen2020}, \cite{Ribeiro2021}, \cite{puduppully-lapata-2021}, \cite{Suadaa2021}, \cite{Lu2022}, \cite{Xu2021}, \cite{Wiseman2021}, \cite{Chen2020b}, \cite{Perez-Beltrachini2018}, \cite{Harkous2020} \\ \cline{2-4} 
    & Prolific & 1 & \cite{Chang2021} \\ \cline{2-4}  
    & CrowdFlower & 1 & \cite{Moussallem2019} \\ \cline{2-4}  
    & Other & 9 & \cite{Gong2020}, \cite{Gehrmann2018}, \cite{Nie2020}, \cite{Nie2018}, \cite{Freitag2018}, \cite{Chang2021}, \cite{Fan2020}, \cite{Moussallem2019}, \cite{Guo2020}\\ \hline

    \multirow{3}{*}{Annotator Experience} &  Expert & 11 &  \cite{Ram2020}, \cite{Harkous2020}, \cite{Chen2020}, \cite{Fillippova2020}, \cite{Rebuffel2022}, \cite{Juraska2019}, \cite{Bai2020}, \cite{laha-etal-2019}, \cite{Moussallem2019}, \cite{Juraska2021}, \cite{Rebuffel2020} \\ \cline{2-4}
    & Graduate student  & 8 &   \cite{Gong2019}, \cite{Nema2018}, \cite{Gong2020a}, \cite{Chen2020}, \cite{Kedzie2019}, \cite{Suadaa2021}, \cite{Kedzie2020}, \cite{Li2021}\\ \cline{2-4}
    & Author & 5 &  \cite{Ferreira2019}, \cite{Harkous2020}, \cite{Juraska2021}, \cite{Arun2020}, \cite{Hargreaves2021} \\ \hline

    \multirow{10}{*}{Scale Sizes}  &  1 - 5 & 42 & \cite{Shimorina2018}, \cite{Hajdik2019}, \cite{Opitz2021}, \cite{Song2019}, \cite{Juraska2019}, \cite{Chen2019}, \cite{Juraska2021}, \cite{Hargreaves2021}, \cite{Heidari2021}, \cite{Lu2022}, \cite{Wiseman2021}, \cite{Perez-Beltrachini2018}, \cite{Nema2018}, \cite{Wiseman2017}, \cite{Ribeiro2021b}, \cite{Puduppully2019}, \cite{Gong2020a}, \cite{Su2021}, \cite{Ram2020}, \cite{Ribeiro2021}, \cite{Suadaa2021}, \cite{Kale2020},\cite{Nie2020}, \cite{Nie2018}, \cite{Wang2020a}, \cite{Gong2019}, \cite{Bai2020}, \cite{Liu2021}, \cite{Wang2021},, \cite{Xu2021}, \cite{Gong2020}, \cite{Li2021}, \cite{Rebuffel2020}, \cite{Chen2020b}, \cite{Kedzie2020}, \cite{laha-etal-2019}, \cite{Ribeiro2020}, \cite{Shao2019}, \cite{Ribeiro2019}, \cite{Chang2021a}, \cite{Chen2020}, \cite{Nan2021} \\ \cline{2-4} 
    & 10 - 30 & 5 & \cite{Rebuffel2022}, \cite{Qader2019}, \cite{VanderLee2018}, \cite{Zhang2020}, \cite{Moussallem2019} \\ \cline{2-4}
    & 50 and above  & 4 & \cite{Chang2021}, \cite{castro-ferreira-etal-2020}, \cite{Moussallem2019}, \cite{puduppully-lapata-2021} \\ \hline 

    \multirow{13}{*}{Scale Rank}  & -1 - +1   & 1 & \cite{Moryossef2019}  \\ \cline{2-4}
    & -100 - +100  & 3 & \cite{Puduppully2019}, \cite{puduppully-lapata-2021}, \cite{Suadaa2021} \\ \cline{2-4}
    & 0-1 & 1 &  \cite{Gong2020a} \\ \cline{2-4}
    & 0-2   & 1 & \cite{Su2021} \\ \cline{2-4}
    & 0-5   & 3 & \cite{Ram2020}, \cite{Chang2021}, \cite{Chang2021a} \\ \cline{2-4}
    & 0-100   & 6 & \cite{Ribeiro2019}, \cite{Zhao2020}, \cite{castro-ferreira-etal-2020}, \cite{Agarwal2020}, \cite{Zhang2020}, \cite{Guo2020} \\ \cline{2-4}
    & 1-3   & 7 & \cite{Chen2020b}, \cite{Lu2022}, \cite{Fillippova2020}, \cite{Rebuffel2022}, \cite{Fan2020}, \cite{Garneau2021}, \cite{Kedzie2020} \\ \cline{2-4} 
    & 1-4   & 1 & \cite{Suadaa2021} \\ \cline{2-4} 
    & 1-5   & 12 & \cite{Shao2019}, \cite{Shen2020}, \cite{laha-etal-2019}, \cite{Moussallem2019}, \cite{Wiseman2021}, \cite{Perez-Beltrachini2018}, \cite{Ribeiro2020}, \cite{Nema2018}, \cite{Qader2019}, \cite{Juraska2019}, \cite{Lin2020}, \cite{Freitag2018} \\ \cline{2-4}
    & 1-6   & 1 & \cite{Hardy2018}  \\ \cline{2-4} 
    & 1-7   & 5 & \cite{Wiseman2017}, \cite{VanderLee2018}, \cite{Ribeiro2021b}, \cite{Harkous2020}, \cite{Ribeiro2021} \\ \cline{2-4}
    & 1-100   & 1 & \cite{Fu2020} \\ \cline{2-4}
    & TrueSkill Algorithm   & 2 & \cite{Gehrmann2018}, \cite{Puzikov2018} \\ \hline 

    \multirow{6}{*}{Inter-annotator tools}  & Krippendorff’s α  & 3 & \cite{VanderLee2018}, \cite{puduppully-lapata-2021}, \cite{Juraska2021} \\ \cline{2-4}
    & Fleiss’ Kappa   & 5 & \cite{Shao2019}, \cite{Wang2020a}, \cite{Ke2021}, \cite{Liu2021}, \cite{Xu2021}  \\ \cline{2-4}
    & Cohen’s Kappa   & 2 & \cite{Harkous2020}, \cite{Chen2019} \\ \cline{2-4} 
    & Weighted Kappa   & 1 &  \cite{VanderLee2018} \\ \hline

    \multirow{4}{*}{Statistical Test } &  ANOVA + posthoc Tukey HSD test & 5 &  \cite{Wiseman2021}, \cite{Perez-Beltrachini2018}, \cite{Wiseman2017}, \cite{Puduppully2019}, \cite{Gong2019} \\ \cline{2-4} 
    & Wilcoxon Rank-Sum Test  & 4 & \cite{castro-ferreira-etal-2020}, \cite{Harkous2020}, \cite{Agarwal2020}, \cite{Guo2020} \\ \cline{2-4}
    & Pearson Correlation Coefficient  & 2 &  \cite{Ram2020}, \cite{laha-etal-2019}  \\ \cline{2-4} 
    & Kendall’s τ  & 1 & \cite{Kedzie2020} \\ \hline

    \multirow{1}{*}{Result Aggregation}  &  average  & 12  & \cite{Garneau2021}, \cite{Shao2019}, \cite{Shen2020}, \cite{Wiseman2021}, \cite{Ribeiro2020}, \cite{Nema2018}, \cite{Ribeiro2021b}, \cite{Ribeiro2019}, \cite{castro-ferreira-etal-2020}, \cite{Ram2020}, \cite{Chang2021}, \cite{Freitag2018} \\ \hline

\end{NiceTabular}%
} \caption{Human Evaluation Criteria and Frameworks. \textit{Amazon Mechanical Turk (AMT)}} 
    \label{tab:human}
\end{table}

\subsection{Application Areas}
\label{sec:app}

\begin{table}[H]
    \centering
    \scriptsize
    \begin{tabular}{|p{0.3\textwidth}|l|p{0.5\textwidth}|}
    \hline
        \textbf{Application Areas} & \textbf{Count} & \textbf{Papers} \\ \hline
        Weather Forecasting & 3 & \cite{VanderLee2018}, \cite{Arun2020}, \cite{Heidari2021}\\ \hline
        Robo-Journalism & 3 & \cite{Teixeira2020}, \cite{Wiseman2017}, \cite{VanderLee2018}\\ \hline
        Translation and Multilingualism & 12 & \cite{Song2019}, \cite{Shao2019}, \cite{castro-ferreira-etal-2020}, \cite{Garneau2021}, \cite{Agarwal2020}, \cite{Lu2022}, \cite{Li2020}, \cite{Teixeira2020}, \cite{Fan2020}, \cite{Xu2021a}, \cite{Nema2018}, \cite{Moussallem2020} \\ \hline
        Relational Databases & 2 & \cite{Nan2021}, \cite{Xu2018} \\ \hline
        Question Generation and Answering & 4 & \cite{Juraska2019}, \cite{Ke2021}, \cite{Ke2021}, \cite{Nan2021} \\ \hline
        Legal Domain & 1 & \cite{Garneau2021} \\ \hline
        Dialogue Systems & 25 & \cite{Harkous2020}, \cite{Chang2021a}, \cite{Kedzie2020}, \cite{Juraska2021}, \cite{Kale2020}, \cite{Wiseman2018}, \cite{Chen2020b}, \cite{Nan2021}, \cite{Zhang2020}, \cite{Lu2022}, \cite{Lin2020}, \cite{Wiseman2021}, \cite{Gehrmann2018}, \cite{Puzikov2018}, \cite{Nie2020}, \cite{Freitag2018}, \cite{Shen2020}, \cite{Shimorina2018}, \cite{Chang2021}, \cite{Kedzie2019}, \cite{Qader2019}, \cite{Juraska2019}, \cite{Wang2021}, \cite{Xu2021}, \cite{Gong2020}\\ \hline
        Financial Reporting & 1 & \cite{Uehara2020} \\ \hline
        Biography Generation & 15 & \cite{Wiseman2018}, \cite{Parikh2020}, \cite{Dhingra}, \cite{Chen2020b}, \cite{Nan2021}, \cite{Nema2018}, \cite{Chen2020}, \cite{Fillippova2020}, \cite{Chen2019}, \cite{Rebuffel2022}, \cite{laha-etal-2019}, \cite{Zhang2020}, \cite{Rebuffel2020}, \cite{Perez-Beltrachini2018}, \cite{Wiseman2021} \\ \hline
        Advertising  & 1 & \cite{Shao2019} \\ \hline
        Recipe Generation  & 1 & \cite{Shao2019} \\ \hline
        Sports Narration  & 18 & \cite{Nie2018}, \cite{Wiseman2017}, \cite{Puduppully2019}, \cite{Parikh2020}, \cite{Puduppully2020}, \cite{Chen2020a}, \cite{Nan2021}, \cite{Gong2019}, \cite{Jiang2020}, \cite{puduppully-lapata-2021}, \cite{Wang2019}, \cite{Suadaa2021}, \cite{Gong2020}, \cite{Li2021}, \cite{Lin2020}, \cite{VanderLee2018}, \cite{Iso2020a}, \cite{Iso2020} \\ \hline
    \end{tabular}
    \caption{Application Areas of Data-to-text Generation.}
    \label{tab:application}
\end{table}

Data-to-text generation finds extensive application in diverse domains, reflecting its versatility and value in addressing specific needs, as shown in Table \ref{tab:application}. It plays a pivotal role in dialogue systems, where 26 studies focus on generating dialogues across various conversational contexts. Another significant application domain is sports narration, with 18 studies employing data-to-text generation to create textual summaries of game match records, encompassing player details, team information, and scores. In the realm of biography generation, 15 studies work on generating biographical texts for individuals with data often sourced from platforms like English Wikipedia. The application extends to translation and multilingualism, with 12 studies leveraging data-to-text techniques to tackle multilingual challenges. Additionally, data-to-text methods have proven effective in question generation and answering, exemplified by four identified studies in this domain. The application footprint extends to weather forecasting, robo-journalism, relational databases, the legal domain, and financial reporting, each with several studies showcasing the practical utility of data-to-text generation in distinct contexts. Furthermore, advertising and recipe generation domains have harnessed data-to-text techniques effectively. This comprehensive coverage highlights the adaptability and broad applicability of data-to-text generation in diverse scenarios and underscores its role in addressing specific needs across multiple domains.

\section{Discussion}
\label{sec:dicuss}

Based on the data gathered from various studies, we will explore researchers' preferences. The datasets outlined in Table \ref{tab:datset} reveal that, in data-to-text NLG, researchers tend to favor WebNLG and E2E over other datasets due to the competition challenges and dedicated conferences associated with them. Moreover, these datasets, being human-curated, offer more natural and fluent content compared to online-extracted datasets, contributing to their widespread acceptance.

In terms of evaluation metrics, BLEU emerges as the most commonly used automatic metric, enjoying broad acceptance across various data-to-text tasks. This observation is supported by the considerable number of studies that incorporate this metric in Table \ref{tab:metric-table}.

In addition, the dominance of large language models is evident in their effectiveness for data-to-text generation, as reflected by the widespread adoption of transformer models, as illustrated in Table \ref{tab:model}. 

Lastly, there is no universally preferred technique for addressing the hallucination problem in data-to-text. However, based on our observations, most mitigation measures are task-specific, with data refinement being a more general and effective method. This involves processes such as deduplication, cleaning, and factuality grounding, exerting control over the content the generation model encounters during training.

\section{Recommendation and Future Directions}
\label{sec:recom}

We have conducted a thorough analysis of the prevailing trends and have provided answers to the research questions within the scope of this literature. Nonetheless, it has come to our attention that there is a need to extend research efforts to encompass more low-resourced languages as seen in Section \ref{sec:lang}. Additionally, in Section \ref{sec:models} owing to the temporal limitations defined in the exclusion criteria, we did not include papers pertaining to recent large language models such as GPT-3.5 and GPT-4 \cite{ye2023comprehensive, Achiam2023GPT4TR} and LLAMA \cite{touvron2023llama, touvron2023llama2}. In future investigations, our emphasis will be on studies that incorporate these advanced technologies into their research.

Moreover, from observations in Section \ref{sec:eval}, we recommend that future studies place greater emphasis on the utilization of contextual evaluation metrics for assessing the performance of data-to-text generation. These metrics have shown notable advantages in terms of semantic accuracy in data-to-text pairs, and their inclusion in evaluation frameworks is a direction worth exploring. 

A standardized approach to human evaluation in the data-to-text field is essential. We strongly recommend authors to provide detailed explanations of their human evaluation procedures, including quality criteria definitions, response elicitation platforms, participants' knowledge backgrounds, etc. We also encourage the broader NLG community to collaboratively establish a universal naming convention to disambiguate similar terms and associated tasks.

Furthermore, referring to Section \ref{sec:hal} and considering the richness of general knowledge that recent LLMs possess, we propose an advancement in their hallucination mitigation methods compared to task-specific LLMs. This improvement could focus more on addressing numerical and logical inference hallucination in the generated text.

\section{Conclusion}
\label{sec:conclude}

This systematic review of data-to-text generation provides a comprehensive overview of the field, including its trends, challenges, and advancements. The review consolidates knowledge on datasets, language considerations, models, hallucination mitigation, and applications to guide future research endeavors. The insights gained from this review contribute to a deeper understanding of data-to-text generation, paving the way for continued innovation and progress. Addressing the identified trends and challenges will be crucial in advancing the capabilities and applicability of data-to-text generation systems as the field continues to evolve.

\backmatter

\bmhead{Acknowledgments}
This work was conducted with the financial support of the Science Foundation Ireland Centre for Research Training in Artificial Intelligence under Grant No. 18/CRT/6223. This publication has emanated from research conducted with the financial support of Science Foundation Ireland under Grant number 18/CRT/6223. For the purpose of Open Access, the author has applied a CC BY public copyright license to any Author Accepted Manuscript version arising from this submission.

\bibliography{sn-bibliography}


\begin{thebibliography}{143}
\ifx \bisbn   \undefined \def \bisbn  #1{ISBN #1}\fi
\ifx \binits  \undefined \def \binits#1{#1}\fi
\ifx \bauthor  \undefined \def \bauthor#1{#1}\fi
\ifx \batitle  \undefined \def \batitle#1{#1}\fi
\ifx \bjtitle  \undefined \def \bjtitle#1{#1}\fi
\ifx \bvolume  \undefined \def \bvolume#1{\textbf{#1}}\fi
\ifx \byear  \undefined \def \byear#1{#1}\fi
\ifx \bissue  \undefined \def \bissue#1{#1}\fi
\ifx \bfpage  \undefined \def \bfpage#1{#1}\fi
\ifx \blpage  \undefined \def \blpage #1{#1}\fi
\ifx \burl  \undefined \def \burl#1{\textsf{#1}}\fi
\ifx \doiurl  \undefined \def \doiurl#1{\url{https://doi.org/#1}}\fi
\ifx \betal  \undefined \def \betal{\textit{et al.}}\fi
\ifx \binstitute  \undefined \def \binstitute#1{#1}\fi
\ifx \binstitutionaled  \undefined \def \binstitutionaled#1{#1}\fi
\ifx \bctitle  \undefined \def \bctitle#1{#1}\fi
\ifx \beditor  \undefined \def \beditor#1{#1}\fi
\ifx \bpublisher  \undefined \def \bpublisher#1{#1}\fi
\ifx \bbtitle  \undefined \def \bbtitle#1{#1}\fi
\ifx \bedition  \undefined \def \bedition#1{#1}\fi
\ifx \bseriesno  \undefined \def \bseriesno#1{#1}\fi
\ifx \blocation  \undefined \def \blocation#1{#1}\fi
\ifx \bsertitle  \undefined \def \bsertitle#1{#1}\fi
\ifx \bsnm \undefined \def \bsnm#1{#1}\fi
\ifx \bsuffix \undefined \def \bsuffix#1{#1}\fi
\ifx \bparticle \undefined \def \bparticle#1{#1}\fi
\ifx \barticle \undefined \def \barticle#1{#1}\fi
\bibcommenthead
\ifx \bconfdate \undefined \def \bconfdate #1{#1}\fi
\ifx \botherref \undefined \def \botherref #1{#1}\fi
\ifx \url \undefined \def \url#1{\textsf{#1}}\fi
\ifx \bchapter \undefined \def \bchapter#1{#1}\fi
\ifx \bbook \undefined \def \bbook#1{#1}\fi
\ifx \bcomment \undefined \def \bcomment#1{#1}\fi
\ifx \oauthor \undefined \def \oauthor#1{#1}\fi
\ifx \citeauthoryear \undefined \def \citeauthoryear#1{#1}\fi
\ifx \endbibitem  \undefined \def \endbibitem {}\fi
\ifx \bconflocation  \undefined \def \bconflocation#1{#1}\fi
\ifx \arxivurl  \undefined \def \arxivurl#1{\textsf{#1}}\fi
\csname PreBibitemsHook\endcsname

\bibitem[\protect\citeauthoryear{Reiter and Dale}{1997}]{Reiter1997}
\begin{botherref}
\oauthor{\bsnm{Reiter}, \binits{E.}},
\oauthor{\bsnm{Dale}, \binits{R.}}:
{Building applied natural language generation systems}
(1997).
\doiurl{10.1017/S1351324997001502}
\end{botherref}
\endbibitem

\bibitem[\protect\citeauthoryear{Hardy and Vlachos}{2018}]{Hardy2018}
\begin{botherref}
\oauthor{\bsnm{Hardy}},
\oauthor{\bsnm{Vlachos}, \binits{A.}}:
{Guided neural language generation for abstractive summarization using abstract meaning representation}.
Proceedings of the 2018 Conference on Empirical Methods in Natural Language Processing, EMNLP 2018,
768--773
(2018)
\doiurl{10.18653/v1/d18-1086}
{\href{https://arxiv.org/abs/1808.09160}{{arXiv:1808.09160}}}
\end{botherref}
\endbibitem

\bibitem[\protect\citeauthoryear{Dou et~al.}{2023}]{dou2023automatic}
\begin{botherref}
\oauthor{\bsnm{Dou}, \binits{Y.}},
\oauthor{\bsnm{Laban}, \binits{P.}},
\oauthor{\bsnm{Gardent}, \binits{C.}},
\oauthor{\bsnm{Xu}, \binits{W.}}:
Automatic and Human-AI Interactive Text Generation
(2023)
\end{botherref}
\endbibitem

\bibitem[\protect\citeauthoryear{Song et~al.}{2019}]{Song2019}
\begin{barticle}
\bauthor{\bsnm{Song}, \binits{L.}},
\bauthor{\bsnm{Gildea}, \binits{D.}},
\bauthor{\bsnm{Zhang}, \binits{Y.}},
\bauthor{\bsnm{Wang}, \binits{Z.}},
\bauthor{\bsnm{Su}, \binits{J.}}:
\batitle{{Semantic Neural Machine Translation Using AMR}}.
\bjtitle{Transactions of the Association for Computational Linguistics}
\bvolume{7},
\bfpage{19}--\blpage{31}
(\byear{2019})
\doiurl{10.1162/tacl_a_00252}
{\href{https://arxiv.org/abs/1902.07282}{{arXiv:1902.07282}}}
\end{barticle}
\endbibitem

\bibitem[\protect\citeauthoryear{Mokady et~al.}{2021}]{mokady2021clipcap}
\begin{botherref}
\oauthor{\bsnm{Mokady}, \binits{R.}},
\oauthor{\bsnm{Hertz}, \binits{A.}},
\oauthor{\bsnm{Bermano}, \binits{A.H.}}:
ClipCap: CLIP Prefix for Image Captioning
(2021)
\end{botherref}
\endbibitem

\bibitem[\protect\citeauthoryear{Harrison et~al.}{2019}]{harrison2019maximizing}
\begin{botherref}
\oauthor{\bsnm{Harrison}, \binits{V.}},
\oauthor{\bsnm{Reed}, \binits{L.}},
\oauthor{\bsnm{Oraby}, \binits{S.}},
\oauthor{\bsnm{Walker}, \binits{M.}}:
Maximizing stylistic control and semantic accuracy in nlg: Personality variation and discourse contrast.
arXiv preprint arXiv:1907.09527
(2019)
\end{botherref}
\endbibitem

\bibitem[\protect\citeauthoryear{Erdem et~al.}{2022}]{Erdem2022}
\begin{barticle}
\bauthor{\bsnm{Erdem}, \binits{E.}},
\bauthor{\bsnm{Plank}, \binits{B.}},
\bauthor{\bsnm{Gatt}, \binits{A.}},
\bauthor{\bsnm{Krahmer}, \binits{E.}},
\bauthor{\bsnm{Sharma}, \binits{M.}},
\bauthor{\bsnm{Gogineni}, \binits{A.}},
\bauthor{\bsnm{Ramakrishnan}, \binits{N.}},
\bauthor{\bsnm{Li}, \binits{W.}},
\bauthor{\bsnm{Wu}, \binits{W.}},
\bauthor{\bsnm{Chen}, \binits{M.}},
\bauthor{\bsnm{Liu}, \binits{J.}},
\bauthor{\bsnm{Xiao}, \binits{X.}},
\bauthor{\bsnm{Wu}, \binits{H.}}:
\batitle{{Neural natural language generation: A survey on multilinguality, multimodality, controllability and learning}}.
\bjtitle{Journal of Artificial Intelligence Research}
\bvolume{73},
\bfpage{1131}--\blpage{1207}
(\byear{2022})
\doiurl{10.1613/jair.5714}
{\href{https://arxiv.org/abs/2207.12571}{{arXiv:2207.12571}}}
\end{barticle}
\endbibitem

\bibitem[\protect\citeauthoryear{Akermi et~al.}{2020}]{akermi2020transformer}
\begin{bchapter}
\bauthor{\bsnm{Akermi}, \binits{I.}},
\bauthor{\bsnm{Heinecke}, \binits{J.}},
\bauthor{\bsnm{Herledan}, \binits{F.}}:
\bctitle{Transformer based natural language generation for question-answering}.
In: \bbtitle{Proceedings of the 13th International Conference on Natural Language Generation},
pp. \bfpage{349}--\blpage{359}
(\byear{2020})
\end{bchapter}
\endbibitem

\bibitem[\protect\citeauthoryear{Wiseman et~al.}{2018}]{Wiseman2018}
\begin{botherref}
\oauthor{\bsnm{Wiseman}, \binits{S.}},
\oauthor{\bsnm{Shieber}, \binits{S.M.}},
\oauthor{\bsnm{Rush}, \binits{A.M.}}:
{Learning Neural Templates for Text Generation}.
Technical report
(2018)
\end{botherref}
\endbibitem

\bibitem[\protect\citeauthoryear{Parikh et~al.}{2020}]{Parikh2020}
\begin{botherref}
\oauthor{\bsnm{Parikh}, \binits{A.P.}},
\oauthor{\bsnm{Wang}, \binits{X.}},
\oauthor{\bsnm{Gehrmann}, \binits{S.}},
\oauthor{\bsnm{Faruqui}, \binits{M.}},
\oauthor{\bsnm{Dhingra}, \binits{B.}},
\oauthor{\bsnm{Yang}, \binits{D.}},
\oauthor{\bsnm{Das}, \binits{D.}}:
{ToTTo: A controlled table-to-text generation dataset}.
EMNLP 2020 - 2020 Conference on Empirical Methods in Natural Language Processing, Proceedings of the Conference,
1173--1186
(2020)
\doiurl{10.18653/v1/2020.emnlp-main.89}
{\href{https://arxiv.org/abs/2004.14373}{{arXiv:2004.14373}}}
\end{botherref}
\endbibitem

\bibitem[\protect\citeauthoryear{Laha et~al.}{2019}]{laha-etal-2019}
\begin{barticle}
\bauthor{\bsnm{Laha}, \binits{A.}},
\bauthor{\bsnm{Jain}, \binits{P.}},
\bauthor{\bsnm{Mishra}, \binits{A.}},
\bauthor{\bsnm{Sankaranarayanan}, \binits{K.}}:
\batitle{{Scalable Micro-planned Generation of Discourse from Structured Data}}.
\bjtitle{Computational Linguistics}
\bvolume{45}(\bissue{4}),
\bfpage{737}--\blpage{763}
(\byear{2019})
\doiurl{10.1162/coli_a_00363}
\end{barticle}
\endbibitem

\bibitem[\protect\citeauthoryear{Teixeira et~al.}{2020}]{Teixeira2020}
\begin{botherref}
\oauthor{\bsnm{Teixeira}, \binits{A.L.R.}},
\oauthor{\bsnm{Campos}, \binits{J.G.M.}},
\oauthor{\bsnm{Cunha}, \binits{R.}},
\oauthor{\bsnm{Ferreira}, \binits{T.C.}},
\oauthor{\bsnm{Pagano}, \binits{A.S.}},
\oauthor{\bsnm{Cozman}, \binits{F.G.}}:
Damata: A robot-journalist covering the brazilian amazon deforestation.
INLG 2020 - 13th International Conference on Natural Language Generation, Proceedings,
103--106
(2020)
\end{botherref}
\endbibitem

\bibitem[\protect\citeauthoryear{Obeid and Hoque}{2020}]{obeid2020chart}
\begin{botherref}
\oauthor{\bsnm{Obeid}, \binits{J.}},
\oauthor{\bsnm{Hoque}, \binits{E.}}:
Chart-to-text: Generating natural language descriptions for charts by adapting the transformer model.
arXiv preprint arXiv:2010.09142
(2020)
\end{botherref}
\endbibitem

\bibitem[\protect\citeauthoryear{Xu}{2018}]{Xu2018}
\begin{botherref}
\oauthor{\bsnm{Xu}, \binits{K.}}:
{SQL-to-Text Generation with Graph-to-Sequence Model},
931--936
(2018)
\end{botherref}
\endbibitem

\bibitem[\protect\citeauthoryear{Gatt and Krahmer}{2018}]{Gatt2018a}
\begin{barticle}
\bauthor{\bsnm{Gatt}, \binits{A.}},
\bauthor{\bsnm{Krahmer}, \binits{E.}}:
\batitle{{Survey of the state of the art in natural language generation: Core tasks, applications and evaluation}}.
\bjtitle{Journal of Artificial Intelligence Research}
\bvolume{61},
\bfpage{1}--\blpage{64}
(\byear{2018})
\doiurl{10.1613/jair.5714}
{\href{https://arxiv.org/abs/1703.09902}{{arXiv:1703.09902}}}
\end{barticle}
\endbibitem

\bibitem[\protect\citeauthoryear{Jiang et~al.}{2020}]{Jiang2020}
\begin{barticle}
\bauthor{\bsnm{Jiang}, \binits{N.}},
\bauthor{\bsnm{Chen}, \binits{J.}},
\bauthor{\bsnm{Zhou}, \binits{R.G.}},
\bauthor{\bsnm{Wu}, \binits{C.}},
\bauthor{\bsnm{Chen}, \binits{H.}},
\bauthor{\bsnm{Zheng}, \binits{J.}},
\bauthor{\bsnm{Wan}, \binits{T.}}:
\batitle{{PAN: Pipeline assisted neural networks model for data-to-text generation in social internet of things}}.
\bjtitle{Information Sciences}
\bvolume{530},
\bfpage{167}--\blpage{179}
(\byear{2020})
\doiurl{10.1016/j.ins.2020.03.080}
\end{barticle}
\endbibitem

\bibitem[\protect\citeauthoryear{{Gonz{\'{a}}lez Corbelle} et~al.}{2022}]{GonzalezCorbelle2022}
\begin{botherref}
\oauthor{\bsnm{{Gonz{\'{a}}lez Corbelle}}, \binits{J.}},
\oauthor{\bsnm{Bugar\'in-Diz}, \binits{A.}},
\oauthor{\bsnm{Alonso-Moral}, \binits{J.}},
\oauthor{\bsnm{Taboada}, \binits{J.}}:
{Dealing with hallucination and omission in neural Natural Language Generation: A use case on meteorology.}
Proceedings of the 15th International Conference on Natural Language Generation,
121--130
(2022)
\end{botherref}
\endbibitem

\bibitem[\protect\citeauthoryear{Zhang et~al.}{2016}]{zhang2016}
\begin{botherref}
\oauthor{\bsnm{Zhang}, \binits{J.}},
\oauthor{\bsnm{Yao}, \binits{J.-g.}},
\oauthor{\bsnm{Wan}, \binits{X.}}:
{Towards constructing sports news from live text commentary},
1361--1371
(2016)
\end{botherref}
\endbibitem

\bibitem[\protect\citeauthoryear{Plachouras et~al.}{2016}]{Plachouras2016}
\begin{botherref}
\oauthor{\bsnm{Plachouras}, \binits{V.}},
\oauthor{\bsnm{Smiley}, \binits{C.}},
\oauthor{\bsnm{Bretz}, \binits{H.}},
\oauthor{\bsnm{Taylor}, \binits{O.}},
\oauthor{\bsnm{Leidner}, \binits{J.L.}},
\oauthor{\bsnm{Song}, \binits{D.}},
\oauthor{\bsnm{Schilder}, \binits{F.}}:
{Interacting with financial data using natural language}.
SIGIR 2016 - Proceedings of the 39th International ACM SIGIR Conference on Research and Development in Information Retrieval,
1121--1124
(2016)
\doiurl{10.1145/2911451.2911457}
\end{botherref}
\endbibitem

\bibitem[\protect\citeauthoryear{Monfroglio et~al.}{2022}]{Monfroglio2022}
\begin{botherref}
\oauthor{\bsnm{Monfroglio}, \binits{E.}},
\oauthor{\bsnm{Anselma}, \binits{L.}},
\oauthor{\bsnm{Mazzei}, \binits{A.}}:
{Personalizing Weekly Diet Reports}
(2022).
\url{https://aclanthology.org/2022.nlg4health-1.5}
\end{botherref}
\endbibitem

\bibitem[\protect\citeauthoryear{Lebret et~al.}{2016}]{Lebret2016}
\begin{botherref}
\oauthor{\bsnm{Lebret}, \binits{R.}},
\oauthor{\bsnm{Grangier}, \binits{D.}},
\oauthor{\bsnm{Auli}, \binits{M.}}:
{Neural Text Generation from Structured Data with Application to the Biography Domain}.
EMNLP 2016 - Conference on Empirical Methods in Natural Language Processing, Proceedings,
1203--1213
(2016)
\doiurl{10.18653/V1/D16-1128}
{\href{https://arxiv.org/abs/1603.07771}{{arXiv:1603.07771}}}
\end{botherref}
\endbibitem

\bibitem[\protect\citeauthoryear{Heidari et~al.}{2021}]{Heidari2021}
\begin{barticle}
\bauthor{\bsnm{Heidari}, \binits{P.}},
\bauthor{\bsnm{Einolghozati}, \binits{A.}},
\bauthor{\bsnm{Jain}, \binits{S.}},
\bauthor{\bsnm{Batra}, \binits{S.}},
\bauthor{\bsnm{Callender}, \binits{L.}},
\bauthor{\bsnm{Arun}, \binits{A.}},
\bauthor{\bsnm{Mei}, \binits{S.}},
\bauthor{\bsnm{Gupta}, \binits{S.}},
\bauthor{\bsnm{Donmez}, \binits{P.}},
\bauthor{\bsnm{Bhardwaj}, \binits{V.}},
\bauthor{\bsnm{Kumar}, \binits{A.}},
\bauthor{\bsnm{White}, \binits{M.}}:
\batitle{{Getting to Production with Few-shot Natural Language Generation Models}}.
\bjtitle{SIGDIAL 2021 - 22nd Annual Meeting of the Special Interest Group on Discourse and Dialogue, Proceedings of the Conference}
\bvolume{2},
\bfpage{66}--\blpage{76}
(\byear{2021})
\end{barticle}
\endbibitem

\bibitem[\protect\citeauthoryear{der Lee et~al.}{2018}]{VanderLee2018}
\begin{botherref}
\oauthor{\bsnm{Lee}, \binits{C.V.}},
\oauthor{\bsnm{Krahmer}, \binits{E.}},
\oauthor{\bsnm{Wubben}, \binits{S.}}:
Automated learning of templates for data-to-text generation: comparing rule-based, statistical and neural methods.
INLG 2018 - 11th International Natural Language Generation Conference, Proceedings of the Conference,
35--45
(2018)
\doiurl{10.18653/v1/w18-6504}
\end{botherref}
\endbibitem

\bibitem[\protect\citeauthoryear{Xu et~al.}{2021}]{Xu2021}
\begin{botherref}
\oauthor{\bsnm{Xu}, \binits{X.}},
\oauthor{\bsnm{Du{\v{s}}ek}, \binits{O.}},
\oauthor{\bsnm{Rieser}, \binits{V.}},
\oauthor{\bsnm{Konstas}, \binits{I.}}:
{AGGGEN: Ordering and aggregating while generating}.
ACL-IJCNLP 2021 - 59th Annual Meeting of the Association for Computational Linguistics and the 11th International Joint Conference on Natural Language Processing, Proceedings of the Conference,
1419--1434
(2021)
\doiurl{10.18653/v1/2021.acl-long.113}
{\href{https://arxiv.org/abs/2106.05580}{{arXiv:2106.05580}}}
\end{botherref}
\endbibitem

\bibitem[\protect\citeauthoryear{Liang et~al.}{2009}]{Liang2009}
\begin{botherref}
\oauthor{\bsnm{Liang}, \binits{P.}},
\oauthor{\bsnm{Jordan}, \binits{M.I.}},
\oauthor{\bsnm{Klein}, \binits{D.}}:
{Learning Semantic Correspondences with Less Supervision}
(2009).
\url{https://aclanthology.org/P09-1011}
\end{botherref}
\endbibitem

\bibitem[\protect\citeauthoryear{Nema et~al.}{2018}]{Nema2018}
\begin{barticle}
\bauthor{\bsnm{Nema}, \binits{P.}},
\bauthor{\bsnm{Shetty}, \binits{S.}},
\bauthor{\bsnm{Jain}, \binits{P.}},
\bauthor{\bsnm{Laha}, \binits{A.}},
\bauthor{\bsnm{Sankaranarayanan}, \binits{K.}},
\bauthor{\bsnm{Khapra}, \binits{M.M.}}:
\batitle{{Generating Descriptions from Structured Data Using a Bifocal Attention Mechanism and Gated Orthogonalization}}.
\bjtitle{NAACL HLT 2018 - 2018 Conference of the North American Chapter of the Association for Computational Linguistics: Human Language Technologies - Proceedings of the Conference}
\bvolume{1},
\bfpage{1539}--\blpage{1550}
(\byear{2018})
\doiurl{10.18653/V1/N18-1139}
{\href{https://arxiv.org/abs/1804.07789}{{arXiv:1804.07789}}}
\end{barticle}
\endbibitem

\bibitem[\protect\citeauthoryear{Shimorina and Gardent}{2018}]{Shimorina2018}
\begin{botherref}
\oauthor{\bsnm{Shimorina}, \binits{A.}},
\oauthor{\bsnm{Gardent}, \binits{C.}}:
{Handling rare items in data-to-text generation}.
INLG 2018 - 11th International Natural Language Generation Conference, Proceedings of the Conference,
360--370
(2018)
\doiurl{10.18653/v1/w18-6543}
\end{botherref}
\endbibitem

\bibitem[\protect\citeauthoryear{Juraska et~al.}{2019}]{Juraska2019}
\begin{botherref}
\oauthor{\bsnm{Juraska}, \binits{J.}},
\oauthor{\bsnm{Bowden}, \binits{K.K.}},
\oauthor{\bsnm{Walker}, \binits{M.}}:
{ViGGO: A Video Game Corpus for Data-To-Text Generation in Open-Domain Conversation}.
INLG 2019 - 12th International Conference on Natural Language Generation, Proceedings of the Conference,
164--172
(2019)
\doiurl{10.18653/V1/W19-8623}
{\href{https://arxiv.org/abs/1910.12129}{{arXiv:1910.12129}}}
\end{botherref}
\endbibitem

\bibitem[\protect\citeauthoryear{Chen et~al.}{2020}]{Chen2020}
\begin{botherref}
\oauthor{\bsnm{Chen}, \binits{Z.}},
\oauthor{\bsnm{Chen}, \binits{W.}},
\oauthor{\bsnm{Zha}, \binits{H.}},
\oauthor{\bsnm{Zhou}, \binits{X.}},
\oauthor{\bsnm{Zhang}, \binits{Y.}},
\oauthor{\bsnm{Sundaresan}, \binits{S.}},
\oauthor{\bsnm{Wang}, \binits{W.Y.}}:
{Logic2Text: High-Fidelity Natural Language Generation from Logical Forms}.
Findings of the Association for Computational Linguistics Findings of ACL: EMNLP 2020,
2096--2111
(2020)
\doiurl{10.18653/V1/2020.FINDINGS-EMNLP.190}
{\href{https://arxiv.org/abs/2004.14579}{{arXiv:2004.14579}}}
\end{botherref}
\endbibitem

\bibitem[\protect\citeauthoryear{Ji et~al.}{2022}]{Ji2022}
\begin{botherref}
\oauthor{\bsnm{Ji}, \binits{Z.}},
\oauthor{\bsnm{Lee}, \binits{N.}},
\oauthor{\bsnm{Frieske}, \binits{R.}},
\oauthor{\bsnm{Yu}, \binits{T.}},
\oauthor{\bsnm{Su}, \binits{D.}},
\oauthor{\bsnm{Xu}, \binits{Y.}},
\oauthor{\bsnm{Ishii}, \binits{E.}},
\oauthor{\bsnm{Bang}, \binits{Y.}},
\oauthor{\bsnm{Madotto}, \binits{A.}},
\oauthor{\bsnm{Fung}, \binits{P.}}:
{Survey of Hallucination in Natural Language Generation}.
ACM Computing Surveys
\textbf{1}(1)
(2022)
\doiurl{10.1145/3571730}
{\href{https://arxiv.org/abs/2202.03629}{{arXiv:2202.03629}}}
\end{botherref}
\endbibitem

\bibitem[\protect\citeauthoryear{Fatima et~al.}{2022}]{Fatima2022}
\begin{barticle}
\bauthor{\bsnm{Fatima}, \binits{N.}},
\bauthor{\bsnm{Imran}, \binits{A.S.}},
\bauthor{\bsnm{Kastrati}, \binits{Z.}},
\bauthor{\bsnm{Daudpota}, \binits{S.M.}},
\bauthor{\bsnm{Soomro}, \binits{A.}}:
\batitle{{A Systematic Literature Review on Text Generation Using Deep Neural Network Models}}.
\bjtitle{IEEE Access}
\bvolume{10},
\bfpage{53490}--\blpage{53503}
(\byear{2022})
\doiurl{10.1109/ACCESS.2022.3174108}
\end{barticle}
\endbibitem

\bibitem[\protect\citeauthoryear{Li et~al.}{2022}]{Li2022}
\begin{botherref}
\oauthor{\bsnm{Li}, \binits{W.}},
\oauthor{\bsnm{Wu}, \binits{W.}},
\oauthor{\bsnm{Chen}, \binits{M.}},
\oauthor{\bsnm{Liu}, \binits{J.}},
\oauthor{\bsnm{Xiao}, \binits{X.}},
\oauthor{\bsnm{Wu}, \binits{H.}}:
{Faithfulness in Natural Language Generation: A Systematic Survey of Analysis, Evaluation and Optimization Methods},
1--52
(2022)
{\href{https://arxiv.org/abs/2203.05227}{{arXiv:2203.05227}}}
\end{botherref}
\endbibitem

\bibitem[\protect\citeauthoryear{Sai et~al.}{2023}]{Sai2023}
\begin{botherref}
\oauthor{\bsnm{Sai}, \binits{A.B.}},
\oauthor{\bsnm{Mohankumar}, \binits{A.K.}},
\oauthor{\bsnm{Khapra}, \binits{M.M.}}:
{A Survey of Evaluation Metrics Used for NLG Systems}.
ACM Computing Surveys
\textbf{55}(2)
(2023)
\doiurl{10.1145/3485766}
{\href{https://arxiv.org/abs/2008.12009}{{arXiv:2008.12009}}}
\end{botherref}
\endbibitem

\bibitem[\protect\citeauthoryear{Lu et~al.}{2018}]{Lu2018}
\begin{botherref}
\oauthor{\bsnm{Lu}, \binits{S.}},
\oauthor{\bsnm{Zhu}, \binits{Y.}},
\oauthor{\bsnm{Zhang}, \binits{W.}},
\oauthor{\bsnm{Wang}, \binits{J.}},
\oauthor{\bsnm{Yu}, \binits{Y.}}:
{Neural Text Generation: Past, Present and Beyond}
(2018)
{\href{https://arxiv.org/abs/1803.07133}{{arXiv:1803.07133}}}
\end{botherref}
\endbibitem

\bibitem[\protect\citeauthoryear{Sharma et~al.}{2022}]{Sharma2022}
\begin{botherref}
\oauthor{\bsnm{Sharma}, \binits{M.}},
\oauthor{\bsnm{Gogineni}, \binits{A.}},
\oauthor{\bsnm{Ramakrishnan}, \binits{N.}}:
{Innovations in Neural Data-to-text Generation},
1--23
(2022)
{\href{https://arxiv.org/abs/2207.12571}{{arXiv:2207.12571}}}
\end{botherref}
\endbibitem

\bibitem[\protect\citeauthoryear{Gardent et~al.}{2017}]{Gardent2017}
\begin{barticle}
\bauthor{\bsnm{Gardent}, \binits{C.}},
\bauthor{\bsnm{Shimorina}, \binits{A.}},
\bauthor{\bsnm{Narayan}, \binits{S.}},
\bauthor{\bsnm{Perez-Beltrachini}, \binits{L.}}:
\batitle{{The WebNLG challenge: Generating text from RDF data}}.
\bjtitle{INLG 2017 - 10th International Natural Language Generation Conference, Proceedings of the Conference}
\bvolume{298},
\bfpage{124}--\blpage{133}
(\byear{2017})
\doiurl{10.18653/v1/w17-3518}
\end{barticle}
\endbibitem

\bibitem[\protect\citeauthoryear{Ribeiro et~al.}{2021}]{Ribeiro2021b}
\begin{botherref}
\oauthor{\bsnm{Ribeiro}, \binits{L.F.R.}},
\oauthor{\bsnm{Schmitt}, \binits{M.}},
\oauthor{\bsnm{Sch{\"{u}}tze}, \binits{H.}},
\oauthor{\bsnm{Gurevych}, \binits{I.}}:
{Investigating Pretrained Language Models for Graph-to-Text Generation},
211--227
(2021)
\doiurl{10.18653/v1/2021.nlp4convai-1.20}
{\href{https://arxiv.org/abs/2007.08426}{{arXiv:2007.08426}}}
\end{botherref}
\endbibitem

\bibitem[\protect\citeauthoryear{Zhang et~al.}{2020}]{Zhang2020}
\begin{botherref}
\oauthor{\bsnm{Zhang}, \binits{Y.}},
\oauthor{\bsnm{Guo}, \binits{Z.}},
\oauthor{\bsnm{Teng}, \binits{Z.}},
\oauthor{\bsnm{Lu}, \binits{W.}},
\oauthor{\bsnm{Cohen}, \binits{S.B.}},
\oauthor{\bsnm{Liu}, \binits{Z.}},
\oauthor{\bsnm{Bing}, \binits{L.}}:
{Lightweight, dynamic graph convolutional networks for AMR-to-text generation}.
EMNLP 2020 - 2020 Conference on Empirical Methods in Natural Language Processing, Proceedings of the Conference,
2162--2172
(2020)
\doiurl{10.18653/v1/2020.emnlp-main.169}
{\href{https://arxiv.org/abs/2010.04383}{{arXiv:2010.04383}}}
\end{botherref}
\endbibitem

\bibitem[\protect\citeauthoryear{Hajdik et~al.}{2019}]{Hajdik2019}
\begin{barticle}
\bauthor{\bsnm{Hajdik}, \binits{V.}},
\bauthor{\bsnm{Buys}, \binits{J.}},
\bauthor{\bsnm{Goodman}, \binits{M.W.}},
\bauthor{\bsnm{Bender}, \binits{E.M.}}:
\batitle{{Neural text generation from rich semantic representations}}.
\bjtitle{NAACL HLT 2019 - 2019 Conference of the North American Chapter of the Association for Computational Linguistics: Human Language Technologies - Proceedings of the Conference}
\bvolume{1},
\bfpage{2259}--\blpage{2266}
(\byear{2019})
\doiurl{10.18653/v1/n19-1235}
{\href{https://arxiv.org/abs/1904.11564}{{arXiv:1904.11564}}}
\end{barticle}
\endbibitem

\bibitem[\protect\citeauthoryear{Page et~al.}{2021}]{Page2021}
\begin{botherref}
\oauthor{\bsnm{Page}, \binits{M.J.}},
\oauthor{\bsnm{McKenzie}, \binits{J.E.}},
\oauthor{\bsnm{Bossuyt}, \binits{P.M.}},
\oauthor{\bsnm{Boutron}, \binits{I.}},
\oauthor{\bsnm{Hoffmann}, \binits{T.C.}},
\oauthor{\bsnm{Mulrow}, \binits{C.D.}},
\oauthor{\bsnm{Shamseer}, \binits{L.}},
\oauthor{\bsnm{Tetzlaff}, \binits{J.M.}},
\oauthor{\bsnm{Akl}, \binits{E.A.}},
\oauthor{\bsnm{Brennan}, \binits{S.E.}},
\oauthor{\bsnm{Chou}, \binits{R.}},
\oauthor{\bsnm{Glanville}, \binits{J.}},
\oauthor{\bsnm{Grimshaw}, \binits{J.M.}},
\oauthor{\bsnm{Hr{\'{o}}bjartsson}, \binits{A.}},
\oauthor{\bsnm{Lalu}, \binits{M.M.}},
\oauthor{\bsnm{Li}, \binits{T.}},
\oauthor{\bsnm{Loder}, \binits{E.W.}},
\oauthor{\bsnm{Mayo-Wilson}, \binits{E.}},
\oauthor{\bsnm{McDonald}, \binits{S.}},
\oauthor{\bsnm{McGuinness}, \binits{L.A.}},
\oauthor{\bsnm{Stewart}, \binits{L.A.}},
\oauthor{\bsnm{Thomas}, \binits{J.}},
\oauthor{\bsnm{Tricco}, \binits{A.C.}},
\oauthor{\bsnm{Welch}, \binits{V.A.}},
\oauthor{\bsnm{Whiting}, \binits{P.}},
\oauthor{\bsnm{Moher}, \binits{D.}}:
{The PRISMA 2020 statement: An updated guideline for reporting systematic reviews}.
The BMJ
\textbf{372}
(2021)
\doiurl{10.1136/bmj.n71}
\end{botherref}
\endbibitem

\bibitem[\protect\citeauthoryear{Fillippova}{2020}]{Fillippova2020}
\begin{botherref}
\oauthor{\bsnm{Fillippova}, \binits{K.}}:
{Controlled hallucinations: learning to generate faithfully from noisy data}.
Findings of the Association for Computational Linguistics Findings of ACL: EMNLP 2020,
864--870
(2020)
\doiurl{10.18653/v1/2020.findings-emnlp.76}
{\href{https://arxiv.org/abs/2010.05873}{{arXiv:2010.05873}}}
\end{botherref}
\endbibitem

\bibitem[\protect\citeauthoryear{Lin et~al.}{2020}]{Lin2020}
\begin{botherref}
\oauthor{\bsnm{Lin}, \binits{S.}},
\oauthor{\bsnm{Wang}, \binits{W.}},
\oauthor{\bsnm{Yang}, \binits{Z.}},
\oauthor{\bsnm{Liang}, \binits{X.}},
\oauthor{\bsnm{Xu}, \binits{F.F.}},
\oauthor{\bsnm{Xing}, \binits{E.P.}},
\oauthor{\bsnm{Hu}, \binits{Z.}}:
{Data-to-text generation with style imitation}.
Findings of the Association for Computational Linguistics Findings of ACL: EMNLP 2020
(Figure 1),
1589--1598
(2020)
\doiurl{10.18653/v1/2020.findings-emnlp.144}
{\href{https://arxiv.org/abs/1901.09501}{{arXiv:1901.09501}}}
\end{botherref}
\endbibitem

\bibitem[\protect\citeauthoryear{Gong et~al.}{2020}]{Gong2020}
\begin{botherref}
\oauthor{\bsnm{Gong}, \binits{H.}},
\oauthor{\bsnm{Bi}, \binits{W.}},
\oauthor{\bsnm{Feng}, \binits{X.}},
\oauthor{\bsnm{Qin}, \binits{B.}},
\oauthor{\bsnm{Liu}, \binits{X.}},
\oauthor{\bsnm{Liu}, \binits{T.}}:
{Enhancing content planning for table-to-text generation with data understanding and verification}.
Findings of the Association for Computational Linguistics Findings of ACL: EMNLP 2020,
905--2914
(2020)
\doiurl{10.18653/v1/2020.findings-emnlp.262}
\end{botherref}
\endbibitem

\bibitem[\protect\citeauthoryear{Garneau and Lamontagne}{2021}]{Garneau2021}
\begin{botherref}
\oauthor{\bsnm{Garneau}, \binits{N.}},
\oauthor{\bsnm{Lamontagne}, \binits{L.}}:
{Trainable Ranking Models to Evaluate the Semantic Accuracy of Data-to-Text Neural Generator}.
Eval4NLP 2021 - Evaluation and Comparison of NLP Systems, Proceedings of the 2nd Workshop,
51--61
(2021)
\doiurl{10.26615/978-954-452-056-4_006}
\end{botherref}
\endbibitem

\bibitem[\protect\citeauthoryear{Su et~al.}{2021}]{Su2021}
\begin{botherref}
\oauthor{\bsnm{Su}, \binits{Y.}},
\oauthor{\bsnm{Meng}, \binits{Z.}},
\oauthor{\bsnm{Baker}, \binits{S.}},
\oauthor{\bsnm{Collier}, \binits{N.}}:
{Few-Shot Table-to-Text Generation with Prototype Memory}.
Findings of the Association for Computational Linguistics, Findings of ACL: EMNLP 2021,
910--917
(2021)
\doiurl{10.18653/v1/2021.findings-emnlp.77}
{\href{https://arxiv.org/abs/2108.12516}{{arXiv:2108.12516}}}
\end{botherref}
\endbibitem

\bibitem[\protect\citeauthoryear{Wiseman et~al.}{2017}]{Wiseman2017}
\begin{botherref}
\oauthor{\bsnm{Wiseman}, \binits{S.}},
\oauthor{\bsnm{Shieber}, \binits{S.M.}},
\oauthor{\bsnm{Rush}, \binits{A.M.}}:
{Challenges in Data-to-Document Generation},
2253--2263
(2017)
\end{botherref}
\endbibitem

\bibitem[\protect\citeauthoryear{Nie et~al.}{2018}]{Nie2018}
\begin{botherref}
\oauthor{\bsnm{Nie}, \binits{F.}},
\oauthor{\bsnm{Wang}, \binits{J.}},
\oauthor{\bsnm{Yao}, \binits{J.G.}},
\oauthor{\bsnm{Pan}, \binits{R.}},
\oauthor{\bsnm{Lin}, \binits{C.Y.}}:
{Operation-guided neural networks for high fidelity data-to-text generation}.
Proceedings of the 2018 Conference on Empirical Methods in Natural Language Processing, EMNLP 2018,
3879--3889
(2018)
\doiurl{10.18653/v1/d18-1422}
{\href{https://arxiv.org/abs/1809.02735}{{arXiv:1809.02735}}}
\end{botherref}
\endbibitem

\bibitem[\protect\citeauthoryear{Freitag and Roy}{2018}]{Freitag2018}
\begin{botherref}
\oauthor{\bsnm{Freitag}, \binits{M.}},
\oauthor{\bsnm{Roy}, \binits{S.}}:
{Unsupervised natural language generation with denoising autoencoders}.
Proceedings of the 2018 Conference on Empirical Methods in Natural Language Processing, EMNLP 2018
(2014),
3922--3929
(2018)
\doiurl{10.18653/v1/d18-1426}
{\href{https://arxiv.org/abs/1804.07899}{{arXiv:1804.07899}}}
\end{botherref}
\endbibitem

\bibitem[\protect\citeauthoryear{Ferreira et~al.}{2019}]{Ferreira2019}
\begin{botherref}
\oauthor{\bsnm{Ferreira}, \binits{T.C.}},
\oauthor{\bsnm{Lee}, \binits{C.}},
\oauthor{\bsnm{Miltenburg}, \binits{E.}},
\oauthor{\bsnm{Krahmer}, \binits{E.}}:
{Neural data-to-text generation: A comparison between pipeline and end-to-end architectures}.
EMNLP-IJCNLP 2019 - 2019 Conference on Empirical Methods in Natural Language Processing and 9th International Joint Conference on Natural Language Processing, Proceedings of the Conference,
552--562
(2019)
\doiurl{10.18653/v1/d19-1052}
{\href{https://arxiv.org/abs/1908.09022}{{arXiv:1908.09022}}}
\end{botherref}
\endbibitem

\bibitem[\protect\citeauthoryear{Shao et~al.}{2019}]{Shao2019}
\begin{botherref}
\oauthor{\bsnm{Shao}, \binits{Z.}},
\oauthor{\bsnm{Huang}, \binits{M.}},
\oauthor{\bsnm{Wen}, \binits{J.}},
\oauthor{\bsnm{Xu}, \binits{W.}},
\oauthor{\bsnm{Zhu}, \binits{X.}}:
{Long and diverse text generation with planning-based hierarchical variational model}.
EMNLP-IJCNLP 2019 - 2019 Conference on Empirical Methods in Natural Language Processing and 9th International Joint Conference on Natural Language Processing, Proceedings of the Conference,
3257--3268
(2019)
\doiurl{10.18653/v1/d19-1321}
{\href{https://arxiv.org/abs/1908.06605}{{arXiv:1908.06605}}}
\end{botherref}
\endbibitem

\bibitem[\protect\citeauthoryear{Ribeiro et~al.}{2019}]{Ribeiro2019}
\begin{botherref}
\oauthor{\bsnm{Ribeiro}, \binits{L.F.R.}},
\oauthor{\bsnm{Gardent}, \binits{C.}},
\oauthor{\bsnm{Gurevych}, \binits{I.}}:
{Enhancing AMR-to-text generation with dual graph representations}.
EMNLP-IJCNLP 2019 - 2019 Conference on Empirical Methods in Natural Language Processing and 9th International Joint Conference on Natural Language Processing, Proceedings of the Conference,
3183--3194
(2019)
\doiurl{10.18653/v1/d19-1314}
{\href{https://arxiv.org/abs/1909.00352}{{arXiv:1909.00352}}}
\end{botherref}
\endbibitem

\bibitem[\protect\citeauthoryear{Gong et~al.}{2019}]{Gong2019}
\begin{botherref}
\oauthor{\bsnm{Gong}, \binits{H.}},
\oauthor{\bsnm{Feng}, \binits{X.}},
\oauthor{\bsnm{Qin}, \binits{B.}},
\oauthor{\bsnm{Liu}, \binits{T.}}:
{Table-to-text generation with effective hierarchical encoder on three dimensions (row, column and time)}.
EMNLP-IJCNLP 2019 - 2019 Conference on Empirical Methods in Natural Language Processing and 9th International Joint Conference on Natural Language Processing, Proceedings of the Conference,
3143--3152
(2019)
\doiurl{10.18653/v1/d19-1310}
{\href{https://arxiv.org/abs/1909.02304}{{arXiv:1909.02304}}}
\end{botherref}
\endbibitem

\bibitem[\protect\citeauthoryear{Chen et~al.}{2019}]{Chen2019}
\begin{botherref}
\oauthor{\bsnm{Chen}, \binits{S.}},
\oauthor{\bsnm{Wang}, \binits{J.}},
\oauthor{\bsnm{Feng}, \binits{X.}},
\oauthor{\bsnm{Jiang}, \binits{F.}},
\oauthor{\bsnm{Qin}, \binits{B.}},
\oauthor{\bsnm{Lin}, \binits{C.Y.}}:
{Enhancing neural data-to-text generation models with external background knowledge}.
EMNLP-IJCNLP 2019 - 2019 Conference on Empirical Methods in Natural Language Processing and 9th International Joint Conference on Natural Language Processing, Proceedings of the Conference,
3022--3032
(2019)
\doiurl{10.18653/v1/d19-1299}
\end{botherref}
\endbibitem

\bibitem[\protect\citeauthoryear{Chen et~al.}{2020}]{Chen2020b}
\begin{botherref}
\oauthor{\bsnm{Chen}, \binits{W.}},
\oauthor{\bsnm{Su}, \binits{Y.}},
\oauthor{\bsnm{Yan}, \binits{X.}},
\oauthor{\bsnm{Wang}, \binits{W.Y.}}:
{KGPT : Knowledge-Grounded Pre-Training for Data-to-Text Generation},
8635--8648
(2020)
\end{botherref}
\endbibitem

\bibitem[\protect\citeauthoryear{Fan et~al.}{2020}]{Fan2020}
\begin{botherref}
\oauthor{\bsnm{Fan}, \binits{A.}},
\oauthor{\bsnm{Loria}, \binits{F.}},
\oauthor{\bsnm{Lorraine}, \binits{D.}},
\oauthor{\bsnm{Gardent}, \binits{C.}},
\oauthor{\bsnm{Loria}, \binits{C.}}:
{Multilingual AMR-to-Text Generation}.
arXiv preprint arXiv:2011.05443,
2889--2901
(2020)
\end{botherref}
\endbibitem

\bibitem[\protect\citeauthoryear{Bai et~al.}{2020}]{Bai2020}
\begin{botherref}
\oauthor{\bsnm{Bai}, \binits{X.}},
\oauthor{\bsnm{Song}, \binits{L.}},
\oauthor{\bsnm{Zhang}, \binits{Y.}}:
{Online Back-Parsing for AMR-to-Text Generation},
1206--1219
(2020)
\end{botherref}
\endbibitem

\bibitem[\protect\citeauthoryear{Fu et~al.}{2020}]{Fu2020}
\begin{botherref}
\oauthor{\bsnm{Fu}, \binits{Z.}},
\oauthor{\bsnm{Shi}, \binits{B.}},
\oauthor{\bsnm{Lam}, \binits{W.}},
\oauthor{\bsnm{Bing}, \binits{L.}},
\oauthor{\bsnm{Liu}, \binits{Z.}}:
{Partially-aligned data-to-text generation with distant supervision}.
EMNLP 2020 - 2020 Conference on Empirical Methods in Natural Language Processing, Proceedings of the Conference,
9183--9193
(2020)
\doiurl{10.18653/v1/2020.emnlp-main.738}
{\href{https://arxiv.org/abs/2010.01268}{{arXiv:2010.01268}}}
\end{botherref}
\endbibitem

\bibitem[\protect\citeauthoryear{Kedzie and McKeown}{2020}]{Kedzie2020}
\begin{botherref}
\oauthor{\bsnm{Kedzie}, \binits{C.}},
\oauthor{\bsnm{McKeown}, \binits{K.}}:
{Controllable meaning representation to text generation: Linearization and data augmentation strategies}.
EMNLP 2020 - 2020 Conference on Empirical Methods in Natural Language Processing, Proceedings of the Conference,
5160--5185
(2020)
\doiurl{10.18653/v1/2020.emnlp-main.419}
\end{botherref}
\endbibitem

\bibitem[\protect\citeauthoryear{Ribeiro et~al.}{2021}]{Ribeiro2021}
\begin{botherref}
\oauthor{\bsnm{Ribeiro}, \binits{L.F.R.}},
\oauthor{\bsnm{Zhang}, \binits{Y.}},
\oauthor{\bsnm{Gurevych}, \binits{I.}}:
{Structural Adapters in Pretrained Language Models for AMR-to-Text Generation}.
EMNLP 2021 - 2021 Conference on Empirical Methods in Natural Language Processing, Proceedings,
4269--4282
(2021)
\doiurl{10.18653/v1/2021.emnlp-main.351}
{\href{https://arxiv.org/abs/2103.09120}{{arXiv:2103.09120}}}
\end{botherref}
\endbibitem

\bibitem[\protect\citeauthoryear{Wiseman et~al.}{2021}]{Wiseman2021}
\begin{botherref}
\oauthor{\bsnm{Wiseman}, \binits{S.}},
\oauthor{\bsnm{Backurs}, \binits{A.}},
\oauthor{\bsnm{Stratos}, \binits{K.}}:
{Data-to-text Generation by Splicing Together Nearest Neighbors}.
EMNLP 2021 - 2021 Conference on Empirical Methods in Natural Language Processing, Proceedings,
4283--4299
(2021)
\doiurl{10.18653/v1/2021.emnlp-main.352}
{\href{https://arxiv.org/abs/2101.08248}{{arXiv:2101.08248}}}
\end{botherref}
\endbibitem

\bibitem[\protect\citeauthoryear{Chang et~al.}{2021}]{Chang2021}
\begin{botherref}
\oauthor{\bsnm{Chang}, \binits{E.}},
\oauthor{\bsnm{Shen}, \binits{X.}},
\oauthor{\bsnm{Zhu}, \binits{D.}},
\oauthor{\bsnm{Demberg}, \binits{V.}},
\oauthor{\bsnm{Su}, \binits{H.}}:
{Neural data-to-text generation with LM-based text augmentation}.
EACL 2021 - 16th Conference of the European Chapter of the Association for Computational Linguistics, Proceedings of the Conference
(1),
758--768
(2021)
\doiurl{10.18653/v1/2021.eacl-main.64}
{\href{https://arxiv.org/abs/2102.03556}{{arXiv:2102.03556}}}
\end{botherref}
\endbibitem

\bibitem[\protect\citeauthoryear{Opitz and Frank}{2021}]{Opitz2021}
\begin{botherref}
\oauthor{\bsnm{Opitz}, \binits{J.}},
\oauthor{\bsnm{Frank}, \binits{A.}}:
{Towards a decomposable metric for explainable evaluation of text generation from AMR}.
EACL 2021 - 16th Conference of the European Chapter of the Association for Computational Linguistics, Proceedings of the Conference
(i),
1504--1518
(2021)
\doiurl{10.18653/v1/2021.eacl-main.129}
{\href{https://arxiv.org/abs/2008.08896}{{arXiv:2008.08896}}}
\end{botherref}
\endbibitem

\bibitem[\protect\citeauthoryear{Hargreaves et~al.}{2021}]{Hargreaves2021}
\begin{botherref}
\oauthor{\bsnm{Hargreaves}, \binits{J.}},
\oauthor{\bsnm{Vlachos}, \binits{A.}},
\oauthor{\bsnm{Emerson}, \binits{G.}}:
{Incremental beam manipulation for natural language generation}.
EACL 2021 - 16th Conference of the European Chapter of the Association for Computational Linguistics, Proceedings of the Conference,
2563--2574
(2021)
\doiurl{10.18653/v1/2021.eacl-main.219}
{\href{https://arxiv.org/abs/2102.02574}{{arXiv:2102.02574}}}
\end{botherref}
\endbibitem

\bibitem[\protect\citeauthoryear{Konstas et~al.}{2017}]{Konstas2017}
\begin{barticle}
\bauthor{\bsnm{Konstas}, \binits{I.}},
\bauthor{\bsnm{Iyer}, \binits{S.}},
\bauthor{\bsnm{Yatskar}, \binits{M.}},
\bauthor{\bsnm{Choi}, \binits{Y.}},
\bauthor{\bsnm{Zettlemoyer}, \binits{L.}}:
\batitle{{Neural AMR: Sequence-to-sequence models for parsing and generation}}.
\bjtitle{ACL 2017 - 55th Annual Meeting of the Association for Computational Linguistics, Proceedings of the Conference (Long Papers)}
\bvolume{1},
\bfpage{146}--\blpage{157}
(\byear{2017})
\doiurl{10.18653/v1/P17-1014}
{\href{https://arxiv.org/abs/1704.08381}{{arXiv:1704.08381}}}
\end{barticle}
\endbibitem

\bibitem[\protect\citeauthoryear{Song et~al.}{2018}]{Song2018}
\begin{botherref}
\oauthor{\bsnm{Song}, \binits{L.}},
\oauthor{\bsnm{Zhang}, \binits{Y.}},
\oauthor{\bsnm{Wang}, \binits{Z.}},
\oauthor{\bsnm{Gildea}, \binits{D.}},
\oauthor{\bsnm{Science}, \binits{C.}}:
{A Graph-to-Sequence Model for AMR-to-Text Generation},
1616--1626
(2018)
\end{botherref}
\endbibitem

\bibitem[\protect\citeauthoryear{Dhingra et~al.}{2020}]{Dhingra}
\begin{botherref}
\oauthor{\bsnm{Dhingra}, \binits{B.}},
\oauthor{\bsnm{Faruqui}, \binits{M.}},
\oauthor{\bsnm{Parikh}, \binits{A.}},
\oauthor{\bsnm{Chang}, \binits{M.W.}},
\oauthor{\bsnm{Das}, \binits{D.}},
\oauthor{\bsnm{Cohen}, \binits{W.W.}}:
Handling divergent reference texts when evaluating table-to-text generation.
ACL 2019 - 57th Annual Meeting of the Association for Computational Linguistics, Proceedings of the Conference,
4884--4895
(2020)
\doiurl{10.18653/v1/p19-1483}
\end{botherref}
\endbibitem

\bibitem[\protect\citeauthoryear{Puduppully et~al.}{2019}]{Puduppully2020}
\begin{botherref}
\oauthor{\bsnm{Puduppully}, \binits{R.}},
\oauthor{\bsnm{Dong}, \binits{L.}},
\oauthor{\bsnm{Lapata}, \binits{M.}}:
{Data-to-text generation with entity modeling}.
ACL 2019 - 57th Annual Meeting of the Association for Computational Linguistics, Proceedings of the Conference,
2023--2035
(2019)
\doiurl{10.18653/v1/p19-1195}
{\href{https://arxiv.org/abs/1906.03221}{{arXiv:1906.03221}}}
\end{botherref}
\endbibitem

\bibitem[\protect\citeauthoryear{Nie et~al.}{2020}]{Nie2020}
\begin{botherref}
\oauthor{\bsnm{Nie}, \binits{F.}},
\oauthor{\bsnm{Yao}, \binits{J.G.}},
\oauthor{\bsnm{Wang}, \binits{J.}},
\oauthor{\bsnm{Pan}, \binits{R.}},
\oauthor{\bsnm{Lin}, \binits{C.Y.}}:
{A simple recipe towards reducing hallucination in neural surface realisation}.
ACL 2019 - 57th Annual Meeting of the Association for Computational Linguistics, Proceedings of the Conference
(2),
2673--2679
(2020)
\doiurl{10.18653/v1/p19-1256}
\end{botherref}
\endbibitem

\bibitem[\protect\citeauthoryear{Iso et~al.}{2020}]{Iso2020}
\begin{botherref}
\oauthor{\bsnm{Iso}, \binits{H.}},
\oauthor{\bsnm{Uehara}, \binits{Y.}},
\oauthor{\bsnm{Ishigaki}, \binits{T.}},
\oauthor{\bsnm{Noji}, \binits{H.}},
\oauthor{\bsnm{Aramaki}, \binits{E.}},
\oauthor{\bsnm{Kobayashi}, \binits{I.}},
\oauthor{\bsnm{Miyao}, \binits{Y.}},
\oauthor{\bsnm{Okazaki}, \binits{N.}},
\oauthor{\bsnm{Takamura}, \binits{H.}}:
{Learning to select, track, and generate for data-to-text}.
ACL 2019 - 57th Annual Meeting of the Association for Computational Linguistics, Proceedings of the Conference,
2102--2113
(2020)
\doiurl{10.5715/jnlp.27.599}
{\href{https://arxiv.org/abs/1907.09699}{{arXiv:1907.09699}}}
\end{botherref}
\endbibitem

\bibitem[\protect\citeauthoryear{Chen et~al.}{2020}]{Chen2020a}
\begin{barticle}
\bauthor{\bsnm{Chen}, \binits{W.}},
\bauthor{\bsnm{Chen}, \binits{J.}},
\bauthor{\bsnm{Su}, \binits{Y.}},
\bauthor{\bsnm{Chen}, \binits{Z.}},
\bauthor{\bsnm{Wang}, \binits{W.Y.}}:
\batitle{{Logical natural language generation from open-domain tables}}.
\bjtitle{Proceedings of the Annual Meeting of the Association for Computational Linguistics}
\bvolume{2},
\bfpage{7929}--\blpage{7942}
(\byear{2020})
\doiurl{10.18653/v1/2020.acl-main.708}
{\href{https://arxiv.org/abs/2004.10404}{{arXiv:2004.10404}}}
\end{barticle}
\endbibitem

\bibitem[\protect\citeauthoryear{Zhao et~al.}{2020}]{Zhao2020}
\begin{botherref}
\oauthor{\bsnm{Zhao}, \binits{C.}},
\oauthor{\bsnm{Walker}, \binits{M.}},
\oauthor{\bsnm{Chaturvedi}, \binits{S.}}:
{Bridging the structural gap between encoding and decoding for data-to-text generation}.
Proceedings of the Annual Meeting of the Association for Computational Linguistics,
2481--2491
(2020)
\doiurl{10.18653/v1/2020.acl-main.224}
\end{botherref}
\endbibitem

\bibitem[\protect\citeauthoryear{Ram et~al.}{2020}]{Ram2020}
\begin{botherref}
\oauthor{\bsnm{Ram}, \binits{M.M.}},
\oauthor{\bsnm{Radu}, \binits{Y.-s.L.}},
\oauthor{\bsnm{Salim}, \binits{F.}}:
{GPT-too : A Language-Model-First Approach for AMR-to-Text Generation},
1846--1852
(2020)
\end{botherref}
\endbibitem

\bibitem[\protect\citeauthoryear{Wang et~al.}{2020}]{Wang2020a}
\begin{botherref}
\oauthor{\bsnm{Wang}, \binits{Z.}},
\oauthor{\bsnm{Wang}, \binits{X.}},
\oauthor{\bsnm{An}, \binits{B.}},
\oauthor{\bsnm{Yu}, \binits{D.}},
\oauthor{\bsnm{Chen}, \binits{C.}}:
{Towards faithful neural table-to-text generation with content-matching constraints}.
Proceedings of the Annual Meeting of the Association for Computational Linguistics,
1072--1086
(2020)
\doiurl{10.18653/v1/2020.acl-main.101}
{\href{https://arxiv.org/abs/2005.00969}{{arXiv:2005.00969}}}
\end{botherref}
\endbibitem

\bibitem[\protect\citeauthoryear{Shen et~al.}{2020}]{Shen2020}
\begin{botherref}
\oauthor{\bsnm{Shen}, \binits{X.}},
\oauthor{\bsnm{Chang}, \binits{E.}},
\oauthor{\bsnm{Su}, \binits{H.}},
\oauthor{\bsnm{Niu}, \binits{C.}},
\oauthor{\bsnm{Klakow}, \binits{D.}}:
{Neural data-to-text generation via jointly learning the segmentation and correspondence}.
Proceedings of the Annual Meeting of the Association for Computational Linguistics
(2019),
7155--7165
(2020)
\doiurl{10.18653/v1/2020.acl-main.641}
{\href{https://arxiv.org/abs/2005.01096}{{arXiv:2005.01096}}}
\end{botherref}
\endbibitem

\bibitem[\protect\citeauthoryear{Iso et~al.}{2020}]{Iso2020a}
\begin{botherref}
\oauthor{\bsnm{Iso}, \binits{H.}},
\oauthor{\bsnm{Qiao}, \binits{C.}},
\oauthor{\bsnm{Li}, \binits{H.}}:
{Fact-based text editing}.
Proceedings of the Annual Meeting of the Association for Computational Linguistics,
171--182
(2020)
\doiurl{10.18653/v1/2020.acl-main.17}
{\href{https://arxiv.org/abs/2007.00916}{{arXiv:2007.00916}}}
\end{botherref}
\endbibitem

\bibitem[\protect\citeauthoryear{Suadaa et~al.}{2021}]{Suadaa2021}
\begin{botherref}
\oauthor{\bsnm{Suadaa}, \binits{L.H.}},
\oauthor{\bsnm{Kamigaito}, \binits{H.}},
\oauthor{\bsnm{Funakoshi}, \binits{K.}},
\oauthor{\bsnm{Okumura}, \binits{M.}},
\oauthor{\bsnm{Takamura}, \binits{H.}}:
{Towards table-to-text generation with numerical reasoning}.
ACL-IJCNLP 2021 - 59th Annual Meeting of the Association for Computational Linguistics and the 11th International Joint Conference on Natural Language Processing, Proceedings of the Conference,
1451--1465
(2021)
\doiurl{10.18653/v1/2021.acl-long.115}
\end{botherref}
\endbibitem

\bibitem[\protect\citeauthoryear{Chang et~al.}{2021}]{Chang2021a}
\begin{barticle}
\bauthor{\bsnm{Chang}, \binits{E.}},
\bauthor{\bsnm{Shen}, \binits{X.}},
\bauthor{\bsnm{Yeh}, \binits{H.S.}},
\bauthor{\bsnm{Demberg}, \binits{V.}}:
\batitle{{On Training Instance Selection for Few-Shot Neural Text Generation}}.
\bjtitle{ACL-IJCNLP 2021 - 59th Annual Meeting of the Association for Computational Linguistics and the 11th International Joint Conference on Natural Language Processing, Proceedings of the Conference}
\bvolume{2},
\bfpage{8}--\blpage{13}
(\byear{2021})
\doiurl{10.18653/v1/2021.acl-short.2}
{\href{https://arxiv.org/abs/2107.03176}{{arXiv:2107.03176}}}
\end{barticle}
\endbibitem

\bibitem[\protect\citeauthoryear{Wang et~al.}{2021}]{Wang2021}
\begin{botherref}
\oauthor{\bsnm{Wang}, \binits{Y.}},
\oauthor{\bsnm{Wood}, \binits{I.D.}},
\oauthor{\bsnm{Wan}, \binits{S.}},
\oauthor{\bsnm{Dras}, \binits{M.}},
\oauthor{\bsnm{Johnson}, \binits{M.}}:
{Mention flags (MF): Constraining transformer-based text generators}.
ACL-IJCNLP 2021 - 59th Annual Meeting of the Association for Computational Linguistics and the 11th International Joint Conference on Natural Language Processing, Proceedings of the Conference,
103--113
(2021)
\doiurl{10.18653/v1/2021.acl-long.9}
\end{botherref}
\endbibitem

\bibitem[\protect\citeauthoryear{Xu et~al.}{2021}]{Xu2021a}
\begin{botherref}
\oauthor{\bsnm{Xu}, \binits{D.}},
\oauthor{\bsnm{Li}, \binits{J.}},
\oauthor{\bsnm{Zhu}, \binits{M.}},
\oauthor{\bsnm{Zhang}, \binits{M.}},
\oauthor{\bsnm{Zhou}, \binits{G.}}:
{XLPT-AMR: Cross-lingual pre-training via multi-task learning for zero-shot AMR parsing and text generation}.
ACL-IJCNLP 2021 - 59th Annual Meeting of the Association for Computational Linguistics and the 11th International Joint Conference on Natural Language Processing, Proceedings of the Conference,
896--907
(2021)
\doiurl{10.18653/v1/2021.acl-long.73}
\end{botherref}
\endbibitem

\bibitem[\protect\citeauthoryear{Li et~al.}{2021}]{Li2021}
\begin{botherref}
\oauthor{\bsnm{Li}, \binits{L.}},
\oauthor{\bsnm{Ma}, \binits{C.}},
\oauthor{\bsnm{Yue}, \binits{Y.}},
\oauthor{\bsnm{Hu}, \binits{D.}}:
{Improving encoder by auxiliary supervision tasks for table-to-text generation}.
ACL-IJCNLP 2021 - 59th Annual Meeting of the Association for Computational Linguistics and the 11th International Joint Conference on Natural Language Processing, Proceedings of the Conference,
5979--5989
(2021)
\doiurl{10.18653/v1/2021.acl-long.466}
\end{botherref}
\endbibitem

\bibitem[\protect\citeauthoryear{Bai et~al.}{2022}]{Bai2022}
\begin{barticle}
\bauthor{\bsnm{Bai}, \binits{X.}},
\bauthor{\bsnm{Chen}, \binits{Y.}},
\bauthor{\bsnm{Zhang}, \binits{Y.}}:
\batitle{{Graph Pre-training for AMR Parsing and Generation}}.
\bjtitle{Proceedings of the Annual Meeting of the Association for Computational Linguistics}
\bvolume{1},
\bfpage{6001}--\blpage{6015}
(\byear{2022})
\doiurl{10.18653/v1/2022.acl-long.415}
{\href{https://arxiv.org/abs/2203.07836}{{arXiv:2203.07836}}}
\end{barticle}
\endbibitem

\bibitem[\protect\citeauthoryear{Perez-Beltrachini and Lapata}{2018}]{Perez-Beltrachini2018}
\begin{barticle}
\bauthor{\bsnm{Perez-Beltrachini}, \binits{L.}},
\bauthor{\bsnm{Lapata}, \binits{M.}}:
\batitle{{Bootstrapping Generators from Noisy Data}}.
\bjtitle{NAACL HLT 2018 - 2018 Conference of the North American Chapter of the Association for Computational Linguistics: Human Language Technologies - Proceedings of the Conference}
\bvolume{1},
\bfpage{1516}--\blpage{1527}
(\byear{2018})
\doiurl{10.18653/V1/N18-1137}
{\href{https://arxiv.org/abs/1804.06385}{{arXiv:1804.06385}}}
\end{barticle}
\endbibitem

\bibitem[\protect\citeauthoryear{Moryossef et~al.}{2019}]{Moryossef2019}
\begin{barticle}
\bauthor{\bsnm{Moryossef}, \binits{A.}},
\bauthor{\bsnm{Goldberg}, \binits{Y.}},
\bauthor{\bsnm{Dagan}, \binits{I.}}:
\batitle{{Step-by-step: Separating planning from realization in neural data-to-text generation}}.
\bjtitle{NAACL HLT 2019 - 2019 Conference of the North American Chapter of the Association for Computational Linguistics: Human Language Technologies - Proceedings of the Conference}
\bvolume{1},
\bfpage{2267}--\blpage{2277}
(\byear{2019})
{\href{https://arxiv.org/abs/1904.03396}{{arXiv:1904.03396}}}
\end{barticle}
\endbibitem

\bibitem[\protect\citeauthoryear{Damonte and Cohen}{2019}]{Damonte2019}
\begin{barticle}
\bauthor{\bsnm{Damonte}, \binits{M.}},
\bauthor{\bsnm{Cohen}, \binits{S.B.}}:
\batitle{{Structural neural encoders for AMR-to-text generation}}.
\bjtitle{NAACL HLT 2019 - 2019 Conference of the North American Chapter of the Association for Computational Linguistics: Human Language Technologies - Proceedings of the Conference}
\bvolume{1},
\bfpage{3649}--\blpage{3658}
(\byear{2019})
{\href{https://arxiv.org/abs/1903.11410}{{arXiv:1903.11410}}}
\end{barticle}
\endbibitem

\bibitem[\protect\citeauthoryear{Nan et~al.}{2021}]{Nan2021}
\begin{botherref}
\oauthor{\bsnm{Nan}, \binits{L.}},
\oauthor{\bsnm{Radev}, \binits{D.}},
\oauthor{\bsnm{Zhang}, \binits{R.}},
\oauthor{\bsnm{Rau}, \binits{A.}},
\oauthor{\bsnm{Sivaprasad}, \binits{A.}},
\oauthor{\bsnm{Hsieh}, \binits{C.}},
\oauthor{\bsnm{Tang}, \binits{X.}},
\oauthor{\bsnm{Vyas}, \binits{A.}},
\oauthor{\bsnm{Verma}, \binits{N.}},
\oauthor{\bsnm{Krishna}, \binits{P.}},
\oauthor{\bsnm{Liu}, \binits{Y.}},
\oauthor{\bsnm{Irwanto}, \binits{N.}},
\oauthor{\bsnm{Pan}, \binits{J.}},
\oauthor{\bsnm{Rahman}, \binits{F.}},
\oauthor{\bsnm{Zaidi}, \binits{A.}},
\oauthor{\bsnm{Mutuma}, \binits{M.}},
\oauthor{\bsnm{Tarabar}, \binits{Y.}},
\oauthor{\bsnm{Gupta}, \binits{A.}},
\oauthor{\bsnm{Yu}, \binits{T.}},
\oauthor{\bsnm{Tan}, \binits{Y.C.}},
\oauthor{\bsnm{Lin}, \binits{X.V.}},
\oauthor{\bsnm{Xiong}, \binits{C.}},
\oauthor{\bsnm{Socher}, \binits{R.}},
\oauthor{\bsnm{Rajani}, \binits{N.F.}}:
{DART: Open-Domain Structured Data Record to Text Generation}.
NAACL-HLT 2021 - 2021 Conference of the North American Chapter of the Association for Computational Linguistics: Human Language Technologies, Proceedings of the Conference,
432--447
(2021)
\doiurl{10.18653/v1/2021.naacl-main.37}
{\href{https://arxiv.org/abs/2007.02871}{{arXiv:2007.02871}}}
\end{botherref}
\endbibitem

\bibitem[\protect\citeauthoryear{Lu et~al.}{2022}]{Lu2022}
\begin{botherref}
\oauthor{\bsnm{Lu}, \binits{X.}},
\oauthor{\bsnm{Welleck}, \binits{S.}},
\oauthor{\bsnm{West}, \binits{P.}}:
{NEUROLOGIC A$^*$esque Decoding: Constrained Text Generation with Lookahead Heuristics},
780--799
(2022)
\end{botherref}
\endbibitem

\bibitem[\protect\citeauthoryear{Wang et~al.}{2020}]{Wang2020}
\begin{barticle}
\bauthor{\bsnm{Wang}, \binits{T.}},
\bauthor{\bsnm{Wan}, \binits{X.}},
\bauthor{\bsnm{Jin}, \binits{H.}}:
\batitle{Amr-to-text generation with graph transformer}.
\bjtitle{Transactions of the Association for Computational Linguistics}
\bvolume{8},
\bfpage{19}--\blpage{33}
(\byear{2020})
\doiurl{10.1162/tacl_a_00297}
\end{barticle}
\endbibitem

\bibitem[\protect\citeauthoryear{Ribeiro et~al.}{2020}]{Ribeiro2020}
\begin{barticle}
\bauthor{\bsnm{Ribeiro}, \binits{L.F.R.}},
\bauthor{\bsnm{Zhang}, \binits{Y.}},
\bauthor{\bsnm{Gardent}, \binits{C.}},
\bauthor{\bsnm{Gurevych}, \binits{I.}}:
\batitle{{Modeling global and local node contexts for text generation from knowledge graphs}}.
\bjtitle{Transactions of the Association for Computational Linguistics}
\bvolume{8},
\bfpage{589}--\blpage{604}
(\byear{2020})
\doiurl{10.1162/tacl_a_00332}
{\href{https://arxiv.org/abs/2001.11003}{{arXiv:2001.11003}}}
\end{barticle}
\endbibitem

\bibitem[\protect\citeauthoryear{Puduppully and Lapata}{2021}]{puduppully-lapata-2021}
\begin{barticle}
\bauthor{\bsnm{Puduppully}, \binits{R.}},
\bauthor{\bsnm{Lapata}, \binits{M.}}:
\batitle{{Data-to-text Generation with Macro Planning}}.
\bjtitle{Transactions of the Association for Computational Linguistics}
\bvolume{9},
\bfpage{510}--\blpage{527}
(\byear{2021})
\doiurl{10.1162/tacl_a_00381}
\end{barticle}
\endbibitem

\bibitem[\protect\citeauthoryear{Ke et~al.}{2021}]{Ke2021}
\begin{botherref}
\oauthor{\bsnm{Ke}, \binits{P.}},
\oauthor{\bsnm{Ji}, \binits{H.}},
\oauthor{\bsnm{Ran}, \binits{Y.}},
\oauthor{\bsnm{Cui}, \binits{X.}},
\oauthor{\bsnm{Wang}, \binits{L.}},
\oauthor{\bsnm{Song}, \binits{L.}},
\oauthor{\bsnm{Zhu}, \binits{X.}},
\oauthor{\bsnm{Huang}, \binits{M.}}:
{JointGT: Graph-Text Joint Representation Learning for Text Generation from Knowledge Graphs}.
Findings of the Association for Computational Linguistics: ACL-IJCNLP 2021,
2526--2538
(2021)
\doiurl{10.18653/v1/2021.findings-acl.223}
{\href{https://arxiv.org/abs/2106.10502}{{arXiv:2106.10502}}}
\end{botherref}
\endbibitem

\bibitem[\protect\citeauthoryear{Harkous et~al.}{2020}]{Harkous2020}
\begin{botherref}
\oauthor{\bsnm{Harkous}, \binits{H.}},
\oauthor{\bsnm{Groves}, \binits{I.}},
\oauthor{\bsnm{Saffari}, \binits{A.}}:
{Have Your Text and Use It Too! End-to-End Neural Data-to-Text Generation with Semantic Fidelity}.
COLING 2020 - 28th International Conference on Computational Linguistics, Proceedings of the Conference,
2410--2424
(2020)
\doiurl{10.18653/v1/2020.coling-main.218}
{\href{https://arxiv.org/abs/2004.06577}{{arXiv:2004.06577}}}
\end{botherref}
\endbibitem

\bibitem[\protect\citeauthoryear{Gong et~al.}{2020}]{Gong2020a}
\begin{botherref}
\oauthor{\bsnm{Gong}, \binits{H.}},
\oauthor{\bsnm{Sun}, \binits{Y.}},
\oauthor{\bsnm{Feng}, \binits{X.}},
\oauthor{\bsnm{Qin}, \binits{B.}},
\oauthor{\bsnm{Bi}, \binits{W.}},
\oauthor{\bsnm{Liu}, \binits{X.}},
\oauthor{\bsnm{Liu}, \binits{T.}}:
{TableGPT: Few-shot Table-to-Text Generation with Table Structure Reconstruction and Content Matching}.
COLING 2020 - 28th International Conference on Computational Linguistics, Proceedings of the Conference,
1978--1988
(2020)
\doiurl{10.18653/v1/2020.coling-main.179}
\end{botherref}
\endbibitem

\bibitem[\protect\citeauthoryear{Uehara et~al.}{2020}]{Uehara2020}
\begin{botherref}
\oauthor{\bsnm{Uehara}, \binits{Y.}},
\oauthor{\bsnm{Ishigaki}, \binits{T.}},
\oauthor{\bsnm{Aoki}, \binits{K.}},
\oauthor{\bsnm{Goshima}, \binits{K.}},
\oauthor{\bsnm{Noji}, \binits{H.}},
\oauthor{\bsnm{Kobayashi}, \binits{I.}},
\oauthor{\bsnm{Takamura}, \binits{H.}},
\oauthor{\bsnm{Miyao}, \binits{Y.}}:
{Learning with Contrastive Examples for Data-to-Text Generation}.
COLING 2020 - 28th International Conference on Computational Linguistics, Proceedings of the Conference,
2352--2362
(2020)
\doiurl{10.18653/v1/2020.coling-main.213}
\end{botherref}
\endbibitem

\bibitem[\protect\citeauthoryear{Arun et~al.}{2020}]{Arun2020}
\begin{botherref}
\oauthor{\bsnm{Arun}, \binits{A.}},
\oauthor{\bsnm{Batra}, \binits{S.}},
\oauthor{\bsnm{Bhardwaj}, \binits{V.}},
\oauthor{\bsnm{Challa}, \binits{A.}},
\oauthor{\bsnm{Donmez}, \binits{P.}},
\oauthor{\bsnm{Heidari}, \binits{P.}},
\oauthor{\bsnm{Inan}, \binits{H.}},
\oauthor{\bsnm{Jain}, \binits{S.}},
\oauthor{\bsnm{Kumar}, \binits{A.}},
\oauthor{\bsnm{Mei}, \binits{S.}},
\oauthor{\bsnm{Mohan}, \binits{K.}},
\oauthor{\bsnm{White}, \binits{M.}}:
{Best Practices for Data-Efficient Modeling in NLG: How to Train Production-Ready Neural Models with Less Data}.
COLING 2020 - 28th International Conference on Computational Linguistics, Proceedings of the Industry Track,
64--77
(2020)
\doiurl{10.18653/v1/2020.coling-industry.7}
{\href{https://arxiv.org/abs/2011.03877}{{arXiv:2011.03877}}}
\end{botherref}
\endbibitem

\bibitem[\protect\citeauthoryear{Moussallem et~al.}{2019}]{Moussallem2019}
\begin{botherref}
\oauthor{\bsnm{Moussallem}, \binits{D.}},
\oauthor{\bsnm{Ferreira}, \binits{T.C.}},
\oauthor{\bsnm{Zampieri}, \binits{M.}},
\oauthor{\bsnm{Cavalcanti}, \binits{M.C.}},
\oauthor{\bsnm{Xex{\'{e}}o}, \binits{G.}},
\oauthor{\bsnm{Neves}, \binits{M.}},
\oauthor{\bsnm{Ngomo}, \binits{A.C.N.}}:
{Rdf2Pt: Generating brazilian Portuguese texts from RDF data}.
LREC 2018 - 11th International Conference on Language Resources and Evaluation,
3043--3050
(2019)
{\href{https://arxiv.org/abs/1802.08150}{{arXiv:1802.08150}}}
\end{botherref}
\endbibitem

\bibitem[\protect\citeauthoryear{Puduppully et~al.}{2019}]{Puduppully2019}
\begin{botherref}
\oauthor{\bsnm{Puduppully}, \binits{R.}},
\oauthor{\bsnm{Dong}, \binits{L.}},
\oauthor{\bsnm{Lapata}, \binits{M.}}:
{Data-to-text generation with content selection and planning}.
33rd AAAI Conference on Artificial Intelligence, AAAI 2019, 31st Innovative Applications of Artificial Intelligence Conference, IAAI 2019 and the 9th AAAI Symposium on Educational Advances in Artificial Intelligence, EAAI 2019,
6908--6915
(2019)
\doiurl{10.1609/aaai.v33i01.33016908}
{\href{https://arxiv.org/abs/1809.00582}{{arXiv:1809.00582}}}
\end{botherref}
\endbibitem

\bibitem[\protect\citeauthoryear{Liu et~al.}{2021}]{Liu2021}
\begin{barticle}
\bauthor{\bsnm{Liu}, \binits{T.}},
\bauthor{\bsnm{Zheng}, \binits{X.}},
\bauthor{\bsnm{Chang}, \binits{B.}},
\bauthor{\bsnm{Sui}, \binits{Z.}}:
\batitle{{Towards Faithfulness in Open Domain Table-to-text Generation from an Entity-centric View}}.
\bjtitle{35th AAAI Conference on Artificial Intelligence, AAAI 2021}
\bvolume{15},
\bfpage{13415}--\blpage{13423}
(\byear{2021})
\doiurl{10.1609/aaai.v35i15.17583}
{\href{https://arxiv.org/abs/2102.08585}{{arXiv:2102.08585}}}
\end{barticle}
\endbibitem

\bibitem[\protect\citeauthoryear{Gehrmann et~al.}{2018}]{Gehrmann2018}
\begin{botherref}
\oauthor{\bsnm{Gehrmann}, \binits{S.}},
\oauthor{\bsnm{Dai}, \binits{F.Z.}},
\oauthor{\bsnm{Elder}, \binits{H.}},
\oauthor{\bsnm{Rush}, \binits{A.M.}}:
{End-to-End Content and Plan Selection for Data-to-Text Generation}.
E2E NLG Challenge System Descriptions,
46--56
(2018)
{\href{https://arxiv.org/abs/1810.04700v1}{{arXiv:1810.04700v1}}}
\end{botherref}
\endbibitem

\bibitem[\protect\citeauthoryear{Puzikov and Gurevych}{2018}]{Puzikov2018}
\begin{botherref}
\oauthor{\bsnm{Puzikov}, \binits{Y.}},
\oauthor{\bsnm{Gurevych}, \binits{I.}}:
{E2E NLG challenge: Neural models vs. templates}.
INLG 2018 - 11th International Natural Language Generation Conference, Proceedings of the Conference,
463--471
(2018)
\doiurl{10.18653/v1/w18-6557}
\end{botherref}
\endbibitem

\bibitem[\protect\citeauthoryear{Kedzie and McKeown}{2019}]{Kedzie2019}
\begin{botherref}
\oauthor{\bsnm{Kedzie}, \binits{C.}},
\oauthor{\bsnm{McKeown}, \binits{K.}}:
{A good sample is hard to find: Noise injection sampling and self-training for neural language generation models}.
INLG 2019 - 12th International Conference on Natural Language Generation, Proceedings of the Conference,
584--593
(2019)
\doiurl{10.18653/v1/w19-8672}
{\href{https://arxiv.org/abs/1911.03373}{{arXiv:1911.03373}}}
\end{botherref}
\endbibitem

\bibitem[\protect\citeauthoryear{Wang}{2019}]{Wang2019}
\begin{botherref}
\oauthor{\bsnm{Wang}, \binits{H.}}:
{Revisiting challenges in data-to-text generation with fact grounding}.
INLG 2019 - 12th International Conference on Natural Language Generation, Proceedings of the Conference,
311--322
(2019)
\doiurl{10.18653/v1/w19-8639}
{\href{https://arxiv.org/abs/2001.03830}{{arXiv:2001.03830}}}
\end{botherref}
\endbibitem

\bibitem[\protect\citeauthoryear{Qader et~al.}{2019}]{Qader2019}
\begin{botherref}
\oauthor{\bsnm{Qader}, \binits{R.}},
\oauthor{\bsnm{Portet}, \binits{F.}},
\oauthor{\bsnm{Labb{\'{e}}}, \binits{C.}}:
{Semi-supervised neural text generation by joint learning of natural language generation and natural language understanding models}.
INLG 2019 - 12th International Conference on Natural Language Generation, Proceedings of the Conference,
552--562
(2019)
\doiurl{10.18653/v1/w19-8669}
{\href{https://arxiv.org/abs/1910.03484}{{arXiv:1910.03484}}}
\end{botherref}
\endbibitem

\bibitem[\protect\citeauthoryear{Kale}{2020}]{Kale2020}
\begin{botherref}
\oauthor{\bsnm{Kale}, \binits{M.}}:
{T5Pretrain-Text-to-Text Pre-Training for Data-to-Text Tasks.pdf},
97--102
(2020)
\end{botherref}
\endbibitem

\bibitem[\protect\citeauthoryear{Rebuffel et~al.}{2020}]{Rebuffel2020}
\begin{botherref}
\oauthor{\bsnm{Rebuffel}, \binits{C.}},
\oauthor{\bsnm{Soulier}, \binits{L.}},
\oauthor{\bsnm{Scoutheeten}, \binits{G.}},
\oauthor{\bsnm{Gallinari}, \binits{P.}}:
{PARENTing via Model-Agnostic Reinforcement Learning to Correct Pathological Behaviors in Data-to-Text Generation}.
INLG 2020 - 13th International Conference on Natural Language Generation, Proceedings,
120--130
(2020)
{\href{https://arxiv.org/abs/2010.10866}{{arXiv:2010.10866}}}
\end{botherref}
\endbibitem

\bibitem[\protect\citeauthoryear{Juraska and Walker}{2021}]{Juraska2021}
\begin{botherref}
\oauthor{\bsnm{Juraska}, \binits{J.}},
\oauthor{\bsnm{Walker}, \binits{M.}}:
{Attention Is Indeed All You Need: Semantically Attention-Guided Decoding for Data-to-Text NLG}.
INLG 2021 - 14th International Conference on Natural Language Generation, Proceedings
(September),
416--431
(2021)
{\href{https://arxiv.org/abs/2109.07043}{{arXiv:2109.07043}}}
\end{botherref}
\endbibitem

\bibitem[\protect\citeauthoryear{Moussallem et~al.}{2020}]{Moussallem2020}
\begin{barticle}
\bauthor{\bsnm{Moussallem}, \binits{D.}},
\bauthor{\bsnm{Gnaneshwar}, \binits{D.}},
\bauthor{\bsnm{{Castro Ferreira}}, \binits{T.}},
\bauthor{\bsnm{{Ngonga Ngomo}}, \binits{A.C.}}:
\batitle{{NABU – Multilingual Graph-Based Neural RDF Verbalizer}}.
\bjtitle{Lecture Notes in Computer Science (including subseries Lecture Notes in Artificial Intelligence and Lecture Notes in Bioinformatics)}
\bvolume{12506 LNCS},
\bfpage{420}--\blpage{437}
(\byear{2020})
\doiurl{10.1007/978-3-030-62419-4_24}
{\href{https://arxiv.org/abs/arXiv:2009.07728v2}{{arXiv:arXiv:2009.07728v2}}}
\end{barticle}
\endbibitem

\bibitem[\protect\citeauthoryear{{Castro Ferreira} et~al.}{2020}]{castro-ferreira-etal-2020}
\begin{bchapter}
\bauthor{\bsnm{{Castro Ferreira}}, \binits{T.}},
\bauthor{\bsnm{Gardent}, \binits{C.}},
\bauthor{\bsnm{Ilinykh}, \binits{N.}},
\bauthor{\bsnm{Lee}, \binits{C.}},
\bauthor{\bsnm{Mille}, \binits{S.}},
\bauthor{\bsnm{Moussallem}, \binits{D.}},
\bauthor{\bsnm{Shimorina}, \binits{A.}}:
\bctitle{{The 2020 Bilingual, Bi-Directional {W}eb{NLG}+ Shared Task: Overview and Evaluation Results ({W}eb{NLG}+ 2020)}}.
In: \bbtitle{Proceedings of the 3rd International Workshop on Natural Language Generation from the Semantic Web (WebNLG+)},
pp. \bfpage{55}--\blpage{76}.
\bpublisher{Association for Computational Linguistics},
\blocation{Dublin, Ireland (Virtual)}
(\byear{2020}).
\burl{https://aclanthology.org/2020.webnlg-1.7}
\end{bchapter}
\endbibitem

\bibitem[\protect\citeauthoryear{Agarwal et~al.}{2020}]{Agarwal2020}
\begin{botherref}
\oauthor{\bsnm{Agarwal}, \binits{O.}},
\oauthor{\bsnm{Kale}, \binits{M.}},
\oauthor{\bsnm{Ge}, \binits{H.}},
\oauthor{\bsnm{Shakeri}, \binits{S.}},
\oauthor{\bsnm{Al-Rfou}, \binits{R.}}:
{Machine Translation Aided Bilingual Data-to-Text Generation and Semantic Parsing}.
Proceedings of the 3rd International Workshop on Natural Language Generation from the Semantic Web (WebNLG+)
(December),
125--130
(2020)
\end{botherref}
\endbibitem

\bibitem[\protect\citeauthoryear{Li et~al.}{2020}]{Li2020}
\begin{botherref}
\oauthor{\bsnm{Li}, \binits{X.}},
\oauthor{\bsnm{Maskharashvili}, \binits{A.}},
\oauthor{\bsnm{{Jory Stevens-Guille}}, \binits{S.}},
\oauthor{\bsnm{White}, \binits{M.}}:
{Leveraging Large Pretrained Models for {W}eb{NLG} 2020}.
Proceedings of the 3rd International Workshop on Natural Language Generation from the Semantic Web (WebNLG+)
(December),
117--124
(2020)
\end{botherref}
\endbibitem

\bibitem[\protect\citeauthoryear{Guo et~al.}{2020}]{Guo2020}
\begin{botherref}
\oauthor{\bsnm{Guo}, \binits{Q.}},
\oauthor{\bsnm{Jin}, \binits{Z.}},
\oauthor{\bsnm{Dai}, \binits{N.}},
\oauthor{\bsnm{Qiu}, \binits{X.}},
\oauthor{\bsnm{Xue}, \binits{X.}},
\oauthor{\bsnm{Wipf}, \binits{D.}},
\oauthor{\bsnm{Zhang}, \binits{Z.}}:
{$\mathcal{P}_2$ : A Plan-and-Pretrain Approach for Knowledge Graph-to-Text Generation}
(December),
100--106
(2020)
\end{botherref}
\endbibitem

\bibitem[\protect\citeauthoryear{Rebuffel et~al.}{2022}]{Rebuffel2022}
\begin{barticle}
\bauthor{\bsnm{Rebuffel}, \binits{C.}},
\bauthor{\bsnm{Roberti}, \binits{M.}},
\bauthor{\bsnm{Soulier}, \binits{L.}},
\bauthor{\bsnm{Scoutheeten}, \binits{G.}},
\bauthor{\bsnm{Cancelliere}, \binits{R.}},
\bauthor{\bsnm{Gallinari}, \binits{P.}}:
\batitle{{Controlling hallucinations at word level in data-to-text generation}}.
\bjtitle{Data Mining and Knowledge Discovery}
\bvolume{36}(\bissue{1}),
\bfpage{318}--\blpage{354}
(\byear{2022})
\doiurl{10.1007/s10618-021-00801-4}
{\href{https://arxiv.org/abs/2102.02810}{{arXiv:2102.02810}}}
\end{barticle}
\endbibitem

\bibitem[\protect\citeauthoryear{Novikova et~al.}{2017}]{novikova2017e2e}
\begin{botherref}
\oauthor{\bsnm{Novikova}, \binits{J.}},
\oauthor{\bsnm{Du{\v{s}}ek}, \binits{O.}},
\oauthor{\bsnm{Rieser}, \binits{V.}}:
The e2e dataset: New challenges for end-to-end generation.
arXiv preprint arXiv:1706.09254
(2017)
\end{botherref}
\endbibitem

\bibitem[\protect\citeauthoryear{Banarescu et~al.}{2013}]{banarescu2013abstract}
\begin{bchapter}
\bauthor{\bsnm{Banarescu}, \binits{L.}},
\bauthor{\bsnm{Bonial}, \binits{C.}},
\bauthor{\bsnm{Cai}, \binits{S.}},
\bauthor{\bsnm{Georgescu}, \binits{M.}},
\bauthor{\bsnm{Griffitt}, \binits{K.}},
\bauthor{\bsnm{Hermjakob}, \binits{U.}},
\bauthor{\bsnm{Knight}, \binits{K.}},
\bauthor{\bsnm{Koehn}, \binits{P.}},
\bauthor{\bsnm{Palmer}, \binits{M.}},
\bauthor{\bsnm{Schneider}, \binits{N.}}:
\bctitle{Abstract meaning representation for sembanking}.
In: \bbtitle{Proceedings of the 7th Linguistic Annotation Workshop and Interoperability with Discourse},
pp. \bfpage{178}--\blpage{186}
(\byear{2013})
\end{bchapter}
\endbibitem

\bibitem[\protect\citeauthoryear{Goodman}{2020}]{goodman2020penman}
\begin{bchapter}
\bauthor{\bsnm{Goodman}, \binits{M.W.}}:
\bctitle{Penman: An open-source library and tool for amr graphs}.
In: \bbtitle{Proceedings of the 58th Annual Meeting of the Association for Computational Linguistics: System Demonstrations},
pp. \bfpage{312}--\blpage{319}
(\byear{2020})
\end{bchapter}
\endbibitem

\bibitem[\protect\citeauthoryear{Tiedemann}{2012}]{tiedemann2012parallel}
\begin{bchapter}
\bauthor{\bsnm{Tiedemann}, \binits{J.}}:
\bctitle{Parallel data, tools and interfaces in opus.}
In: \bbtitle{Lrec},
vol. \bseriesno{2012},
pp. \bfpage{2214}--\blpage{2218}
(\byear{2012}).
\bcomment{Citeseer}
\end{bchapter}
\endbibitem

\bibitem[\protect\citeauthoryear{Bojar et~al.}{2017}]{Bojar2017}
\begin{barticle}
\bauthor{\bsnm{Bojar}, \binits{O.}},
\bauthor{\bsnm{Chatterjee}, \binits{R.}},
\bauthor{\bsnm{Federmann}, \binits{C.}},
\bauthor{\bsnm{Graham}, \binits{Y.}},
\bauthor{\bsnm{Haddow}, \binits{B.}},
\bauthor{\bsnm{Huang}, \binits{S.}},
\bauthor{\bsnm{Huck}, \binits{M.}},
\bauthor{\bsnm{Koehn}, \binits{P.}},
\bauthor{\bsnm{Liu}, \binits{Q.}},
\bauthor{\bsnm{Logacheva}, \binits{V.}},
\bauthor{\bsnm{Monz}, \binits{C.}},
\bauthor{\bsnm{Negri}, \binits{M.}},
\bauthor{\bsnm{Post}, \binits{M.}},
\bauthor{\bsnm{Rubino}, \binits{R.}},
\bauthor{\bsnm{Specia}, \binits{L.}},
\bauthor{\bsnm{Turchi}, \binits{M.}}:
\batitle{Findings of the 2017 conference on machine translation (wmt17)}.
\bjtitle{WMT 2017 - 2nd Conference on Machine Translation, Proceedings}
\bvolume{2},
\bfpage{169}--\blpage{214}
(\byear{2017})
\doiurl{10.18653/V1/W17-4717}
\end{barticle}
\endbibitem

\bibitem[\protect\citeauthoryear{Barrault et~al.}{2019}]{Barrault2019}
\begin{barticle}
\bauthor{\bsnm{Barrault}, \binits{L.}},
\bauthor{\bsnm{Biesialska}, \binits{M.}},
\bauthor{\bsnm{Bojar}, \binits{O.}},
\bauthor{\bsnm{Costa-Jussà}, \binits{M.R.}},
\bauthor{\bsnm{Federmann}, \binits{C.}},
\bauthor{\bsnm{Graham}, \binits{Y.}},
\bauthor{\bsnm{Grundkiewicz}, \binits{R.}},
\bauthor{\bsnm{Haddow}, \binits{B.}},
\bauthor{\bsnm{Huck}, \binits{M.}},
\bauthor{\bsnm{Joanis}, \binits{E.}},
\bauthor{\bsnm{Kocmi}, \binits{T.}},
\bauthor{\bsnm{Koehn}, \binits{P.}},
\bauthor{\bsnm{Lo}, \binits{C.K.}},
\bauthor{\bsnm{Ljubešić}, \binits{N.}},
\bauthor{\bsnm{Monz}, \binits{C.}},
\bauthor{\bsnm{Morishita}, \binits{M.}},
\bauthor{\bsnm{Nagata}, \binits{M.}},
\bauthor{\bsnm{Nakazawa}, \binits{T.}},
\bauthor{\bsnm{Pal}, \binits{S.}},
\bauthor{\bsnm{Post}, \binits{M.}},
\bauthor{\bsnm{Zampieri}, \binits{M.}}:
\batitle{Findings of the 2019 conference on machine translation (wmt19)}.
\bjtitle{5th Conference on Machine Translation, WMT 2020 - Proceedings}
\bvolume{2},
\bfpage{1}--\blpage{61}
(\byear{2019})
\doiurl{10.18653/V1/W19-5301}
\end{barticle}
\endbibitem

\bibitem[\protect\citeauthoryear{Bahdanau et~al.}{2014}]{bahdanau2014neural}
\begin{botherref}
\oauthor{\bsnm{Bahdanau}, \binits{D.}},
\oauthor{\bsnm{Cho}, \binits{K.}},
\oauthor{\bsnm{Bengio}, \binits{Y.}}:
Neural machine translation by jointly learning to align and translate.
arXiv preprint arXiv:1409.0473
(2014)
\end{botherref}
\endbibitem

\bibitem[\protect\citeauthoryear{Vaswani et~al.}{2017}]{vaswani2017attention}
\begin{botherref}
\oauthor{\bsnm{Vaswani}, \binits{A.}},
\oauthor{\bsnm{Shazeer}, \binits{N.}},
\oauthor{\bsnm{Parmar}, \binits{N.}},
\oauthor{\bsnm{Uszkoreit}, \binits{J.}},
\oauthor{\bsnm{Jones}, \binits{L.}},
\oauthor{\bsnm{Gomez}, \binits{A.N.}},
\oauthor{\bsnm{Kaiser}, \binits{{\L}.}},
\oauthor{\bsnm{Polosukhin}, \binits{I.}}:
Attention is all you need.
Advances in neural information processing systems
\textbf{30}
(2017)
\end{botherref}
\endbibitem

\bibitem[\protect\citeauthoryear{Hochreiter and Schmidhuber}{1997}]{hochreiter1997long}
\begin{barticle}
\bauthor{\bsnm{Hochreiter}, \binits{S.}},
\bauthor{\bsnm{Schmidhuber}, \binits{J.}}:
\batitle{Long short-term memory}.
\bjtitle{Neural computation}
\bvolume{9}(\bissue{8}),
\bfpage{1735}--\blpage{1780}
(\byear{1997})
\end{barticle}
\endbibitem

\bibitem[\protect\citeauthoryear{Cho et~al.}{2014}]{cho2014properties}
\begin{botherref}
\oauthor{\bsnm{Cho}, \binits{K.}},
\oauthor{\bsnm{Van~Merri{\"e}nboer}, \binits{B.}},
\oauthor{\bsnm{Bahdanau}, \binits{D.}},
\oauthor{\bsnm{Bengio}, \binits{Y.}}:
On the properties of neural machine translation: Encoder-decoder approaches.
arXiv preprint arXiv:1409.1259
(2014)
\end{botherref}
\endbibitem

\bibitem[\protect\citeauthoryear{Raffel et~al.}{2020}]{raffel2020exploring}
\begin{barticle}
\bauthor{\bsnm{Raffel}, \binits{C.}},
\bauthor{\bsnm{Shazeer}, \binits{N.}},
\bauthor{\bsnm{Roberts}, \binits{A.}},
\bauthor{\bsnm{Lee}, \binits{K.}},
\bauthor{\bsnm{Narang}, \binits{S.}},
\bauthor{\bsnm{Matena}, \binits{M.}},
\bauthor{\bsnm{Zhou}, \binits{Y.}},
\bauthor{\bsnm{Li}, \binits{W.}},
\bauthor{\bsnm{Liu}, \binits{P.J.}}:
\batitle{Exploring the limits of transfer learning with a unified text-to-text transformer}.
\bjtitle{The Journal of Machine Learning Research}
\bvolume{21}(\bissue{1}),
\bfpage{5485}--\blpage{5551}
(\byear{2020})
\end{barticle}
\endbibitem

\bibitem[\protect\citeauthoryear{Lewis et~al.}{2019}]{lewis2019bart}
\begin{botherref}
\oauthor{\bsnm{Lewis}, \binits{M.}},
\oauthor{\bsnm{Liu}, \binits{Y.}},
\oauthor{\bsnm{Goyal}, \binits{N.}},
\oauthor{\bsnm{Ghazvininejad}, \binits{M.}},
\oauthor{\bsnm{Mohamed}, \binits{A.}},
\oauthor{\bsnm{Levy}, \binits{O.}},
\oauthor{\bsnm{Stoyanov}, \binits{V.}},
\oauthor{\bsnm{Zettlemoyer}, \binits{L.}}:
Bart: Denoising sequence-to-sequence pre-training for natural language generation, translation, and comprehension.
arXiv preprint arXiv:1910.13461
(2019)
\end{botherref}
\endbibitem

\bibitem[\protect\citeauthoryear{Lample and Conneau}{2019}]{lample2019cross}
\begin{botherref}
\oauthor{\bsnm{Lample}, \binits{G.}},
\oauthor{\bsnm{Conneau}, \binits{A.}}:
Cross-lingual language model pretraining.
arXiv preprint arXiv:1901.07291
(2019)
\end{botherref}
\endbibitem

\bibitem[\protect\citeauthoryear{Radford et~al.}{2019}]{radford2019language}
\begin{barticle}
\bauthor{\bsnm{Radford}, \binits{A.}},
\bauthor{\bsnm{Wu}, \binits{J.}},
\bauthor{\bsnm{Child}, \binits{R.}},
\bauthor{\bsnm{Luan}, \binits{D.}},
\bauthor{\bsnm{Amodei}, \binits{D.}},
\bauthor{\bsnm{Sutskever}, \binits{I.}}, \betal:
\batitle{Language models are unsupervised multitask learners}.
\bjtitle{OpenAI blog}
\bvolume{1}(\bissue{8}),
\bfpage{9}
(\byear{2019})
\end{barticle}
\endbibitem

\bibitem[\protect\citeauthoryear{Papineni et~al.}{2002}]{papineni2002bleu}
\begin{bchapter}
\bauthor{\bsnm{Papineni}, \binits{K.}},
\bauthor{\bsnm{Roukos}, \binits{S.}},
\bauthor{\bsnm{Ward}, \binits{T.}},
\bauthor{\bsnm{Zhu}, \binits{W.-J.}}:
\bctitle{Bleu: a method for automatic evaluation of machine translation}.
In: \bbtitle{Proceedings of the 40th Annual Meeting of the Association for Computational Linguistics},
pp. \bfpage{311}--\blpage{318}
(\byear{2002})
\end{bchapter}
\endbibitem

\bibitem[\protect\citeauthoryear{Denkowski and Lavie}{2014}]{denkowski2014meteor}
\begin{bchapter}
\bauthor{\bsnm{Denkowski}, \binits{M.}},
\bauthor{\bsnm{Lavie}, \binits{A.}}:
\bctitle{Meteor universal: Language specific translation evaluation for any target language}.
In: \bbtitle{Proceedings of the Ninth Workshop on Statistical Machine Translation},
pp. \bfpage{376}--\blpage{380}
(\byear{2014})
\end{bchapter}
\endbibitem

\bibitem[\protect\citeauthoryear{Lin}{2004}]{lin2004rouge}
\begin{bchapter}
\bauthor{\bsnm{Lin}, \binits{C.-Y.}}:
\bctitle{Rouge: A package for automatic evaluation of summaries}.
In: \bbtitle{Text Summarization Branches Out},
pp. \bfpage{74}--\blpage{81}
(\byear{2004})
\end{bchapter}
\endbibitem

\bibitem[\protect\citeauthoryear{Popovi{\'c}}{2015}]{popovic2015chrf}
\begin{bchapter}
\bauthor{\bsnm{Popovi{\'c}}, \binits{M.}}:
\bctitle{chrf: character n-gram f-score for automatic mt evaluation}.
In: \bbtitle{Proceedings of the Tenth Workshop on Statistical Machine Translation},
pp. \bfpage{392}--\blpage{395}
(\byear{2015})
\end{bchapter}
\endbibitem

\bibitem[\protect\citeauthoryear{Doddington}{2002}]{doddington2002automatic}
\begin{bchapter}
\bauthor{\bsnm{Doddington}, \binits{G.}}:
\bctitle{Automatic evaluation of machine translation quality using n-gram co-occurrence statistics}.
In: \bbtitle{Proceedings of the Second International Conference on Human Language Technology Research},
pp. \bfpage{138}--\blpage{145}
(\byear{2002})
\end{bchapter}
\endbibitem

\bibitem[\protect\citeauthoryear{Snover et~al.}{2006}]{snover-etal-2006-study}
\begin{bchapter}
\bauthor{\bsnm{Snover}, \binits{M.}},
\bauthor{\bsnm{Dorr}, \binits{B.}},
\bauthor{\bsnm{Schwartz}, \binits{R.}},
\bauthor{\bsnm{Micciulla}, \binits{L.}},
\bauthor{\bsnm{Makhoul}, \binits{J.}}:
\bctitle{A study of translation edit rate with targeted human annotation}.
In: \bbtitle{Proceedings of the 7th Conference of the Association for Machine Translation in the Americas: Technical Papers},
pp. \bfpage{223}--\blpage{231}.
\bpublisher{Association for Machine Translation in the Americas},
\blocation{Cambridge, Massachusetts, USA}
(\byear{2006}).
\burl{https://aclanthology.org/2006.amta-papers.25}
\end{bchapter}
\endbibitem

\bibitem[\protect\citeauthoryear{Vedantam et~al.}{2015}]{vedantam2015cider}
\begin{bchapter}
\bauthor{\bsnm{Vedantam}, \binits{R.}},
\bauthor{\bsnm{Lawrence~Zitnick}, \binits{C.}},
\bauthor{\bsnm{Parikh}, \binits{D.}}:
\bctitle{Cider: Consensus-based image description evaluation}.
In: \bbtitle{Proceedings of the IEEE Conference on Computer Vision and Pattern Recognition},
pp. \bfpage{4566}--\blpage{4575}
(\byear{2015})
\end{bchapter}
\endbibitem

\bibitem[\protect\citeauthoryear{Sai et~al.}{2022}]{sai2022survey}
\begin{barticle}
\bauthor{\bsnm{Sai}, \binits{A.B.}},
\bauthor{\bsnm{Mohankumar}, \binits{A.K.}},
\bauthor{\bsnm{Khapra}, \binits{M.M.}}:
\batitle{A survey of evaluation metrics used for nlg systems}.
\bjtitle{ACM Computing Surveys (CSUR)}
\bvolume{55}(\bissue{2}),
\bfpage{1}--\blpage{39}
(\byear{2022})
\end{barticle}
\endbibitem

\bibitem[\protect\citeauthoryear{Anderson et~al.}{2016}]{anderson2016spice}
\begin{bchapter}
\bauthor{\bsnm{Anderson}, \binits{P.}},
\bauthor{\bsnm{Fernando}, \binits{B.}},
\bauthor{\bsnm{Johnson}, \binits{M.}},
\bauthor{\bsnm{Gould}, \binits{S.}}:
\bctitle{Spice: Semantic propositional image caption evaluation}.
In: \bbtitle{Computer Vision--ECCV 2016: 14th European Conference, Amsterdam, The Netherlands, October 11-14, 2016, Proceedings, Part V 14},
pp. \bfpage{382}--\blpage{398}
(\byear{2016}).
\bcomment{Springer}
\end{bchapter}
\endbibitem

\bibitem[\protect\citeauthoryear{Zhang et~al.}{2019}]{zhang2019bertscore}
\begin{botherref}
\oauthor{\bsnm{Zhang}, \binits{T.}},
\oauthor{\bsnm{Kishore}, \binits{V.}},
\oauthor{\bsnm{Wu}, \binits{F.}},
\oauthor{\bsnm{Weinberger}, \binits{K.Q.}},
\oauthor{\bsnm{Artzi}, \binits{Y.}}:
Bertscore: Evaluating text generation with bert.
arXiv preprint arXiv:1904.09675
(2019)
\end{botherref}
\endbibitem

\bibitem[\protect\citeauthoryear{Zhao et~al.}{2019}]{zhao2019moverscore}
\begin{botherref}
\oauthor{\bsnm{Zhao}, \binits{W.}},
\oauthor{\bsnm{Peyrard}, \binits{M.}},
\oauthor{\bsnm{Liu}, \binits{F.}},
\oauthor{\bsnm{Gao}, \binits{Y.}},
\oauthor{\bsnm{Meyer}, \binits{C.M.}},
\oauthor{\bsnm{Eger}, \binits{S.}}:
Moverscore: Text generation evaluating with contextualized embeddings and earth mover distance.
arXiv preprint arXiv:1909.02622
(2019)
\end{botherref}
\endbibitem

\bibitem[\protect\citeauthoryear{Sellam et~al.}{2020}]{sellam2020bleurt}
\begin{botherref}
\oauthor{\bsnm{Sellam}, \binits{T.}},
\oauthor{\bsnm{Das}, \binits{D.}},
\oauthor{\bsnm{Parikh}, \binits{A.P.}}:
Bleurt: Learning robust metrics for text generation.
arXiv preprint arXiv:2004.04696
(2020)
\end{botherref}
\endbibitem

\bibitem[\protect\citeauthoryear{Van Der~Lee et~al.}{2019}]{van2019best}
\begin{bchapter}
\bauthor{\bsnm{Van Der~Lee}, \binits{C.}},
\bauthor{\bsnm{Gatt}, \binits{A.}},
\bauthor{\bsnm{Van~Miltenburg}, \binits{E.}},
\bauthor{\bsnm{Wubben}, \binits{S.}},
\bauthor{\bsnm{Krahmer}, \binits{E.}}:
\bctitle{Best practices for the human evaluation of automatically generated text}.
In: \bbtitle{Proceedings of the 12th International Conference on Natural Language Generation},
pp. \bfpage{355}--\blpage{368}
(\byear{2019})
\end{bchapter}
\endbibitem

\bibitem[\protect\citeauthoryear{Belz et~al.}{2020}]{belz2020disentangling}
\begin{bchapter}
\bauthor{\bsnm{Belz}, \binits{A.}},
\bauthor{\bsnm{Mille}, \binits{S.}},
\bauthor{\bsnm{Howcroft}, \binits{D.M.}}:
\bctitle{Disentangling the properties of human evaluation methods: A classification system to support comparability, meta-evaluation and reproducibility testing}.
(\byear{2020}).
\bcomment{Association for Computational Linguistics (ACL)}
\end{bchapter}
\endbibitem

\bibitem[\protect\citeauthoryear{Ye et~al.}{2023}]{ye2023comprehensive}
\begin{botherref}
\oauthor{\bsnm{Ye}, \binits{J.}},
\oauthor{\bsnm{Chen}, \binits{X.}},
\oauthor{\bsnm{Xu}, \binits{N.}},
\oauthor{\bsnm{Zu}, \binits{C.}},
\oauthor{\bsnm{Shao}, \binits{Z.}},
\oauthor{\bsnm{Liu}, \binits{S.}},
\oauthor{\bsnm{Cui}, \binits{Y.}},
\oauthor{\bsnm{Zhou}, \binits{Z.}},
\oauthor{\bsnm{Gong}, \binits{C.}},
\oauthor{\bsnm{Shen}, \binits{Y.}}, et al.:
A comprehensive capability analysis of gpt-3 and gpt-3.5 series models.
arXiv preprint arXiv:2303.10420
(2023)
\end{botherref}
\endbibitem

\bibitem[\protect\citeauthoryear{Achiam et~al.}{2023}]{Achiam2023GPT4TR}
\begin{bchapter}
\bauthor{\bsnm{Achiam}, \binits{O.J.}},
\bauthor{\bsnm{Adler}, \binits{S.}},
\bauthor{\bsnm{Agarwal}, \binits{S.}},
\bauthor{\bsnm{Ahmad}, \binits{L.}},
\bauthor{\bsnm{Akkaya}, \binits{I.}},
\bauthor{\bsnm{Aleman}, \binits{F.L.}},
\bauthor{\bsnm{Diogo}}, \betal:
\bctitle{Gpt-4 technical report}.
(\byear{2023}).
\burl{https://api.semanticscholar.org/CorpusID:257532815}
\end{bchapter}
\endbibitem

\bibitem[\protect\citeauthoryear{Touvron et~al.}{2023a}]{touvron2023llama}
\begin{botherref}
\oauthor{\bsnm{Touvron}, \binits{H.}},
\oauthor{\bsnm{Lavril}, \binits{T.}},
\oauthor{\bsnm{Izacard}, \binits{G.}},
\oauthor{\bsnm{Martinet}, \binits{X.}},
\oauthor{\bsnm{Lachaux}, \binits{M.-A.}},
\oauthor{\bsnm{Lacroix}, \binits{T.}},
\oauthor{\bsnm{Rozi{\`e}re}, \binits{B.}},
\oauthor{\bsnm{Goyal}, \binits{N.}},
\oauthor{\bsnm{Hambro}, \binits{E.}},
\oauthor{\bsnm{Azhar}, \binits{F.}}, et al.:
Llama: Open and efficient foundation language models.
arXiv preprint arXiv:2302.13971
(2023)
\end{botherref}
\endbibitem

\bibitem[\protect\citeauthoryear{Touvron et~al.}{2023b}]{touvron2023llama2}
\begin{botherref}
\oauthor{\bsnm{Touvron}, \binits{H.}},
\oauthor{\bsnm{Martin}, \binits{L.}},
\oauthor{\bsnm{Stone}, \binits{K.}},
\oauthor{\bsnm{Albert}, \binits{P.}},
\oauthor{\bsnm{Almahairi}, \binits{A.}},
\oauthor{\bsnm{Babaei}, \binits{Y.}},
\oauthor{\bsnm{Bashlykov}, \binits{N.}},
\oauthor{\bsnm{Batra}, \binits{S.}},
\oauthor{\bsnm{Bhargava}, \binits{P.}},
\oauthor{\bsnm{Bhosale}, \binits{S.}}, et al.:
Llama 2: Open foundation and fine-tuned chat models.
arXiv preprint arXiv:2307.09288
(2023)
\end{botherref}
\endbibitem

\end{thebibliography}

\end{document}